\documentclass[times,twocolumn,final,authoryear]{elsarticle}

\usepackage{diagbox}
\usepackage{CJK}

\newcommand{\tabincell}[2]{\begin{tabular}{@{}#1@{}}#2\end{tabular}}

\usepackage{ycviu}
\usepackage{framed,multirow}

\usepackage{amssymb}
\usepackage{latexsym}

\usepackage{bbding}
\usepackage{pifont}

\newcommand{\cmark}{\ding{51}}%
\newcommand{\xmark}{\ding{55}}%

\usepackage{url}
\usepackage{xcolor}
\definecolor{newcolor}{rgb}{.8,.349,.1}

\journal{Computer Vision and Image Understanding}

\begin{document}
\clearpage
\ifpreprint
  \setcounter{page}{1}
\else
  \setcounter{page}{1}
\fi

\begin{frontmatter}
\title{Monocular Human Pose Estimation: A Survey of Deep Learning-based Methods}

\author[1]{Yucheng Chen\corref{cor1}}
\ead{chenyucheng@mail.nwpu.edu.cn}

\author[2]{Yingli Tian\corref{cor2}}
\ead{ytian@ccny.cuny.edu}
\cortext[cor2]{Corresponding author}
\author[1]{Mingyi He\corref{cor3}}
\ead{myhe@nwpu.edu.cn}

\address[1]{Northwestern Polytechnical University, Xi'an, China}
\address[2]{The City College, City University of New York, NY 10031, USA}

\received{1 May 2013}
\finalform{10 May 2013}
\accepted{13 May 2013}
\availableonline{15 May 2013}
\communicated{S. Sarkar}

\begin{abstract}
Vision-based monocular human pose estimation, as one of the most fundamental and challenging problems in computer vision, aims to obtain posture of the human body from input images or video sequences. The recent developments of deep learning techniques have been brought significant progress and remarkable breakthroughs in the field of human pose estimation. This survey extensively reviews the recent deep learning-based 2D and 3D human pose estimation methods published since 2014. This paper summarizes the challenges, main frameworks, benchmark datasets, evaluation metrics, performance comparison, and discusses some promising future research directions.
\end{abstract}

\end{frontmatter}


\section{Introduction}\label{sec1}

The human pose estimation (HPE) task, which has been developed for decades, aims to obtain posture of the human body from given sensor inputs. Vision-based approaches are often used to provide such a solution by using cameras. In recent years, with deep learning shows good performance on many computer version tasks such as image classification \citep{krizhevsky2012imagenet}, object detection \citep{ren2015faster}, semantic segmentation \citep{long2015fully}, etc., HPE also achieves rapid progress by employing deep learning technology. The main developments include well-designed networks with great estimation capability, richer datasets \citep{lin2014microsoft, Joo_2017_TPAMI, mehta2017monocular} for feeding networks and more practical exploration of body models \citep{loper2015smpl, kanazawa2018end}. Although there are some existing reviews for HPE, however, there still lacks a survey to summarize the most recent deep learning-based achievements. This paper extensively reviews deep learning-based 2D/3D human pose estimation methods from monocular images or video footage of humans. Algorithms relied on other sensors such as depth \citep{shotton2012efficient}, infrared light source \citep{faessler2014monocular}, radio frequency signal \citep{zhao2018through}, and multi-view inputs \citep{rhodin2018learning} are not included in this survey.

As one of the fundamental computer vision tasks, HPE is a very important research field and can be applied to many applications such as action/activity recognition \citep{li2017Skeleton, luvizon20182d, li20183d}, action detection \citep{li2017Skeletonbox}, human tracking \citep{insafutdinov2017arttrack}, Movies and animation, Virtual reality, Human-computer interaction, Video surveillance, Medical assistance, Self-driving, Sports motion analysis, etc.

\textit{Movies and animation:} The generation of various vivid digital characters is inseparable from the capture of human movements. Cheap and accurate human motion capture system can better promote the development of the digital entertainment industry.

\textit{Virtual reality:} Virtual reality is a very promising technology that can be applied in both education and entertainment. Estimation of human posture can further clarify the relation between human and virtual reality world and enhance the interactive experience.

\textit{Human–computer interaction (HCI):} HPE is very important for computers and robots to better understand the identification, location, and action of people. With the posture of human (e.g. gesture), computers and robots can execute instructions in an easy way and be more intelligent.

\textit{Video surveillance:} Video surveillance is one of the early applications to adopt HPE technology in tracking, action recognition, re-identification people within a specific range.

\textit{Medical assistance:} In the application of medical assistance, HPE can provide physicians with quantitative human motion information especially for rehabilitation training and physical therapy.

\textit{Self-driving:} Advanced self-driving has been developed rapidly. With HPE, self-driving cars can respond more appropriately to pedestrians and offer more comprehensive interaction with traffic coordinators.

\textit{Sport motion analysis:} Estimating players' posture in sport videos can further obtain the statistics of athletes' indicators (e.g. running distance, number of jumps). During training, HPE can provide a quantitative analysis of action details. In physical education, instructors can make more objective evaluations of students with HPE.

\begin{figure}[t]
\centering
\includegraphics[width=0.46\textwidth]{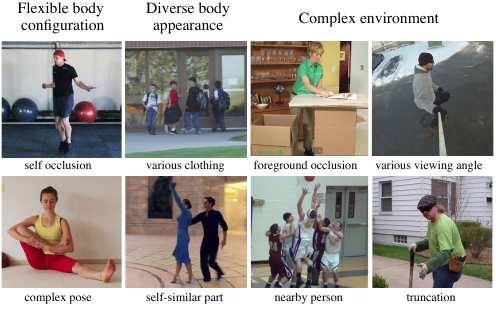}
\caption{Typical challenges of HPE in monocular images or videos. Example images are from Max Planck Institute for Informatics (MPII) dataset \citep{andriluka20142d}.}
\label{fig:challenges}
\end{figure}

Monocular human pose estimation has some unique characteristics and challenges.  As shown in Fig. \ref{fig:challenges}, the challenges of human pose estimation mainly fall in three aspects:

\begin{itemize}
\item Flexible body configuration indicates complex interdependent joints and high degree-of-freedom limbs, which may cause self-occlusions or rare/complex poses.
\item Diverse body appearance includes different clothing and self-similar parts.
\item Complex environment may cause foreground occlusion, occlusion or similar parts from nearby persons, various viewing angles, and truncation in the camera view.
\end{itemize}

The papers of human pose estimation can be categorized in different ways. Based on whether to use designed human body models or not, the methods can be categorized into generative methods (model-based) and discriminative methods (model-free). According to from which level (high-level abstraction or low-level pixel evidence) to start the processing, they can be classified into top-down methods and bottom-up methods. More details of different category strategies for HPE approaches are summarized in Table \ref{tab:overview_all} and described in Section \ref{sec2.1}.

As listed in Table \ref{tab:surveys}, with the development of human pose estimation in the past decades, several notable surveys summarized the research work in this area. The surveys \citep{aggarwal1999human, gavrila1999visual, poppe2007vision, ji2010advances, moeslund2011visual} reviewed the early work of human motion analysis in many aspects (e.g., detection and tracking, pose estimation, recognition) and described the relation between human pose estimation and other related tasks.
While \citet{hu2004survey} summarized the research of human motion analysis for video surveillance application, the reviews \citep{moeslund2001survey, moeslund2006survey} focused on the human motion capture systems.  More recent surveys were mainly focusing on relatively narrow directions, such as RGB-D-based action recognition\citep{chen2013survey, wang2018rgb}, 3D HPE \citep{sminchisescu20083d, holte2012human, sarafianos20163d}, model-based HPE \citep{holte2012human, perez2014survey}, body parts-based HPE \citep{liu2015survey}, and monocular-based HPE \citep{sminchisescu20083d, gong2016human}.

Different from existing review papers, this survey extensively summarizes the recent milestone work of deep learning-based human pose estimation methods, which were mainly published from 2014. In order to provide a comprehensive summary, this survey includes a few research work which has been discussed in some surveys \citep{liu2015survey, gong2016human, sarafianos20163d}, but most of the recent advances are not been presented in any survey before.

The remainder of this paper is organized as follows.
Section \ref{sec2} introduces the existing review papers for human motion analysis and HPE, different ways to category HPE methods, and the widely used human body models. Sections \ref{sec3} and \ref{sec4} describe 2D HPE and 3D HPE approaches respectively. In each section, we further describe HPE approaches for both single person pose estimation and multi-person pose estimation. Since data are a very important and fundamental element for deep learning-based methods, the recent HPE datasets and the evaluation metrics are summarized in Section \ref{sec5}. Finally, Section \ref{sec6} concludes the paper and discusses several promising future research directions.

\begin{table*}[!htb]
\small
    \centering
    \footnotesize
    \caption{Summary of the related surveys of human motion analysis and HPE.}
    \label{tab:surveys}
    \begin{tabular}{|c|p{6cm}<{\centering}|c|p{8.5cm}<{\centering}|}
        \hline
        No. & Survey $\&$ Reference & Venue & Content \\
        \hline
        \tabincell{c}{1} & Human motion analysis: A review \citep{aggarwal1999human} & CVIU
        & A review of human motion analysis including body structure analysis, motion tracking and action recognition.\\
        \hline

        2 & The visual analysis of human movement: A survey \citep{gavrila1999visual} & CVIU
        &  A survey of whole-body and hand motion analysis.\\
        \hline

        3 & A survey of computer vision-based human motion capture \citep{moeslund2001survey} & CVIU
        & An overview based on motion capture system, including initialization, tracking, pose estimation, and recognition.\\
        \hline

        4 & A survey on visual surveillance of object motion and behaviors \citep{hu2004survey} & TSMCS
        & A summary of human motion analysis based one the framework of visual surveillance in dynamic scenes. \\
        \hline

        5 & A survey of advances in vision-based human motion capture and analysis \citep{moeslund2006survey} & CVIU
        & Further summary of human motion capture and analysis from 2000 to 2006, following \citep{moeslund2001survey}.\\
        \hline

        6 & Vision-based human motion analysis: An overview \citep{poppe2007vision} & CVIU
        & A summary of vision-based human motion analysis with markerless data. \\
        \hline

        7 & 3D human motion analysis in monocular video: techniques and challenges \citep{sminchisescu20083d} & \tabincell{c}{Book\\ Chapter}
        & An overview of reconstructing 3D human motion with video sequences from single-view camera. \\
        \hline

        8 & Advances in view-invariant human motion analysis: A review \citep{ji2010advances} & TSMCS
        & A summary of human motion analysis, including human detection, view-invariant pose representation and estimation, and behavior understanding. \\
        \hline

        9 & Visual analysis of humans \citep{moeslund2011visual} & Book
        & A comprehensive overview of human analysis, including detection and tracking, pose estimation, recognition, and applications with human body and face.\\
        \hline

        10 & Human pose estimation and activity recognition from multi-view videos: Comparative explorations of recent developments \citep{holte2012human} & JSTSP
        & A review of model-based 3D HPE and action recognition methods under multi-view. \\
        \hline
        
        11 & A survey of human motion analysis using depth imagery \citep{chen2013survey} & PRL & A survey of traditional RGB-D-based human action recognition methods, including description of sensors, corresponding datasets, and approaches. \\
        \hline

        12 & A survey on model based approaches for 2D and 3D visual human pose recovery \citep{perez2014survey} & Sensors
        & A survey of model-based approaches for HPE, grouped in five main modules: appearance, viewpoint, spatial relations, temporal consistence, and behavior. \\
        \hline

        13 & A survey of human pose estimation: the body parts parsing based methods \citep{liu2015survey} & JVCIR
        & A survey of body parts parsing-based HPE methods under both single-view and multiple-view from different input sources(images, videos, depth). \\
        \hline

        14 & Human pose estimation from monocular images: A comprehensive survey \citep{gong2016human} & Sensors
        & A survey of monocular-based traditional HPE methods with a few deep learning-based methods. \\
        \hline

        15 & 3d human pose estimation: A review of the literature and analysis of covariates \citep{sarafianos20163d} & CVIU
        & A review of 3D HPE methods with different type of inputs(e.g., single image or video, monocular or multi-view). \\
        \hline

        16 & RGB-D-based human motion recognition with deep learning: A survey \citep{wang2018rgb} & CVIU
        & A survey of RGB-D-based motion recognition in four categories: RGB-based, depth-based, skeleton-based, and RGB-D-based. \\
        \hline

        \textbf{17} & \textbf{Monocular Human Pose Estimation: A Survey of Deep Learning-based Methods} & \textbf{Ours}
        & \textbf{A comprehensive survey of deep learning-based monocular HPE research and human pose datasets, organized into four groups: 2D single HPE, 2D multi-HPE, 3D single HPE and 3D multi-HPE} \\
        \hline
    \end{tabular}
\end{table*}

\section{Categories of HPE Methods and Human Body Models}\label{sec2}

\begin{table*}[!htb]
    \centering
    \footnotesize
    
    \caption{The Categories of deep learning-based monocular human pose estimation.}
    \label{tab:overview_all}
    \begin{tabular}{|c|c|c|p{11.3cm}|}
        \hline
        Direction & Sub-direction & Categories & Sub-categories \\ 
        \hline

        \multirow{12}*{\raisebox{-4cm}[0pt]{2D HPE}}
        & \multirow{8}*{\raisebox{-2.5cm}[0pt]{2D Single}}
        & \multirow{3}*{\raisebox{-0.8cm}[0pt]{Regression-based}}
        & \textbf{(1) Direct prediction}: \citep{krizhevsky2012imagenet}, on video~\citep{pfister2014deep} \\
        
        &
        &
        & \textbf{(2) Supervision improvement}: transform heatmaps to joint coordinates~\citep{luvizon2017human, nibali2018numerical}, recursive refinement~\citep{carreira2016human}, bone-based constraint~\citep{sun2017compositional}\\

        &
        &
        & \textbf{(3) Multi-task}: with body part detection~\citep{li2014heterogeneous}, with person detection and action classification~\citep{gkioxari2014r}, with heatmap-based joint detection~\citep{fan2015combining}, with action recognition on video sequences~\citep{luvizon20182d}\\
        \cline{3-4}

        &
        & \multirow{5}*{\raisebox{-1.6cm}[0pt]{Detection-based}}
        & \textbf{(1) Patch-based}: \citep{jain2013learning, chen2014articulated, ramakrishna2014pose} \\

        &
        &
        & \textbf{(2) Network design}: \citep{tompson2015efficient, bulat2016human, xiao2018simple}, multi-scale inputs~\citep{rafi2016efficient}, heatmap-based improvement~\citep{papandreou2017towards}, Hourglass~\citep{newell2016stacked}, CPM~\citep{wei2016convolutional}, PRM~\citep{yang2017learning}, feed forward module~\citep{belagiannis2017recurrent}, HRNet~\citep{sun2019deep}, GAN~\citep{chou2017self, chen2017adversarial, peng2018jointly}  \\
        
        &
        &
        & \textbf{(3) Body structure constraint}: \citep{tompson2014joint, lifshitz2016human, yang2016end, gkioxari2016chained, chu2016structured, chu2017multi, ning2018knowledge, ke2018multi, tang2018deeply, tang2019does} \\
        
        &
        &
        & \textbf{(4) Temporal constraint}: \citep{jain2014modeep, pfister2015flowing, luo2018lstm} \\
        
        &
        &
        & \textbf{(5) Network compression}: \citep{tang2018quantized, debnath2018adapting, Zhang2019Fast} \\
        \cline{2-4}

        & \multirow{4}*{\raisebox{-0.8cm}[0pt]{2D Multiple}}
        & \multirow{1}*{\raisebox{-0.2cm}[0pt]{Top-down}}
        & coarse-to-fine~\citep{iqbal2016multi, huang2017coarse}, bounding box refinement~\citep{fang2017rmpe}, multi-level feature fusion~\citep{xiao2018simple, Chen2018CPN}, results refinement~\citep{moon2019posefix} \\
        \cline{3-4}

        &
        & \multirow{3}*{\raisebox{-0.4cm}[0pt]{Bottom-up}}
        & \textbf{(1) Two-stage}: DeepCut~\citep{pishchulin2016deepcut}, DeeperCut~\citep{insafutdinov2016deepercut}, OpenPose~\citep{cao2016realtime}, PPN~\citep{nie2018pose}, PifPafNet~\citep{kreiss2019pifpaf} \\

        &
        &
        & \textbf{(2) Single-stage}: heatmaps and associative embedding maps~\citep{newell2017associative} \\
        
        &
        &
        & \textbf{(3) Multi-task}: instance segmentation~\citep{papandreou2018personlab}, keypoint detection and semantic segmentation~\citep{kocabas2018multiposenet} \\
        \hline

        \multirow{5}*{\raisebox{-1.8cm}[0pt]{3D HPE}}
        & \multirow{4}*{\raisebox{-1.3cm}[0pt]{3D Single}}
        & \multirow{2}*{\raisebox{-0.3cm}[0pt]{Model-free}}
        & \textbf{(1) Single-stage}: direct prediction~\citep{li20143d, pavlakos2017coarse}, body structure constraint~\citep{li2015maximum, tekin2016structured, sun2017compositional, pavlakos2018ordinal} \\

        &
        &
        & \textbf{(2) 2D-to-3D}: \citep{martinez2017simple, zhou2017towards, tekin2017learning, li2019generating, qammaz2019mocapnet, chen20173d, moreno20173d, wang2018drpose3d, yang20183d} \\
        \cline{3-4}

        &
        & \multirow{2}*{\raisebox{-0.5cm}[0pt]{Model-based}}
        & \textbf{(1) SMPL-based}: \citep{bogo2016keep, tan2017indirect, pavlakos2018learning, omran2018neural, varol2018bodynet, kanazawa2018end, arnab2019exploiting} \\
        
        &
        &
        & \textbf{(2) Kinematic model-based}: \citep{mehta2017monocular, xiaohan2017monocular, zhou2016deep, mehta2017vnect, rhodin2018unsupervised} \\

        &
        &
        & \textbf{(3) Other model-based}: probabilistic model~\citep{tome2017lifting} \\
        \cline{2-4}

        & \multirow{1}*{\raisebox{-0.1cm}[0pt]{3D Multiple}}
        & 
        & bottom-up~\citep{mehta2017single}, top-down~\citep{rogez2017lcr}, SMPL-based~\citep{zanfir2018monocular}, real-time~\citep{mehta2019xnect} \\
        \hline

    \end{tabular}
\end{table*}

\subsection{HPE Method Categories}\label{sec2.1}
This section summarizes the different categories of deep learning-based HPE methods based on different characteristics: 1) generative (human body model-based) and discriminative (human body model-free);
2) top-down (from high-level abstraction to low-level pixel evidence) and bottom-up (from low-level pixel evidence to high-level abstraction);
3) regression-based (directly mapping from input images to body joint positions) and detection-based (generating intermediate image patches or heatmaps of joint locations);
and 4) one-stage (end-to-end training) and multi-stage (stage-by-stage training).

\textbf{Generative V.S. Discriminative:}
The main difference between generative and discriminative methods is whether a method uses human body models or not. Based on the different representations of human body models, generative methods can be processed in different ways such as prior beliefs about the structure of the body model, geometrically projection from different views to 2D or 3D space, high-dimensional parametric space optimization in regression manners. More details of human body model representation can be found in Section \ref{sec2.2}.
Discriminative methods directly learn a mapping from input sources to human pose space (learning-based) or search in existing examples (example-based) without using human body models. Discriminative methods are usually faster than generative methods but may have less robustness for poses never trained with.

\textbf{Top-down V.S. Bottom-up:}
For multi-person pose estimation, HPE methods can generally be classified as top-down and bottom-up methods according to the starting point of the prediction: high-level abstraction or low-level pixel evidence. Top-down methods start from high-level abstraction to first detect persons and generate the person locations in bounding boxes.  Then pose estimation is conducted for each person. In contrast, bottom-up methods first predict all body parts of every person in the input image and then group them either by  human body model fitting or other algorithms. Note that body parts could be joints, limbs, or small template patches depending on different methods.
With an increased number of people in an image, the computation cost of top-down methods  significantly increases, while keeps stable for bottom-up methods. However, if there are some people with a large overlap, bottom-up methods face challenges to group corresponding body parts.

\textbf{Regression-based V.S. Detection-based:}
Based on the different problem formulations, deep learning-based human pose estimation methods can be split into regression-based or detection-based methods. The regression-based methods directly map the input image to the coordinates of body joints or the parameters of human body models. The detection-based methods treat the body parts as detection targets based on two widely used representations: image patches and heatmaps of joint locations.
Direct mapping from images to joint coordinates is very difficult since it is a highly nonlinear problem, while small-region representation provides dense pixel information with stronger robustness. Compared to the original image size, the detected results of small-region representation limit the accuracy of the final joint coordinates.

\textbf{One-stage V.S. Multi-stage:}
The deep learning-based one-stage methods aim to map the input image to human poses by employing end-to-end networks, while multi-stage methods usually predict human pose in multiple stages and are accompanied by intermediate supervision.
For example, some multi-person pose estimation methods first detect the locations of people and then estimate the human pose for each detected person. Other 3D human pose estimation methods first predict joint locations in the 2D surface, then extend them to 3D space.
The training of one-stage methods is easier than multi-stage methods, but with less intermediate constraints.

This survey reviews the recent work in two main sections: 2D human pose estimation (Section 3) and 3D human pose estimation (Section 4). For each section, we further divide them into subsections based on their respective characteristics (see a summary of all the categories and the corresponding papers in Table \ref{tab:overview_all}.)
{\color{blue} }

\subsection{Human Body models}\label{sec2.2}

\begin{figure}
\centering
\includegraphics[width=0.46\textwidth]{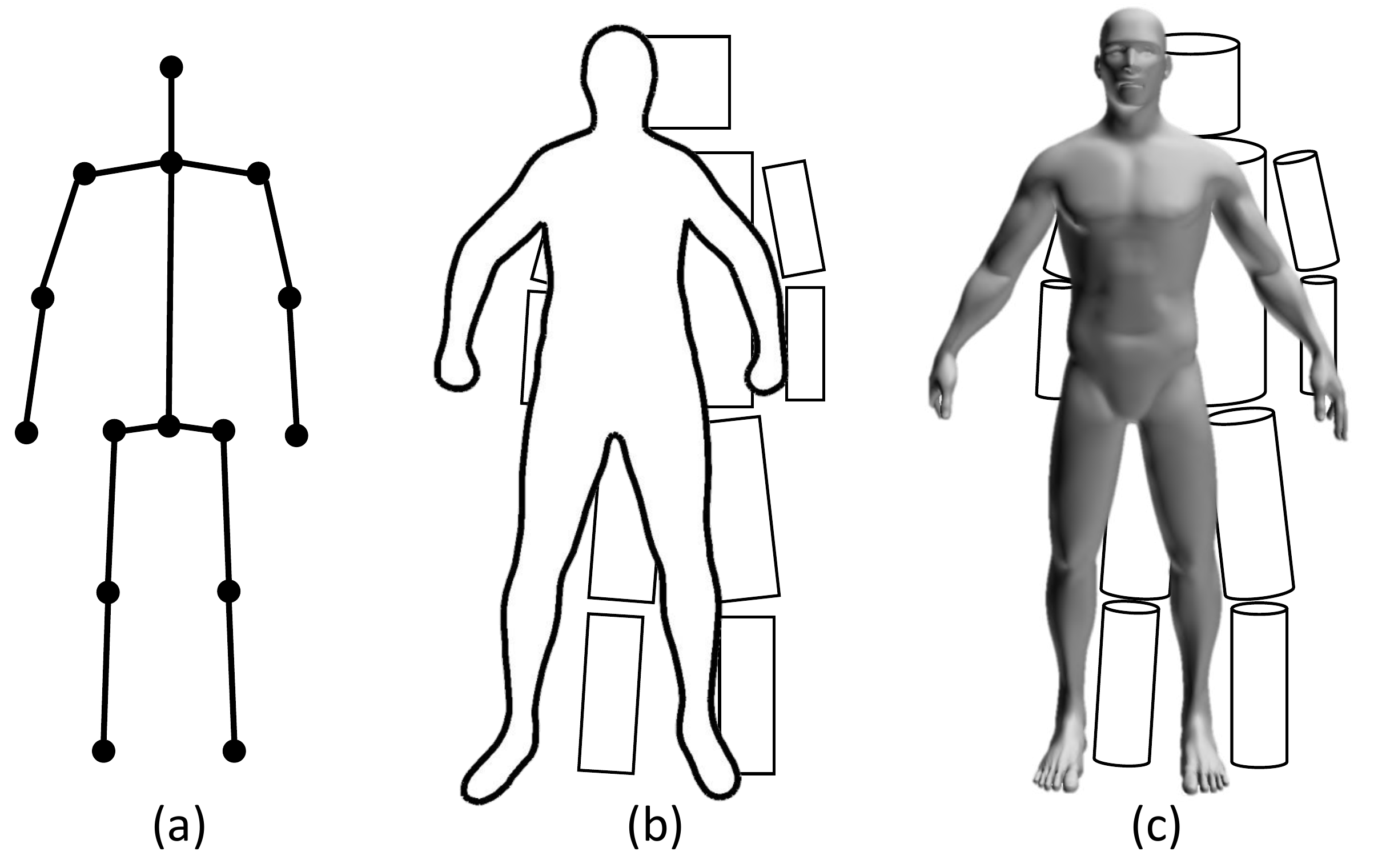}
\caption{Commonly used human body models. (a) skeleton-based model; (b) contour-based models; (c) volume-based models.}
\label{fig:Human_body_models}
\end{figure}

Human body modeling is a key component of HPE. Human body is a flexible and complex non-rigid object and has many specific characteristics like kinematic structure, body shape, surface texture, the position of body parts or body joints, etc. A mature model for human body is not necessary to contain all human body attributes but should satisfy the requirements for specific tasks to build and describe human body pose.
Based on different levels of representations and application scenarios, as shown in Fig. \ref{fig:Human_body_models}, there are three types of commonly used human body models in HPE: skeleton-based model, contour-based model, and volume-based model. For more detailed descriptions of human body models, we refer interested readers to two well-summarized papers \citep{liu2015survey, gong2016human}.

\textbf{Skeleton-based Model:}
Skeleton-based model, also known as stick-figure or kinematic model, represents a set of joint (typically between 10 to 30) locations and the corresponding limb orientations following the human body skeletal structure. The skeleton-based model can also be described as a graph where vertices indicating joints and edges encoding constraints or prior connections of joints within the skeleton structure \citep{felzenszwalb2005pictorial}. This human body topology is very simple and flexible which is widely utilized in both 2D and 3D HPE \citep{cao2016realtime, mehta2017vnect} and human pose datasets \citep{andriluka20142d, wu2017ai}.
With obvious advantages of simple and flexible representing, it also has many shortcomings such as lacking texture information which indicates there is no width and contour information of human body.

\textbf{Contour-based Model:}
The contour-based model is widely used in earlier HPE methods which contains the rough width and contour information of body limbs and torso. Human body parts are approximately represented with rectangles or boundaries of person silhouette. Widely used contour-based models include cardboard models \citep{ju1996cardboard} and Active Shape Models (ASMs) \citep{cootes1995active}.

\textbf{Volume-based Model:}
3D human body shapes and poses are generally represented by volume-based models with geometric shapes or meshes. Earlier geometric shapes for modeling body parts include cylinders, conics, etc. \citep{sidenbladh2000framework}. Modern volume-based models are represented in mesh form, normally captured with 3D scans. Widely used volume-based models includes Shape Completion and Animation of People (SCAPE) \citep{anguelov2005scape}, Skinned Multi-Person Linear model (SMPL) \citep{loper2015smpl}, and a unified deformation model \citep{joo2018total}.

\section{2D Human Pose Estimation}\label{sec3}

2D human pose estimation calculates the locations of human joints from monocular images or videos. Before deep learning brings a huge impact on vision-based human pose estimation, traditional 2D HPE algorithms adopt hand-craft feature extraction and sophisticated body models to obtain local representations and global pose structures \citep{dantone2013human, chen2014articulated, gkioxari2014using}. Here, the recent deep learning-based 2D human pose estimation methods are categorized into "single person pose estimation" and "multi-person pose estimation."

\subsection{2D single person pose estimation}\label{sec3.1}

2D single person pose estimation is to localize body joint positions of a single person in an input image. For images with more persons, pre-processing is needed to crop the original image so that there is only one person in the input image such as using an upper-body detector \citep{eichner2010upperbodydetector} or full-body detector \citep{ren2015faster}, and cropping from original images based on the annotated person center and body scale \citep{andriluka20142d, newell2016stacked}. Early work of introducing deep learning into human pose estimation mainly extended  traditional HPE methods by simply replaced some components of frameworks by neural networks \citep{jain2013learning, ouyang2014multi}.

Based on the different formulations of human pose estimation task, the proposed methods using CNNs can be classified into two categories: regression-based methods and detection-based methods. Regression-based methods attempt to learn a mapping from image to kinematic body joint coordinates by an end-to-end framework and generally directly produce joint coordinates \citep{toshev2014deeppose}. Detection-based methods are intended to predict approximate locations of body parts \citep{chen2014articulated} or joints \citep{newell2016stacked}, usually are supervised by a sequence of rectangular windows (each including a specific body part) \citep{jain2013learning, chen2014articulated} or heatmaps (each indicating one joint position by a 2D Gaussian distribution centered at the joint location) \citep{newell2016stacked, wei2016convolutional}. Each of these two kinds of methods has its advantages and disadvantages. Direct regression learning of only one single point is a difficulty since it is a highly nonlinear problem and lacks robustness, while heatmap learning is supervised by dense pixel information which results in better robustness. Compared to the original image size, heatmap representation has much lower resolution due to the pooling operation in CNNs, which limits the accuracy of joint coordinate estimation. And obtaining joint coordinates from heatmap is normally a non-differentiable process that blocks the network to be trained end-to-end. The recent representative work for 2D single person pose estimation are summarized in Table \ref{tab:2D_HPE}, the last column is the comparisons of PCKh@0.5 scores on the MPII testing set. More details of datasets and evaluation metrics are described in Section~\ref{sec5}.

\begin{table*}[!htb]
\small
    \centering
    \footnotesize
    \caption{Summary of 2D single person pose estimation methods. Note that the last column shows the PCKh@0.5 scores on the Max Planck Institute for Informatics (MPII) Human Pose testing set.}
    \label{tab:2D_HPE}
    \begin{tabular}{|l|c|c|p{9cm}<{\centering}|c|}
        \hline
        Methods & Backbone & Input size & Highlights & PCKh (\%) \\
        \hline
        \multicolumn{5}{|l|}{\textbf{Regression-based}}  \\
        \hline
        \citep{toshev2014deeppose} & \tabincell{c}{AlexNet} & 220$\times$220
        & Direct regression, multi-stage refinement & -\\
        \hline

        \citep{carreira2016human} & GoogleNet & 224$\times$224
        & Iterative error feedback refinement from initial pose. & 81.3\\
        \hline

        \citep{sun2017compositional} & ResNet-50 & 224$\times$224
        & Bone based representation as additional constraint, general for both 2D/3D HPE & 86.4\\
        \hline

        \citep{luvizon2017human} & \tabincell{c}{Inception-v4+\\Hourglass} & 256$\times$256
        & Multi-stage architecture, proposed soft-argmax function to convert heatmaps into joint locations & 91.2\\
        \hline

        \multicolumn{5}{|l|}{\textbf{Detection-based}}  \\
        \hline

        \citep{tompson2014joint} & AlexNet & 320$\times$240
        & Heatmap representation, multi-scale input, MRF-like Spatial-Model & 79.6\\ \hline

        \citep{yang2016end} & VGG & 112$\times$112
        & Jointly learning DCNNs with deformable mixture of parts models  & -\\ \hline

        \citep{newell2016stacked} & Hourglass & 256$\times$256
        & Proposed stacked Hourglass architecture with intermediate supervision.  & 90.9\\ \hline

        \citep{wei2016convolutional} & CPM & 368$\times$368
        & Proposed Convolutional Pose Machines (CPM) with intermediate input and supervision, learn spatial correlations among body parts & 88.5\\ \hline

        \citep{chu2017multi} & Hourglass & 256$\times$256
        & Multi-resolution attention maps from multi-scale features, proposed micro hourglass residual units to increase the receptive field & 91.5\\ \hline

        \citep{yang2017learning} & Hourglass & 256$\times$256
        & Proposed Pyramid Residual Module (PRM) learns filters for input features with different resolutions & 92.0\\ \hline

        \citep{chen2017adversarial} & conv-deconv & 256$\times$256
        & GAN, stacked conv-deconv architecture, multi-task for pose and occlusion, two discriminators for distinguishing whether the pose is 'real' and the confidence is strong & 91.9\\ \hline

        \citep{peng2018jointly} & Hourglass & 256$\times$256
        & GAN, proposed augmentation network to generate data augmentations without looking for more data & 91.5 \\ \hline

        \citep{ke2018multi} & Hourglass & 256$\times$256
        & Improved Hourglass network with multi-scale intermediate supervision, multi-scale feature combination, structure-aware loss and data augmentation of joints masking & 92.1 \\ \hline

        \citep{tang2018deeply} & Hourglass & 256$\times$256
        & Compositional model, hierarchical representation of body parts for intermediate supervision & 92.3\\ \hline

        \citep{sun2019deep} & HRNet & 256$\times$256
        & high-resolution representations of features across the whole network, multi-scale fusion. & 92.3\\ \hline

        \citep{tang2019does} & Hourglass & 256$\times$256
        & data-driven joint grouping, proposed part-based branching network (PBN) to learn representations specific to each part group. & 92.7\\ \hline
    \end{tabular}
\end{table*}

\subsubsection{Regression-based methods}\label{sec3.1.1}

\begin{figure}[!h]
\centering
\includegraphics[width=0.46\textwidth]{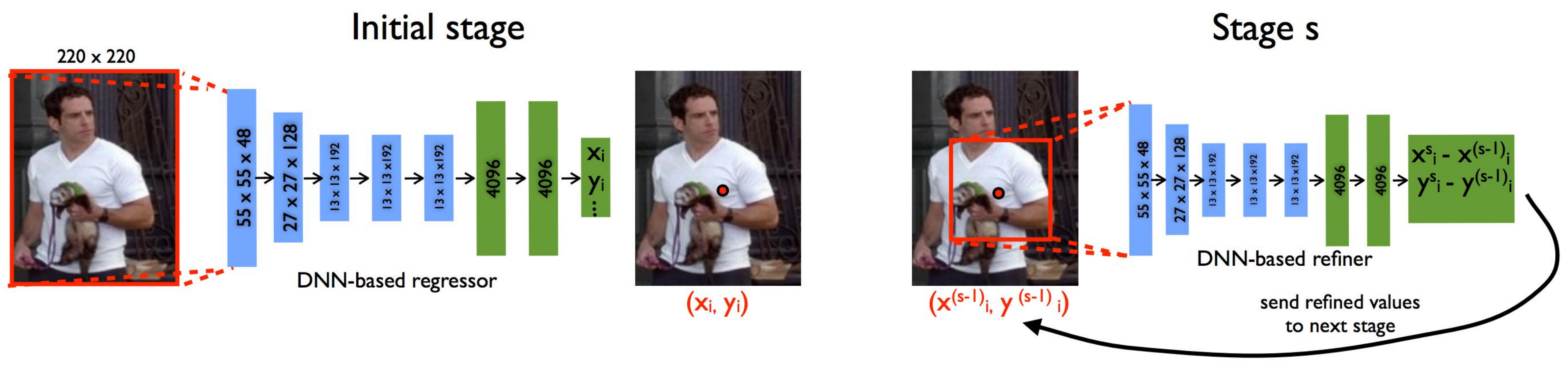}
\caption{The framework of DeepPose \citep{toshev2014deeppose}.}
\label{fig:deeppose}
\end{figure}

AlexNet \citep{krizhevsky2012imagenet} was one of the early networks for deep learning-based HPE methods due to its simple architecture and impressive performance. \citet{toshev2014deeppose} firstly attempted to train an AlexNet-like deep neural network to learn joint coordinates from full images in a very straightforward manner without using any body model or part detectors as shown in Fig.~\ref{fig:deeppose}. Moreover, a cascade architecture of multi-stage refining regressors is employed to refine the cropped images from the previous stage and show improved performance. \citet{pfister2014deep} also applied an AlexNet-like network using a sequence of concatenated frames as input to predict the human pose in the videos.

Only using joints without the surrounding information lacks robustness. Converting heatmap supervision to numerical joint positions supervision can retain the advantages of both representations. \citet{luvizon2017human} proposed a Soft-argmax function to transform heatmaps to joint coordinates which can convert a detection-based network to a differentiable regression-based one. \citet{nibali2018numerical} designed a differentiable spatial to numerical transform (DSNT) layer to calculate joint coordinates from heatmaps, which worked well with low-resolution heatmaps.

Prediction of joint coordinates directly from input images with few constrains is very hard, therefore more powerful networks were introduced with a refinement or body model structure.
\citet{carreira2016human} proposed an Iterative Error Feedback network based on GoogleNet which recursively processes the combination of the input image and output results. The final pose is improved from an initial mean pose after iterations.
\citet{sun2017compositional} proposed a structure-aware regression approach based on a ResNet-50. Instead of using joints to represent pose,  a bone-based representation is designed by involving body structure information to achieve more stable results than only using joint positions. The bone-based representation also works on 3D HPE.

Networks handling multiple closely related tasks of human body may learn diverse features to improve the prediction of joint coordinates.
\citet{li2014heterogeneous} employed an AlexNet-like multi-task framework to handle the joint coordinate prediction task from full images in a regression way, and the body part detection task from image patches obtained by a sliding-window.
\citet{gkioxari2014r} used a R-CNN architecture to synchronously detect person, estimate pose, and classify action.
\citet{fan2015combining} proposed a dual-source deep CNNs which take image patches and full images as inputs and output heatmap represented joint detection results of sliding windows together with coordinate represented joint localization results. The final estimated posture is obtained from the combination of the two results.
\citet{luvizon20182d} designed a network that can jointly handle 2D/3D pose estimation and action recognition from video sequences. The pose estimated in the middle of the network can be used as a reference for action recognition.

\subsubsection{Detection-based methods}\label{sec3.1.2}

Detection-based methods are developed from body part detection methods. In traditional part-based HPE methods, body parts are first detected from image patch candidates and then are assembled to fit a human body model. The detected body parts in early work are relatively big and generally represented by rectangular sliding windows or patches. We refer to~\citep{poppe2007vision, gong2016human} for a more detailed introduction. Some early methods use neural networks as body part detectors to distinguish whether a candidate patch is a specific body part~\citep{jain2013learning}, classify a candidate patch among predefined templates~\citep{chen2014articulated} or predict the confidence map belonging to multiple classes~\citep{ramakrishna2014pose}. Body part detection methods are usually sensitive to complexity background and body occlusions. Therefore the independent image patches with only local appearance may not be sufficiently discriminative for body part detection.

\begin{figure}[!h]
\centering
\includegraphics[width=0.46\textwidth]{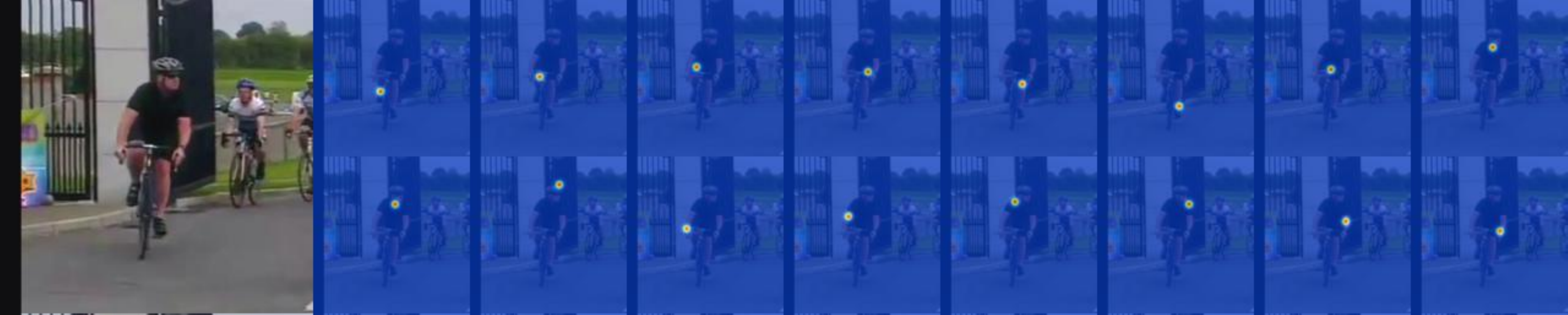}
\caption{Heatmap representation of different joints.}
\label{fig:heatmap}
\end{figure}

In order to provide more supervision information than just joint coordinates and to facilitate the training of CNNs, more recent work employed heatmap to indicate the ground truth of the joint location~\citep{tompson2014joint, jain2014modeep}. As shown in Fig.~\ref{fig:heatmap}, each joint occupies a heatmap channel with a 2D Gaussian distribution centered at the target joint location. Moreover, \citet{papandreou2017towards} proposed an improved representation of the joint location, which is a combination of binary activation heatmap and corresponding offset. Since heatmap representation is more robust than coordinate representation, most of the recent research is based on heatmap representation.

The neural network architecture is very important to make better use of input information.
Some approaches are mainly based on classic networks with appropriate improvements, such as GoogLeNet-based network with multi-scale inputs \citep{rafi2016efficient}, ResNet-based network with deconvolutional layers \citet{xiao2018simple}.
In terms of iterative refinement, some work designed networks in a multi-stage style to refine results from coarse prediction via end-to-end learning \citep{tompson2015efficient, bulat2016human, newell2016stacked, wei2016convolutional, yang2017learning, belagiannis2017recurrent}. Such networks generally use intermediate supervision to address vanishing gradients.
\citet{newell2016stacked} proposed a novel \textit{stacked hourglass} architecture by using a residual module as the component unit.
\citet{wei2016convolutional} proposed a multi-stage prediction framework with input image for each stage.
\citet{yang2017learning} designed a Pyramid Residual Module (PRMs) to replace the residual module of the Hourglass network to enhance the invariance across scales of DCNNs by learning features on various scales.
\citet{belagiannis2017recurrent} combined a 7 layers feedforward module with a recurrent module to iteratively refine the results. This model learns to predict location heatmaps for both joints and body limbs. Also, they analyzed keypoint visibility with unbalanced ground truth distribution.
To keep high-resolution representations of features across the whole network, \citet{sun2019deep} proposed a novel High-Resolution Net (HRNet) with multi-scale feature fusion.

Different from earlier work which attempted to fit detected body parts into body models, some recent work tried to encode human body structure information into networks.
\citet{tompson2014joint} jointly trained a network with a MRF-like spatial-model for learning typical spatial relations between joints.
\citet{lifshitz2016human} discretized an image into log-polar bins centered around each joint and employed a VGG-based network to predict joint category confident for each pair-wise joints (binary terms). With all relative confident scores, the final heatmap for each joint can be generated by a deconvolutional network.
\citet{yang2016end} designed a two-stage network. Stage one is a convolutional neural network to predict joint locations in heatmap representation. Stage two is a message-passing model connected manually according to the human body structure to find optimal joint locations with a max-sum algorithm. 
\citet{gkioxari2016chained} proposed a convolutional Recurrent Neural Network to output joint location one by one following a chain model. The output of each step depends on both the input image and the previously predicted output. The network can handle both images and videos with different connection strategy.
\citet{chu2016structured} proposed to transform kernels by a bi-directional tree to pass information between corresponding joints in a tree body model.
\citet{chu2017multi} replaced the residual modules of the Hourglass network with more sophisticated ones. The Conditional Random Field (CRF) is utilized for attention maps as intermediate supervisions for learning body structure information.
\citep{ning2018knowledge} designed a fractal network to impose body prior knowledge to guide the network. The external knowledge visual features are encoded into the basic network by using a learned projection matrix.
\citet{ke2018multi} proposed a multi-scale structure-aware network based on Hourglass network  with multi-scale supervision, multi-scale feature combination, structure-aware loss, and data augmentation of joints masking.
On the basic framework of Hourglass network, \citet{tang2018deeply} designed a hierarchical representation of body parts for intermediate supervision to replace heatmap for each joint. Thus the network  learns the bottom-up/top-down body structure, rather than only scattered joints.
\citet{tang2019does} proposed a part-based branching network (PBN) to learn specific representations of each part group rather than predict all joint heatmaps from one branch. The  data-driven part groups are then split by calculating mutual information of joints.

Generative Adversarial Networks (GANs) are also employed to provide adversarial supervision for learning body structure or network training.
\citet{chou2017self} introduced adversarial learning with two same Hourglass networks as generator and discriminator respectively. The generator predicts heatmap location of each joint, while the discriminator distinguishes ground truth heatmaps from generated heatmaps.
\citet{chen2017adversarial} proposed a structure-aware convolutional network with one generator and two discriminators to incorporate priors of human body structure. The generator is designed from the Hourglass network to predict joint heatmaps as well as occlusion heatmaps. The pose discriminator can discriminate against reasonable body configuration from unreasonable body configuration. The confidence discriminator shows the confidence score of predictions.
\citet{peng2018jointly} studied how to jointly optimize data augmentation and network training without looking for more data. Instead of using random data augmentation, they applied augmentations to increase the network loss while the pose network learns from the generated augmentations.

Utilization of temporal information is also very important to estimate 2D human poses in monocular video sequences. \citet{jain2014modeep} designed a framework contains two-branch CNNs taking multi-scale RGB frames and optical-flow maps as inputs. The extracted features are concatenated before the last convolutional layers. \citet{pfister2015flowing} used optical-flow maps as a guide to align predicted heatmaps from neighboring frames based on the temporal context of videos. \citet{luo2018lstm} exploited temporal information with a Recurrent Neural Network redesigned from CPM by changing multi-stage architecture with LSTM structure.

In order to estimate human poses on low-capacity devices, network parameters can be reduced while still maintaining competitive performance. \citet{tang2018quantized} committed to improving the network structure by proposing a densely connected U-Nets and efficient usage of memory. This network is similar to the idea of the Hourglass network while utilizing U-Net as each component with a more optimized global connection across each stage resulting in fewer parameters and small model size. \citet{debnath2018adapting} adapted MobileNets~\citep{howard2017mobilenets} for pose estimation by designing a split stream architecture at the final two layers of the MobileNets. \citet{Zhang2019Fast} designed a lightweight variant of Hourglass network and trained it with a full teacher Hourglass network by a Fast Pose Distillation (FPD) training strategy.

In summary, the heatmap representation is more suitable for network training than coordinate representation from detection-based methods in deep learning-based 2D single person pose estimation.

\subsection{2D multi-person pose estimation}\label{sec3.2}

Different from single person pose estimation, multi-person pose estimation needs to handle both detection and localization tasks since there is no prompt of how many persons in the input images. According to from which level (high-level abstraction or low-level pixel evidence) to start the calculation, human pose estimation methods can be classified into top-down methods and bottom-up methods.

Top-down methods generally employ person detectors to obtain a set of the bounding box of people in the input image and then directly leverage existing single-person pose estimators to predict human poses. The predicted poses heavily depend on the precision of the person detection. The runtime for the whole system is proportional based on the number of persons. While bottom-up methods directly predict all the 2D joints of all persons and then assemble them into independent skeletons. Correct grouping of joint points in a complex environment is a challenging research task. Table \ref{tab:2D_M_HPE} summarizes recent deep learning-based work about 2D multi-person pose estimation methods in both top-down and bottom-up categories. The last column of Table \ref{tab:2D_M_HPE} is the Average Precision (AP) scores on the COCO test-dev dataset. More details of datasets and evaluation metrics are described in Section~\ref{sec5}.

\begin{table*}[!htb]
\small
    \centering
    \footnotesize
    \caption{Comparison of 2D multi-person pose estimation methods. Note that the last column shows the Average Precision (AP) scores on the COCO test-dev set. The results with * were obtained with COCO16 training set, while others with COCO17 training set.}
    \label{tab:2D_M_HPE}
    \begin{tabular}{|l|p{1.7cm}|p{10.5cm}<{\centering}|c|}
        \hline
        Methods & Network type & Highlights & AP Score (\%)\\
        \hline
        \multicolumn{4}{|l|}{\textbf{Top-down}}  \\
        \hline
        \citep{iqbal2016multi} & Faster R-CNN + CPM 
        & After person detection and single HPE, refines detected local joint candidates with Integer Linear Programming (ILP). & -\\ \hline
        
        \citep{fang2017rmpe}  & Faster R-CNN + Hourglass 
        & Combines symmetric spatial transformer network (SSTN) and Hourglass model to do SPPE on detected results; proposes a parametric pose NMS for refining pose proposals; designs a pose-guided proposals generator to augment the existing training samples & $63.3^{*}$\\ \hline
        
        \citep{papandreou2017towards}  & Faster R-CNN + ResNet-101
        & Produces heatmap and offset map of each joint for SPPE and combines them with an aggregation procedure; uses keypoint-based NMS to avoid duplicate poses & $64.9^{*}$\\ \hline
        
        \citep{huang2017coarse}  & Faster R-CNN + Inception-v2
        & Produces coarse and fine poses for SPPE with multi-level supervisions; multi-scale features fusion & $72.2^{*}$\\ \hline

        \citep{he2017mask}  & Mask R-CNN + ResNet-FPN
        & An extension of Mask R-CNN framework; predicts keypoints and human mask synchronously & $63.1^{*}$\\ \hline

        \citep{xiao2018simple}  & Faster R-CNN + ResNet
        & Simply adds a few deconvolutional layers after ResNet to generate heatmaps from deep and low resolution features & 73.7\\ \hline

        \citep{Chen2018CPN}  & FPN + CPN
        & Proposes CPN with feature pyramid; two-stage network; online hard keypoints mining & 73.0\\ \hline

        \citep{moon2019posefix} & ResNet + upsampling
        & proposes PoseFix net to refine estimated pose from any HPE methods based on pose error distributions & - \\ \hline

        \citep{sun2019deep}  & Faster R-CNN + HRNet
        & high-resolution representations of features across the whole network, multi-scale fusion & 75.5 \\ \hline


        \multicolumn{4}{|l|}{\textbf{Bottom-up}}  \\
        \hline
        \citep{pishchulin2016deepcut}  & Fast R-CNN
        & Formulate the distinguishing different persons as an ILP problem; cluster detected part candidates; combine person clusters and labeled parts to obtain final poses & -\\ \hline

        \citep{insafutdinov2016deepercut}  & ResNet
        & Employs image-conditioned pairwise terms to assemble the part proposals & -\\ \hline

        \citep{cao2016realtime}  & VGG-19 + CPM
        & OpenPose; real-time; Simultaneous joints detection and association in a two-branch architecture; propose Part Affinity Fields (PAFs) to encode the location and orientation of limbs & $61.8^{*}$\\ \hline

        \citep{newell2017associative}  & Hourglass
        & Simultaneous joints detection and association in one branch; propose dense associative embedding tags for detected joints grouping & 65.5\\ \hline

        \citep{nie2018pose}  & Hourglass
        & Simultaneous joints detection and association in a two-branch architecture; generate partitions in the embedding space parameterized by person centroids over joint candidates; estimate pose instances by a local greedy inference approach & -\\ \hline

        \citep{papandreou2018personlab}  & ResNet
        & Multi-task (pose estimation and instance segmentation) network; simultaneous joints detection and association in a multi-branch architecture; multi-range joint offsets following tree-structured kinematic graph to guide joints grouping & 68.7\\ \hline

        \citep{kocabas2018multiposenet}  & ResNet-FPN + RetinaNet
        & Multi-task (pose estimation, person detection and person segmentation) network; simultaneous keypoint detection and person detection in a two-branch architecture; proposes a Pose Residual Network (PRN) to assign keypoint detection to person instances & 69.6\\ \hline

        \citep{kreiss2019pifpaf} & ResNet-50
        & predicts Part Intensity Fields (PIF) and Part Association Fields (PAF) to represent body joints location and body joints association; works well under low-resolution & 66.7 \\ \hline

    \end{tabular}
\end{table*}

\subsubsection{Top-down methods}\label{sec3.2.1}

The two most important components of top-down HPE methods are human body region candidate detector and a single person pose estimator. Most of the research focused on human part estimation based on existing human detectors such as Faster R-CNN~\citep{ren2015faster}, Mask R-CNN~\citep{he2017mask}, FPN~\citep{lin2017feature}.
\citet{iqbal2016multi} utilized a convolutional pose machine-based pose estimator to generate initial poses. Then integer linear programming (ILP) is applied to obtain the final poses. \citet{fang2017rmpe} adopted spatial transformer network (STN) \citep{jaderberg2015spatial}, Non-Maximum-Suppression (NMS), and Hourglass network to facilitate pose estimation in the presence of inaccurate human bounding boxes.
\citet{huang2017coarse} developed a coarse-fine network (CFN) with Inception-v2 network \citep{szegedy2016rethinking} as the backbone. The network is supervised in multiple levels for learning coarse and fine prediction.
\citet{xiao2018simple} added several deconvolutional layers over the last convolution layer of ResNet to generate heatmaps from deep and low-resolution features.
\citet{Chen2018CPN} proposed a cascade pyramid network (CPN) by employing multi-scale feature maps from different layers to obtain more inference from local and global features with an online hard keypoint mining loss for difficulty joints.
Based on similar pose error distributions of different HPE approaches, \citet{moon2019posefix} designed PoseFix net to refine estimated poses from any methods.

Top-down HPE methods can be easily implemented by combining existing detection networks and single HPE networks. Meanwhile, the performance of this kind of methods is affected by person detection results and the operation speed is usually not real-time.

\subsubsection{Bottom-up methods}\label{sec3.2.2}

The main components of bottom-up HPE methods include body joint detection and joint candidate grouping. Most algorithms handle these two components separately.
DeepCut~\citep{pishchulin2016deepcut} employed a Fast R-CNN based body part detector to first detect all the body part candidates, then labeled each part to its corresponding part category, and assembled these parts with integer linear programming to a complete skeleton.
DeeperCut~\citep{insafutdinov2016deepercut} improved the DeepCut by using a stronger part detector based on ResNet and a better incremental optimization strategy exploring geometric and appearance constraints among joint candidates.
OpenPose~\citep{cao2016realtime} used CPM to predict candidates of all body joints  with Part Affinity Fields (PAFs). The proposed PAFs can encode locations and orientations of limbs to assemble the estimated joints into different poses of persons.
\citet{nie2018pose} proposed a Pose Partition Network (PPN) to conduct both joint detection and dense regression for joint partition. Then PPN performs local inference for joint configurations with joint partition.
Similar to OpenPose, \citet{kreiss2019pifpaf} designed a PifPaf net to predict a Part Intensity Field (PIF) and a Part Association Field (PAF) to represent body joint locations and body joint association. It works well on low-resolution images due to the fine-grained PAF and the utilization of Laplace loss.

The above methods are all following a separation of joint detection and joint grouping. Recently, some methods can do the prediction in one stage.
\citet{newell2017associative} introduced a single-stage deep network architecture to simultaneously perform both detection and grouping. This network can produce detection heatmaps for each joint, and associative embedding maps that contain the grouping tags of each joint.

Some methods employed multi-task structures.
\citet{papandreou2018personlab} proposed a box-free multi-task network for pose estimation and instance segmentation. The ResNet-based network can synchronously predict joint heatmaps of all keypoints for every person and their relative displacements. Then the grouping starts from the most confident detection with a greedy decoding process based on a tree-structured kinematic graph.
The network proposed by \citet{kocabas2018multiposenet} combines a multi-task model with a novel assignment method to handle human keypoint estimation, detection, and semantic segmentation tasks altogether. Its backbone network is a combination of ResNet and FPN with shared features for keypoints and person detection subnets. The human detection results are used as constraints of the spatial position of people.

Currently, the processing speed of bottom-up methods is very fast, and some \citep{cao2016realtime, nie2018pose} can run in real-time. However, the performance can be very influenced by the complex background and human occlusions. The top-down approaches achieved state-of-the-art performance in almost all benchmark datasets while the processing speed is limited by the number of detected people.

\section{3D Human Pose Estimation}\label{sec4}

3D human pose estimation is to predict locations of body joints in 3D space from images or other input sources. Although commercial products such as Kinect~\citep{Kinect} with depth sensor, VICON~\citep{Vicon} with optical sensor and TheCaptury~\citep{TheCaptury} with multiple cameras have been employed for 3D body pose estimation, all these systems work in very constrained environments or need special markers on human body. Monocular camera, as the most widely used sensor, is very important for 3D human pose estimation. Deep neural networks have the capability to estimate the dense depth~\citep{li2015depth, li2018monocular, li2019learning} and sparse depth points (joints) as well from monocular images. Moreover, the progress of 3D human pose estimation from monocular inputs can further improve multi-view 3D human pose estimation in constrained environments. Thus, this section focuses on the deep learning-based methods that estimate 3D human pose from monocular RGB images and videos including 3D single person pose estimation and 3D multi-person pose estimation.

\subsection{3D single person pose estimation}\label{sec4.1}

Compared to 2D HPE, 3D HPE is more challenging since it needs to predict the depth information of body joints. In addition, the training data for 3D HPE are not easy to obtain as 2D HPE. Most existing datasets are obtained under constrained environments with limited generalizability. For single person pose estimation, the bounding box of the person in the image is normally provided, and hence it is not necessary to combine the process of person detection. In this section, we divide the methods of 3D single person pose estimation into model-free and model-based categories and summarize the recent work in Table~\ref{tab:3D_S_HPE}. The last column of Table~\ref{tab:3D_S_HPE} is the comparisons of Mean Per Joint Position Error (MPJPE) in millimeter on Human3.6M dataset under protocol \#1. More details of datasets and evaluation metrics are described in Section~\ref{sec5}.

\begin{table*}[!htb]
\small
    \centering
    \footnotesize
    \caption{Comparison of 3D single person pose estimation methods. Here “E.” stands for “Extra data” and “T.” indicates “Temporal info”. The last column is the Mean Per Joint Position Error (MPJPE) in millimeter on Human3.6M dataset under protocol \#1. The results with $*$ were reported from 6 actions in testing set, while others from all 17 actions. The results with $\dagger$ were reported with 2D joint ground truth. The methods with $\#$ report joint rotation as well.}
    \label{tab:3D_S_HPE}
    \begin{tabular}{|l|p{1.7cm}<{\centering}|c|c|p{9.5cm}<{\centering}|p{0.8cm}<{\centering}|}
        \hline
        Methods & Backbone & E. & T. & Highlights & MPJPE (mm)\\
        \hline
        \multicolumn{6}{|l|}{\textbf{Model-free}}  \\
        \hline

        \citep{li20143d} & shallow CNNs & \xmark & \xmark & A multi-task network to predict of body part detection with sliding windows and 3D pose estimation jointly  & $132.2^{*}$\\ \hline

        \citep{li2015maximum} & shallow CNNs & \xmark & \xmark & Compute matching score of image-pose pairs & $120.2^{*}$\\ \hline

        \citep{tekin2016structured} & auto-encoder+ shallow CNNs & \xmark & \xmark & Employ an auto-encoder to learn a high-dimensional representation of 3D pose; use a shallow CNNs network to learn the high-dimensional pose representation & $116.8^{*}$\\ \hline

        \citep{tekin2017learning} & Hourglass & \cmark & \xmark & Predict 2D heatmaps for joints first; then use a trainable fusion architecture to combine 2D heatmaps and extracted features; 2D module is pre-trained with MPII & $69.7$\\ \hline
		
        \citep{chen20173d} & CPM & \cmark & \xmark & Estimate 2D poses from images first; then estimate depth of them by matching to a library of 3D poses; 2D module is pre-trained with MPII & $82.7$ $/57.5^{\dagger}$\\ \hline

        \citep{moreno20173d} & CPM & \cmark & \xmark & Use Euclidean Distance Matrices (EDMs) to encoding pairwise distances of 2D and 3D body joints; train a network to learn 2D-to-3D EDM regression; jointly trained with other 3D (Humaneva-I) dataset & $87.3$\\ \hline

        \citep{pavlakos2017coarse} & Hourglass & \cmark & \xmark & Volumetric representation for 3D human pose; a coarse-to-fine prediction scheme; 2D module is pre-trained with MPII & $71.9$\\ \hline

        \citep{zhou2017towards} & Hourglass & \cmark & \xmark & A proposed loss induced from a geometric constraint for 2D data; bone-length constraints; jointly trained with 2D (MPII) dataset & $64.9$\\ \hline

        \citep{martinez2017simple} & Hourglass & \cmark & \xmark & Directly map predicted 2D poses to 3D poses with two linear layers; 2D module is pre-trained with MPII; process in real-time & $62.9$ $/45.5^{\dagger}$\\ \hline

        \citep{sun2017compositional}$^{\#}$ & ResNet & \cmark & \xmark & A bone-based representation involving body structure information to enhance robustness; bone-length constraints; jointly trained with 2D (MPII) dataset & $48.3$\\ \hline

        \citep{yang20183d} & Hourglass & \cmark & \xmark & Adversarial learning for domain adaptation of 2D/3D datasets; adopted generator from \citep{zhou2017towards}; multi-source discriminator with image, pairwise geometric structure and joint location; jointly trained with 2D (MPII) dataset & $58.6$\\ \hline

        \citet{pavlakos2018ordinal} & Hourglass & \cmark & \xmark & Volumetric representation for 3D human pose; additional ordinal depths annotations for human joints; jointly trained with 2D (MPII) and 3D (Humaneva-I) datasets & $56.2$\\ \hline

        \citep{sun2018integral} & Mask R-CNN & \cmark & \xmark & Volumetric representation for 3D human pose; integral operation unifies the heat map representation and joint regression; jointly trained with 2D (MPII) dataset & $40.6$\\ \hline
        
        \citep{li2019generating} & Hourglass & \cmark & \xmark & Multiple hypotheses of 3D poses are generated from 2D poses; the best one is chosen by 2D reprojections; 2D module is pre-trained with MPII &  $52.7$\\ \hline 

        \multicolumn{6}{|l|}{\textbf{Model-based}}  \\
        \hline

        \citep{bogo2016keep}$^{\#}$ & DeepCut & \xmark & \xmark & SMPL model; fit SMPL model to 2D joints by minimizing the distance between 2D joints and projected 3D model joints & $82.3$\\ \hline

        \citep{zhou2016deep}$^{\#}$ & ResNet & \xmark & \xmark & kinematic model; embedded a kinematic object model into network for general articulated object pose estimation; orientation and rotational constrains & $107.3$\\ \hline
        
        \citep{mehta2017vnect}$^{\#}$ & ResNet & \cmark & \cmark & A real-time pipeline with temporal smooth filter and model-based kinematic skeleton fitting; 2D module is pre-trained with MPII and LSP; process in real-time; provide body height & $80.5$\\ \hline

        \citep{tan2017indirect} & shallow CNNs & \xmark & \xmark & SMPL model; first train a decoder to predict a 2D body silhouette from parameters of SMPL; then train a encoder-decoder network with images and corresponding silhouettes; the trained encoder can predict parameters of SMPL from images & -\\ \hline

        \citep{mehta2017monocular} & Resnet & \cmark & \xmark & Kinematic model; transfer learning from features learned for 2D pose estimation; 2D pose prediction as auxiliary task; predict relative joint locations following the kinematic tree body model; jointly trained with 2D (MPII and LSP) datasets & $74.1$\\ \hline

        \citep{xiaohan2017monocular} & RMPE + LSTM & \cmark & \xmark & Kinematic model; joint depth estimation from global 2D pose with skeleton-LSTM and local body parts with patch-LSTM; 2D module is pre-trained with MPII & $79.5$\\ \hline

        \citep{kanazawa2018end}$^{\#}$ & ResNet & \cmark & \xmark & SMPL model; adversarial learning for domain adaptation of 2D images and 3D human body model; propose a framework to learn parameters of SMPL; jointly trained with 2D (LSP, MPII and COCO) datasets; process in real-time & $88.0$\\ \hline

        \citep{pavlakos2018learning}$^{\#}$ & Hourglass & \cmark & \xmark & SMPL model; first predict 2D heatmaps of joint and human silhouette; second generate parameters of SMPL; 2D module is trained with MPII and LSP & $75.9$\\ \hline

        \citep{omran2018neural}$^{\#}$ & RefineNet & \xmark & \xmark & SMPL model; first predict 2D body parts segmentation from the RGB image; second take this segmentation to predict the parameters of SMPL & $59.9$\\ \hline

        \citep{varol2018bodynet} & Hourglass & \cmark & \xmark & SMPL model; first predict 2D pose and 2D body parts segmentation; second predict 3D pose; finally predict volumetric shape to fit SMPL model; 2D modules are trained with MPII and SURREAL & $49.0$\\ \hline
        
        \citep{arnab2019exploiting}$^{\#}$ & ResNet & \cmark & \cmark & SMPL model; 2D keypoints, SMPL and camera parameters estimation; off-line bundle adjustment with temporal constraints; 2D module is trained with COCO & $77.8$ $/63.3^{\dagger}$\\ \hline
        
        \citep{tome2017lifting} & CPM & \cmark & \xmark & Pre-trained probabilistic 3D pose model; 3D lifting and projection by probabilistic model within the CPM-like network; 2D module is pre-trained with MPII; process in real-time & $88.4$\\ \hline

        \citep{rhodin2018unsupervised} & Hourglass & \xmark & \xmark & A latent variable body model learned from multi-view images; an encoder-decoder to predict a novel view image from a given one; the pre-trained encoder with additional shallow layers to predict 3D poses from images & -\\ \hline

    \end{tabular}
\end{table*}

\subsubsection{Model-free methods}\label{sec4.1.1}

The model-free methods do not employ human body models as the predicted target or intermediate cues. They can be roughly categorized into two types: 1) directly map an image to 3D pose, and 2) estimate depth following intermediately predicted 2D pose from 2D pose estimation methods.

Approaches that directly estimate the 3D pose from image features usually contain very few constraints.
\citet{li20143d} employed a shallow network to regress 3D joint coordinates directly with synchronous task of body part detection with sliding windows.
\citet{pavlakos2017coarse} proposed a volumetric representation for 3D human pose and employed a coarse-to-fine prediction scheme to refine predictions with a multi-stage structure. 
Some researchers attempted to add body structure information or the dependencies between human joints to the deep learning networks.
\citet{li2015maximum} designed an embedding sub-network learning latent pose structure information to guide the 3D joint coordinates mapping. The sub-network can assign matching scores for input image-pose pairs with a maximum-margin cost function.
\citet{tekin2016structured} pre-trained an unsupervised auto-encoder to learn a high-dimensional latent pose representation of 3D pose for adding implicit constraints about the human body and then used a shallow network to learn the high-dimensional pose representation.
\citet{sun2017compositional} proposed a structure-aware regression approach. They designed a bone-based representation involving body structure information which is more stable than only using joint positions.
\citet{pavlakos2018ordinal} trained the network with additional ordinal depths of human joints as constraints, by which the 2D human datasets can also be feed in with ordinal depths annotations.

The 3D HPE methods which intermediately estimate 2D poses gain the advantages of 2D HPE, and can easily utilize images from 2D human datasets. Some of them adopt off-the-shelf 2D HPE modules to first estimate 2D poses, then extend to 3D poses. 
\citep{martinez2017simple} designed a 2D-to-3D pose predictor with only two linear layers.
\citep{zhou2017towards} presented a depth regression module to predict 3D pose from 2D heatmaps with a proposed geometric constraint loss for 2D data.
\citet{tekin2017learning} proposed a two-branch framework to predict 2D heatmaps and extract features from images. The extracted features are fused with 2D heatmaps by a trainable fusion scheme instead of being hand-crafted to obtain the final 3D joint coordinates.
\citet{li2019generating} considered 3D HPE as an inverse problem with multiple feasible solutions. Multiple feasible hypotheses of 3D poses are generated from 2D poses and the best one is chosen by 2D reprojections.
\citet{qammaz2019mocapnet} proposed MocapNET directly encoding 2D poses  into the 3D BVH \citep{meredith2001motion} format for subsequent rendering. By consolidating OpenPose~\citep{cao2016realtime} the architecture estimated and rendered 3D human pose in real-time using only CPU processing.

When mapping 2D pose to 3D pose, different strategies may be applied. \citet{chen20173d} used a matching strategy for an estimated 2D pose and 3D pose from a library. \citet{moreno20173d} encoded pairwise distances of 2D and 3D body joints into two Euclidean Distance Matrices (EDMs) and trained a regression network to learn the mapping of the two matrices. \citet{wang2018drpose3d} predicted depth rankings of human joints as a cue to infer 3D joint positions from a 2D pose. \citet{yang20183d} adopted a generator from \citep{zhou2017towards} and designed a multi-source discriminator with image, pairwise geometric structure, and joint location information.

\subsubsection{Model-based methods}\label{sec4.1.2}

Model-based methods generally employ a parametric body model or template to estimate human pose and shape from images. Early geometric-based models are not included in this paper. More recent models are estimated from multiple scans of diverse people \citep{hasler2009statistical, loper2015smpl, pons2015dyna, zuffi2015stitched} or combination of different body models \citep{joo2018total}. These models are typically parameterized by separate body pose and shape components.

Some work employed the body model of SMPL \citep{loper2015smpl} and attempted to estimate the 3D parameters from images. For example,
\citet{bogo2016keep} fit SMPL model to estimated 2D joints and proposed an optimization-based method to recover SMPL parameters from 2D joints.
\citet{tan2017indirect} inferred SMPL parameters by first training a decoder to predict silhouettes from SMPL parameters with synthetic data, and then learning an image encoder with the trained decoder. The trained encoder can predict SMPL parameters from input images.
Directly learning parameters of SMPL is hard, some work predicted intermediate cues as constrains. For example, intermediate 2D pose and human body segmentation \citep{pavlakos2018learning}, body parts segmentation \citep{omran2018neural}, 2D pose and body parts segmentation \citep{varol2018bodynet}.
In order to overcome the problem of lacking training data for the human body model, \citep{kanazawa2018end} employed adversarial learning by using a generator to predict parameters of SMPL, and a discriminator to distinguish the real SMPL model and the predicted ones.
\citep{arnab2019exploiting} reconstructed person from video sequences which explored the multiple views information.

Kinematic model is widely used for 3D HPE. \citep{mehta2017monocular} predicted relative joint locations from 2D heatmaps following the kinematic tree body model.
\citep{xiaohan2017monocular} employed LSTM to exploit global 2D joint locations and
local body part images following kinematic tree body model which are two cues for joint depth estimation.
\citet{zhou2016deep} embedded a kinematic object model into a network for general articulated object pose estimation which provides orientation and rotational constrains.
\citet{mehta2017vnect} proposed a pipeline for 3D single HPE running in real-time. The temporal information and kinematic body model are used as a smooth filter and skeleton fitting respectively.
\citet{rhodin2018unsupervised} used an encoder-decoder network to learn a latent variable body model without 2D or 3D annotations under self-supervision, then employed the pre-trained encoder to predict 3D poses.

Additional to those typical body models,  latent 3D pose model learned from data is also used for 3D HPE.
\citet{tome2017lifting} proposed a multi-stage CPM-like network including a pre-trained probabilistic 3D pose model layer which can generate 3D pose from 2D heatmaps.

\begin{table*}[t]
\small
    \centering
    \footnotesize
    \caption{Summary of 3D multi-person pose estimation methods.}
    \label{tab:3D_M_HPE}
    \begin{tabular}{|l|p{1.9cm}|p{11.5cm}<{\centering}|}
        \hline
        Methods & Network type & Highlights \\
        \hline
        \citep{mehta2017single}  & ResNet
        & Propose an occlusion-robust pose-maps (ORPM) for full-body pose inference even under (self-)occlusions; combine 2D pose and part affinity fields to infer person instances\\ \hline

        \citep{rogez2017lcr} & \tabincell{c}{Faster R-CNN\\+ VGG-16}
        & Localize human bounding boxes with Faster R-CNN; classify the closest anchor-pose for each proposal; regress anchor-pose to get final pose \\ \hline

        \citep{zanfir2018monocular}  & DMHS 
        & Feed forward process of body parts semantic segmentation and 3d pose estimates; feed backward process of refining pose and shape parameters of a body model SMPL \\ \hline
        
        \citet{mehta2019xnect}  & SelecSLS Net
        & Real-time; a new CNN architecture that uses selective long and short range skip connections; 2D and 3D pose features prediction along with identity assignments for all visible joints of all individuals; complete 3D pose reconstruction including occluded joints; temporal stability refinement and kinematic skeleton fitting. \\ \hline

    \end{tabular}
\end{table*}

\subsection{3D multi-person pose estimation}\label{sec4.2}

The achievements of monocular 3D multi-person pose estimation are based on 3D single person pose estimation and other deep learning methods. This research field is pretty new and only a few methods are proposed. Table \ref{tab:3D_M_HPE} summarizes these methods.

\citet{mehta2017single} proposed a bottom-up method by using 2D pose and part affinity fields to infer person instances. An occlusion-robust pose-maps (ORPM) is proposed to provide multi-style occlusion information regardless of the number of people. \citet{rogez2017lcr} proposed a Localization-Classification-Regression Network (LCR-Net) following three-stage processing. First, Faster R-CNN is employed to detect people locations. Second, each pose proposal is assigned with the closest anchor-pose scored by a classifier. The final poses are refined with a regressor respectively.
\citet{zanfir2018monocular} proposed a framework with feed forward and feed backward stages for 3D multi-person pose and shape estimation. The feed forward process includes semantic segmentation of body parts and 3D pose estimates based on DMHS \citep{popa2017deep}. Then the feed backward process refines the pose and shape parameters of SMPL \citep{loper2015smpl}. \citet{mehta2019xnect} estimated multiple poses in real-time with three stages. First,  SelecSLS Net infers 2D pose and intermediate 3D pose encoding for visible body joints. Then based on each detected person, it reconstructs the complete 3D pose, including occluded joints. Finally, refinement is provided for temporal stability and kinematic skeleton fitting.

\section{Datasets and evaluation protocols}\label{sec5}

Datasets play an important role in deep learning-baed human pose estimation. Datasets not only are essential for fair comparison of different algorithms but also bring more challenges and complexity through their expansion and improvement. With the maturity of the commercial motion capture systems and crowdsourcing services, recent datasets are no longer limited by the data quantity or lab environments.

This section discusses the popular publicly available human pose datasets for 2D and 3D human pose estimation, introduces the characteristics and the evaluation methods, as well as the performance of recent state-of-the-art work on several popular datasets. In addition to these basic datasets, some researchers have extended the existing datasets in their own way~\citep{pavlakos2018ordinal, lassner2017unite}. In addition, some relevant human datasets are also within the scope of this section~\citep{guler2018densepose}. A brief description of how researchers collected all the annotated images of each dataset is also provided to bring inspiration to readers who want to generate their own datasets.

\subsection{Datasets for 2D human pose estimation}\label{sec5.1}

Before deep learning brings significant progress for 2D HPE, there are many 2D human pose datasets for specific scenarios and tasks. Upper body pose datasets include Buffy Stickmen \citep{ferrari2008progressive} (frontal-facing view, from indoor TV show), ETHZ PASCAL Stickmen \citep{eichner2009better} (frontal-facing view, from PASCAL VOC \citep{everingham2010pascal}), We Are Family \citep{eichner2010we} (Group photo scenario), Video Pose 2 \citep{sapp2011parsing} (from indoor TV show), Sync. Activities \citep{eichner2012human} (sports, full-body image, upper body annotation). full-body pose datasets include PASCAL Person Layout \citep{everingham2010pascal} (daily scene, from PASCAL VOC \citep{everingham2010pascal}), Sport \citep{wang2011learning} (sport scenes) and UIUC people \citep{li2007and} (sport scenes). For detailed description of these datasets, we refer interested readers to several well-summarized papers \citep{andriluka20142d} and \citep{gong2016human}.

\begin{figure*}[!htb]
\centering
\includegraphics[width=0.97\textwidth]{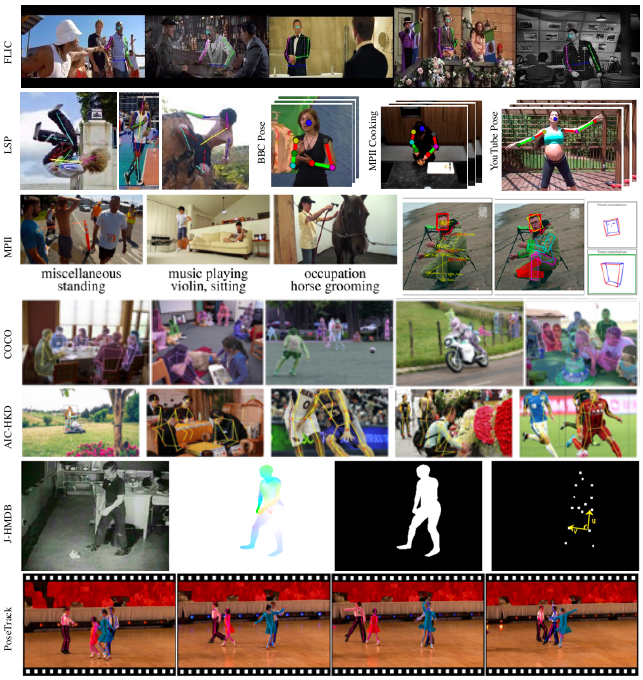}
\caption{Some selected example images with annotations from typical 2D human pose estimation datasets.}
\label{fig:datasets_2D}
\end{figure*}

Above earlier datasets for 2D human pose estimation have many shortcomings such as few scenes, monotonous view angle, lack of diverse activities, and limited number of images. The scale is the most important aspect of a dataset for deep learning-based methods. Small training sets are insufficient for learning robust features, unsuitable for networks with deep layers and complex design, and may easily cause overfitting. Thus in this section, we only introduce 2D human pose datasets with the number of images for training over 1,000. The features of these selected 2D HPE datasets are summarized in Table~\ref{tab:datasets_2D} and some sample images with annotations are illustrated in Fig.~\ref{fig:datasets_2D}.

\begin{table*}[!htb]
    \centering
    \footnotesize
    
    \caption{Popular 2D databases for human pose estimation. Selected example images with annotations are shown in Fig. \ref{fig:datasets_2D}. Here Jnt. indicates the number of joints}
    \label{tab:datasets_2D}
    \begin{tabular}{|p{1.1cm}<{\centering}|p{1.0cm}<{\centering}|c|c|c|c|c|p{7.6cm}<{\centering}|}
        \hline
        Dataset
        & Single/
        & \multirow{2}*{Jnt.}
        & \multicolumn{3}{c|}{Number of images/videos}
        & Evaluation
        & \multirow{2}*{Highlights}\\
        \cline{4-6}
        name
        & Multiple
        &
        &Train&Val&Test
        & protocol
        & \\
        \hline
        \multicolumn{8}{|l|}{\textbf{Image-based}}  \\
        \hline

        \raisebox{-0.3cm}[0pt]{FLIC}
        & \multirow{3}*{\raisebox{-0.7cm}[0pt]{single}}
        & \multirow{3}*{\raisebox{-0.7cm}[0pt]{10}}
        & \raisebox{-0.3cm}[0pt]{$\approx$5k}
        & \raisebox{-0.3cm}[0pt]{0}
        & \raisebox{-0.3cm}[0pt]{$\approx$1k}
        & \multirow{3}{1.5cm}{\raisebox{-0.8cm}[0pt]{PCP\&PCK}}
        & \multirow{3}{7.5cm}{Upper body poses; Sampled from movies; FLIC-full is complete version \citep{modec13}; FLIC-plus is cleaned version \citep{tompson2014joint}; FLIC is a simple version with no difficult poses.}\\[9pt]
        \cline{1-1}
        \cline{4-6}
        \raisebox{-0.3cm}[0pt]{\tabincell{c}{FLIC-full}}
        & &
        & \raisebox{-0.3cm}[0pt]{$\approx$20k}
        & \raisebox{-0.3cm}[0pt]{0}
        & \raisebox{-0.3cm}[0pt]{0}
        & & \\[9pt]
        \cline{1-1}
        \cline{4-6}
        \raisebox{-0.3cm}[0pt]{\tabincell{c}{FLIC-plus}}
        & &
        & \raisebox{-0.3cm}[0pt]{$\approx$17k}
        & \raisebox{-0.3cm}[0pt]{0}
        & \raisebox{-0.3cm}[0pt]{0}
        & & \\[9pt]
        \hline

        \raisebox{-0.3cm}[0pt]{LSP}
        & \multirow{2}*{\raisebox{-0.7cm}[0pt]{single}}
        & \multirow{2}*{\raisebox{-0.7cm}[0pt]{14}}
        & \raisebox{-0.3cm}[0pt]{$\approx$1k}
        & \raisebox{-0.3cm}[0pt]{0}
        & \raisebox{-0.3cm}[0pt]{$\approx$1k}
        & \multirow{2}*{\raisebox{-0.7cm}[0pt]{PCP}}
        & \multirow{2}{7.5cm}{full-body poses; From Flickr with 8 sports tags \citep{Johnson10}; Extended by adding most challenging poses lie in 3 tags \citep{Johnson11}.}\\[13pt]
        \cline{1-1}
        \cline{4-6}
        \raisebox{-0.3cm}[0pt]{\tabincell{c}{LSP-\\extended}}
        & &
        & \raisebox{-0.3cm}[0pt]{$\approx$10k}
        & \raisebox{-0.3cm}[0pt]{0}
        & \raisebox{-0.3cm}[0pt]{0}
        & & \\[13pt]
        \hline

        \multirow{2}*{\raisebox{-0.7cm}[0pt]{MPII}}
        & \raisebox{-0.3cm}[0pt]{single}
        & \multirow{2}*{\raisebox{-0.7cm}[0pt]{16}}
        & \raisebox{-0.3cm}[0pt]{$\approx$29k}
        & \raisebox{-0.3cm}[0pt]{0}
        & \raisebox{-0.3cm}[0pt]{$\approx$12k}
        & \raisebox{-0.3cm}[0pt]{PCPm/PCKh}
        & \multirow{2}{7.5cm}{Various body poses; Downloaded videos from YouTube; Multiple annotations (bounding boxes, 3D viewpoint of the head and torso, position of the eyes and nose, joint locations); \citep{andriluka20142d}.}\\[13pt]
        \cline{2-2}
        \cline{4-7}
        & \raisebox{-0.3cm}[0pt]{multiple}
        &
        & \raisebox{-0.3cm}[0pt]{$\approx$3.8k}
        & \raisebox{-0.3cm}[0pt]{0}
        & \raisebox{-0.3cm}[0pt]{$\approx$1.7k}
        & \raisebox{-0.3cm}[0pt]{mAP}
        & \\[13pt]
        \hline

        \raisebox{-0.3cm}[0pt]{COCO16}
        & \multirow{2}*{\raisebox{-0.7cm}[0pt]{multiple}}
        & \multirow{2}*{\raisebox{-0.7cm}[0pt]{17}}
        & \raisebox{-0.3cm}[0pt]{$\approx$45k}
        & \raisebox{-0.3cm}[0pt]{$\approx$22k}
        & \raisebox{-0.3cm}[0pt]{$\approx$80k}
        
        & \multirow{2}*{\raisebox{-0.7cm}[0pt]{AP}}
        & \multirow{2}{7.5cm}{Various body poses; From Google, Bing and Flickr; Multiple annotations (bounding boxes, human body masks, joint locations); With about 120K unlabeled images for semi-supervised learning; \citep{lin2014microsoft}}\\[13pt]
        \cline{1-1}
        \cline{4-6}
        \raisebox{-0.3cm}[0pt]{COCO17}
        &
        &
        & \raisebox{-0.3cm}[0pt]{$\approx$64k}
        & \raisebox{-0.3cm}[0pt]{$\approx$2.7k}
        & \raisebox{-0.3cm}[0pt]{$\approx$40k}
        &
        & \\[13pt]
        \hline

        \tabincell{c}{AIC-\\HKD}
        & multiple & 14
        & $\approx$210k & $\approx$30k
        & $\approx$60k
        & AP
        & Various body poses; From Internet search engines; Multiple annotations (bounding boxes, joint locations); \citep{wu2017ai}\\
        \hline
        \multicolumn{8}{|l|}{\textbf{Video-based}}  \\
        \hline

        \tabincell{c}{Penn\\Action}
        & single & 13 & $\approx$1k & 0 & $\approx$1k
        & -
        & \tabincell{l}{full-body poses; From YouTube; 15 actions; Multiple annotations\\(joint locations, bounding boxes, action classes) \citep{zhang2013actemes}.}\\
        
        \hline

        J-HMDB & single & 15
        & $\approx$0.6k
        & 0
        & $\approx$0.3k
        & -
        
        & \tabincell{l}{full-body poses; Generated from action recognition dataset; 21\\actions; Multiple annotations (joint positions and relations, optical\\flows, segmentation masks) \citep{Jhuang:ICCV:2013}.}\\
        \hline

        \tabincell{c}{PoseTrack}
        & multiple & 15
        & 292 & 50 & 208
        & mAP
        & \tabincell{l}{Various body poses; Extended from MPII; Dense annotations\\(joint locations, head bounding boxes) \citep{andriluka2018posetrack}.}\\
        \hline
    \end{tabular}
\end{table*}

\textbf{Frames Labeled In Cinema (FLIC) Dataset} \citep{modec13} contains $5,003$ images collected from popular Hollywood movies. For every tenth frame of 30 movies, a person detector~\citep{bourdev2009poselets} was run to obtain about 20K person candidates. Then all candidates are sent to Amazon Mechanical Turk to obtain ground truth labeling for 10 upper body joints. Finally, images with person occluded or severely non-frontal views are manually deleted. The undeleted original set called FLIC-full consisting of occluded, non-frontal, or just plain mislabeled examples ($20,928$ examples) is also available. Moreover, in \citep{tompson2014joint}, the FLIC-full dataset is further cleaned to FLIC-plus to make sure that the training subset does not include any images from the same scene as the test subset.

\textbf{Leeds Sports Pose (LSP) Dataset} \citep{Johnson10} contains $2,000$ images of full-body poses collected from Flickr by downloading with 8 sports tags (athletics, badminton, baseball, gymnastics, parkour, soccer, tennis, and volleyball). Each image is annotated with up to 14 visible joint locations. Further, the extension version Leeds Sports Pose Extended (LSP-extended) training dataset \citep{Johnson11} is gathered to extend the LSP dataset only for training. It contains $10,000$ images collected from Flickr searches with 3 most challenging tags (parkour, gymnastics, and athletics). The annotations were conducted through Amazon Mechanical Turk and the accuracy cannot be guaranteed.

\textbf{Max Planck Institute for Informatics (MPII) Human Pose Dataset} \citep{andriluka20142d} is one of current the state-of-the-art benchmarks for evaluation of articulated human pose estimation with rich annotations. First, with guidance from a two-level hierarchy of human activities from \citep{ainsworth20112011}, $3,913$ videos spanning $491$ different activities are downloaded from YouTube. Then frames that either contains different people in the video or the same person in a very different pose were manually selected which results in $24,920$ frames. Rich annotations including 16 body joints, the 3D viewpoint of the head and torso and position of the eyes and nose are labeled by in-house workers and on Amazon Mechanical Turk. For corresponding joints, visibility and left/right labels are also annotated in a person-centric way. Images in MPII have various body poses and are suitable for many tasks such as 2D single/multiple human pose estimation, action recognition, etc.

\textbf{Microsoft Common Objects in Context (COCO) Dataset} \citep{lin2014microsoft} is a large-scale dataset that was originally proposed for daily object detection and segmentation in natural environments. With improvements and extensions, the usage of COCO covers image captioning and keypoint detection. Images are collected from Google, Bing, and Flickr image search with isolated or pairwise object categories. Annotations were conducted on Amazon Mechanical Turk. The whole set contains more than $200,000$ images and $250,000$ labeled person instances. Suitable examples are selected for human pose estimation, thus forming two datasets: COCO keypoints 2016 and COCO keypoints 2017, corresponding to two public keypoint detection challenges respectively. The only difference between these two versions is the train/val/test splitting strategy based on community feedback (shown in Table \ref{tab:datasets_2D}), and cross-year results can be compared directly since the images in the test set are same. The COCO Keypoint Detection Challenge aims to localize keypoints of people in uncontrolled images. The annotations for each person include 17 body joints with visibility and left/right labels, and instance human body segmentation. Note that COCO dataset contains about 120K unlabeled images following the same class distribution as the labeled images which can be used for unsupervised or semi-supervised learning.

\textbf{AI Challenger Human Keypoint Detection (AIC-HKD) Dataset} \citep{wu2017ai} has the largest number of training examples. It contains $210,000$, $30,000$, $30,000$, and $30,000$ images for training, validation, test A, and test B respectively. The images, focusing on the daily life of people, were collected from Internet search engines. Then, after removing inappropriate examples (e.g. with the political, constabulary, violent and sexual contents; too small or too crowded human figures), each person in the images were annotated with a bounding box and 14 keypoints. Each keypoint has the visibility and left/right labels.

In addition to the datasets described above which are in static image style, datasets with densely annotated video frames are collected in closer to real-life application scenarios which offer the possibility to utilize temporal information and can be used for action recognition. Some of them focus on single individuals \citep{zhang2013actemes, Jhuang:ICCV:2013, charles2016personalizing} and others have pose annotations for multiple people  \citep{insafutdinov2017arttrack, iqbal2016posetrack, andriluka2018posetrack}.

\textbf{Penn Action Dataset} \citep{zhang2013actemes} consists of $2,326$ videos downloaded from YouTube covering 15 actions: baseball pitch, baseball swing, bench press, bowling, clean and jerk, golf swing, jump rope, jumping jacks, pull up, push up, sit up, squat, strum guitar, tennis forehand, and tennis serve. Annotations for each frame were labeled by VATIC \citep{vondrick2013efficiently} (an annotation tool) on Amazon Mechanical Turk. Each video involves an action class label and each video frame contains a bounding box of human and 13 joint locations with the visibility and left/right labels.

\textbf{Joint-annotated Human Motion Database (J-HMDB)} \citep{Jhuang:ICCV:2013} is based on the HMDB51 \citep{jhuang2011large} which is originally collected for action recognition. First, 21 action categories with relatively large body movements were selected from original 51 actions in HMDB51, including: brush hair, catch, clap, climb stairs, golf, jump, kick ball, pick, pour, pull-up, push, run, shoot ball, shoot bow, shoot gun, sit, stand, swing baseball, throw, walk, and wave. Then, after a selection-and-cleaning process, $928$ clips comprising $31,838$ annotated frames are selected. Finally, a 2D articulated human puppet model \citep{zuffi2012pictorial} is employed to generate all the needed annotations using Amazon Mechanical Turk. The 2D puppet model is an articulated human body model that provides scale, pose, segmentation, coarse viewpoint, and dense optical flow for the humans in actions. The annotations include 15 joint positions and relations, 2D optical flow corresponding to the human motion, human body segmentation mask. The 70$\%$ images are used for training and the 30$\%$ images for testing. J-HMDB can also be used for action recognition and human detection tasks.

There are several video datasets annotated with human upper body pose. \textbf{BBC Pose} \citep{charles2014automatic} contains 20 videos (10/5/5 for train/val/test, 1.5 million frames in total) with 9 sign language signers. $2,000$ frames for validation and test are manually annotated and the rest of the frames are annotated with a semi-automatic method. \textbf{Extended BBC Pose} dataset \citep{pfister2014deep} adds $72$ additional training videos for \textbf{BBC Pose} which has about $7$ million frames in total. \textbf{MPII Cooking} \citep{rohrbach2012database} dataset contains $1,071$ frames for training and $1,277$ frames for testing with manually annotated joint locations for cooking activities. \textbf{YouTube Pose} dataset \citep{charles2016personalizing} contains $50$ YouTube videos with single person in. The activities cover dancing, stand-up comedy, how-to, sports, disk jockeys, performing arts and dancing, and sign language signers. $100$ frames of each video are manually annotated with joint locations of the upper body. The scenes of these datasets are relatively simple, with static views and the characters are normally in a small motion range.

From unlabeled MPII Human Pose \citep{andriluka20142d} video data, there are several extended versions result in dense annotations of video frames. The general approach is to extend the original labeled frame with the connected frames both forward and backward and annotate unlabeled frames in the same way as the labeled frame. \textbf{MPII Video Pose} dataset \citep{insafutdinov2017arttrack} provides 28 videos containing 21 frames each by selecting the challenging labeled images and unlabeled neighboring +/-10 frames from the MPII dataset. In \textbf{Multi-Person PoseTrack} dataset \citep{iqbal2016posetrack}, each selected labeled frame is extended with unlabeled clips ranging +/-20 frames, and each person has a unique ID. Also, additional videos of more than 41 frames are provided for longer and variable-length sequences. In total, it contains 60 videos with additional videos with more than 41 frames for longer and variable-length sequences. \textbf{PoseTrack} dataset \citep{andriluka2018posetrack} is the integrated expansion of the above two datasets and is the current largest multi-person pose estimation and tracking dataset. Each person in the video has a unique track ID with annotations of  a head bounding box and 15 body joint locations. All pose annotations are labeled with VATIC \citep{vondrick2013efficiently}. PoseTrack contains $550$ video sequences with the frames mainly ranging between $41$ and $151$ frames in a wide variety of everyday human activities and is divided into $292$, $50$, and $208$ videos for training, validation, and testing, following original MPII split strategy.

\subsection{Evaluation Metrics of 2D human pose estimation}\label{sec5.2}

Different datasets have different features (e.g. various range of human body sizes, upper/full human body) and different task requirements (single/multiple pose estimation), so there are several evaluation metrics for 2D human pose estimation. The summary of different evaluation metrics which are commonly used are listed in Table \ref{tab:evaluation_2D}.

\begin{table*}[!htb]
\small
    \centering
    \footnotesize
    
    \caption{Summary of commonly used evaluation metrics for 2D HPE.}
    \label{tab:evaluation_2D}
    \begin{tabular}{|c|c|l|l|}
        \hline
        Metric
        & Meaning
        & \multicolumn{2}{c|}{Typical datasets and Description}\\
        \hline
        \multicolumn{4}{|l|}{\textbf{Single person}}  \\
        \hline
        PCP
        & \tabincell{c}{Percentage\\of Correct\\Parts}
        & LSP
        & \tabincell{c}{Percentage of correct predicted Parts which their end points fall within a threshold}  \\
        \hline
        PCK
        & \tabincell{c}{Percentage\\of Correct\\Keypoints}
        & \tabincell{c}{LSP\\MPII}
        & \tabincell{c}{Percentage of correct predicted joints which fall within a threshold} \\
        \hline
        \multicolumn{4}{|l|}{\textbf{Multiple person}}  \\
        \hline
        \multirow{7}{*}{AP}
        & \multirow{7}{*}{\tabincell{c}{Average\\Precision}}
        & \tabincell{c}{MPII\\PoseTrack}
        & \tabincell{l}{mean AP (mAP) is reported by AP for each body part after assigning predicted\\pose to the ground truth pose by PCKh score.}\\
        \cline{3-4}
        &
        & \multirow{5}{*}{\tabincell{c}{COCO}}
        & $\bullet$ $AP^{}_{coco}$: at OKS=.50:.05:.95 (primary metric)\\
        & & & $\bullet$ $AP^{OKS=.50}_{coco}$: at OKS=.50 (loose metric)\\
        & & & $\bullet$ $AP^{OKS=.75}_{coco}$: at OKS=.75 (strict metric)\\
        & & & $\bullet$ $AP^{medium}_{coco}$: for medium objects: $32^2$ $<$ area $<$ $96^2$\\
        & & & $\bullet$ $AP^{large}_{coco}$: for large objects: area $>$ $96^2$\\
        \hline
        \multirow{5}{*}{AR}
        & \multirow{5}{*}{\tabincell{c}{Average\\Recall}}
        & \multirow{5}{*}{\tabincell{c}{COCO}}
        & $\bullet$ $AR^{}_{coco}$: at OKS=.50:.05:.95 \\
        & & & $\bullet$ $AR^{OKS=.50}_{coco}$: at OKS=.50\\
        & & & $\bullet$ $AR^{OKS=.75}_{coco}$: at OKS=.75\\
        & & & $\bullet$ $AR^{medium}_{coco}$: for medium objects: $32^2$ $<$ area $<$ $96^2$\\
        & & & $\bullet$ $AR^{large}_{coco}$: for large objects: area $>$ $96^2$\\
        \hline
        OKS
        & \tabincell{c}{Object\\Keypoint\\Similarity}
        & COCO
        & \tabincell{c}{A similar role as the Intersection over Union (IoU) for AP/AR.}\\
        \hline

    \end{tabular}
\end{table*}

\textbf{Percentage of Correct Parts (PCP)} \citep{ferrari2008progressive} is widely used in early research. It reports the localization accuracy for limbs. A limb is correctly localized if its two endpoints are within a threshold from the corresponding ground truth endpoints. The threshold can be $50\%$ of the limb length. Besides a mean PCP, some limbs PCP (torso, upper legs, lower legs, upper arms, forearms, head) normally are also reported \citep{Johnson10}. And percentage curves for each limb can be obtained with the variation of threshold in the metric \citep{gkioxari2013articulated}. The similar metrics PCPm from \citep{andriluka20142d} use $50\%$ of the mean ground-truth segment length over the entire test set as a matching threshold.

\textbf{Percentage of Correct Keypoints (PCK)} \citep{yang2013articulated} measures the accuracy of the localization of the body joints. A candidate body joint is considered as correct if it falls within the threshold pixels of the ground-truth joint. The threshold can be a fraction of the person bounding box size \citep{yang2013articulated}, pixel radius that normalized by the torso height of each test sample \citep{modec13} (denoted as Percent of Detected Joints (PDJ) in \citep{toshev2014deeppose}), $50\%$ of the head segment length of each test image (denoted as \textbf{PCKh@0.5} in \citep{andriluka20142d}). Also, with the variation of a threshold, Area Under the Curve (AUC) can be generated for further analysis.

\textbf{The Average Precision (AP)}. For systems in which there are only joint locations but no annotated bounding boxes for human bodies/heads or number of people in the image as ground truth at testing, the detection problem must be addressed as well. Similar to object detection, an Average Precision (AP) evaluation method is proposed, which is first called Average Precision of Keypoints (APK) in \citep{yang2013articulated}. In AP measure, if a predicted joint falls within a threshold of the ground-truth joint location, it is counted as a true positive. Note that correspondence between candidates and ground-truth poses are established separately for each keypoint. For multi-person pose evaluation, all predicted poses are assigned to the ground truth poses one by one based on the PCKh score order, while unassigned predictions are counted as false positives \citep{pishchulin2016deepcut}. The mean average precision (mAP) is reported from the AP of each body joint.

\textbf{Average Precision (AP), Average Recall (AR) and their variants}. In \citep{lin2014microsoft}, evaluating multi-person pose estimation results as an object detection problem is further designed. AP, AR, and their variants are reported based on an analogous similarity measure: object keypoint similarity (OKS) which plays the same role as the Intersection over Union (IoU). Additional, AP/AR with different human body scales are also reported in COCO dataset. Table \ref{tab:evaluation_2D} summarizes all above evaluation metrics.

\textbf{Frame Rate, Number of Weights and Giga Floating-point Operations Per Second (GFLOPs)}. The computational performance metrics are also very important for HPE. Frame Rate indicates the processing speed of input data, generally expressed by Frames Per Second (FPS) or seconds per image (s/image) \citep{cao2016realtime}. Number of Weights and GFLOPs show the efficiency of the network, mainly related to the network design and the specific used GPUs/CPUs \citep{sun2019deep}. These computational performance metrics are suitable for 3D HPE as well.

\subsection{Datasets for 3D human pose estimation}\label{sec5.3}

\begin{figure*}[!htb]
\centering
\includegraphics[width=0.97\textwidth]{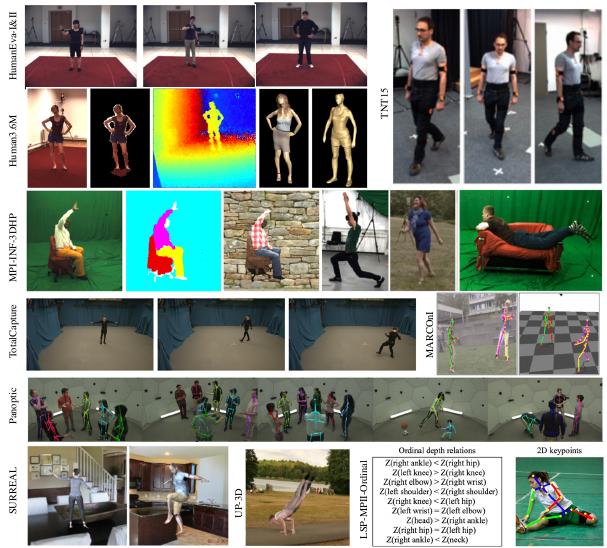}
\caption{Some selected example images with annotations from typical 3D human pose estimation datasets.}
\label{fig:datasets_3D}
\end{figure*}

For a better understanding of the human body in 3D space, there are many kinds of body representations with different modern equipment. 3D human body shape scans, such as \textbf{SCAPE} \citep{anguelov2005scape}, \textbf{INRIA4D} \citep{INRIA4D} and \textbf{FAUST} \citep{Bogo:CVPR:2014, dfaust:CVPR:2017}, 3D human body surface cloud points with time of flight (TOF) depth sensors \citep{shahroudy2016ntu}, 3D human body reflective markers capture with motion capture systems (MoCap) \citep{sigal2010humaneva, ionescu2014human3}, orientation and acceleration of 3D human body data with Inertial Measurement Unit (IMU) \citep{von2016human, von2018recovering}. It is difficult to summarize them all, this paper summarizes the datasets that involve RGB images and 3D joint coordinates. The details of the selected 3D datasets are summarized in Table \ref{tab:datasets_3D} and some example images with annotations are shown in Fig. \ref{fig:datasets_3D}.

\begin{table*}[!htb]
\small
    \centering
    \caption{Popular databases for 3D human pose estimation. Selected example images with annotations are shown in Fig. \ref{fig:datasets_3D} (Cams. indicates the number of cameras; Jnt. indicates the number of joints)}
    \label{tab:datasets_3D}
    \begin{tabular}{|c|p{0.7cm}<{\centering}|p{0.5cm}<{\centering}|c|c|c|p{1.5cm}<{\centering}|p{7.5cm}<{\centering}|}
        \hline
        Dataset
        & \multirow{2}*{Cams.}
        & \multirow{2}*{Jnt.}
        & \multicolumn{3}{c|}{Number of frames/videos}
        & Evaluation
        & \multirow{2}*{Highlights}\\
        \cline{4-6}
        name
        &
        &
        &Train&Val&Test
        & protocol
        & \\
        \hline
        \multicolumn{8}{|l|}{\textbf{Single person}}  \\
        \hline

        \raisebox{-0.3cm}[0pt]{HumanEva-I}
        & \raisebox{-0.3cm}[0pt]{7}
        & \multirow{2}*{\raisebox{-0.5cm}[0pt]{15}}
        & \raisebox{-0.3cm}[0pt]{$\approx$6.8k}
        & \raisebox{-0.3cm}[0pt]{$\approx$6.8k}
        & \raisebox{-0.3cm}[0pt]{$\approx$24k}
        & \multirow{2}{1.5cm}{\raisebox{-0.5cm}[0pt]{~~~~MPJPE}}
        & \multirow{2}{7.5cm}{4/2 (I/II) subjects, 6/1 (I/II) actions, Vicon data, indoor environment. \citep{sigal2010humaneva}}\\[9pt]
        \cline{1-2}
        \cline{4-6}
        \raisebox{-0.3cm}[0pt]{\tabincell{c}{HumanEva-II}}
        & \raisebox{-0.3cm}[0pt]{4}
        &
        & \raisebox{-0.3cm}[0pt]{0}
        & \raisebox{-0.3cm}[0pt]{0}
        & \raisebox{-0.3cm}[0pt]{$\approx$2.5k}
        & & \\[9pt]
        \hline

        \tabincell{c}{Human3.6M}
        & 4 & 17
        & $\approx$1.5M & $\approx$0.6M & $\approx$1.5M
        & MPJPE
        & 11 subjects, 17 actions, Vicon data, multi-annotation (3D joints, person bounding boxes, depth data, 3D body scans), indoor environment. \citep{ionescu2014human3}\\
        \hline

        \tabincell{c}{TNT15}
        & 8 & 15
        & \multicolumn{3}{c|}{$\approx$13k}
        & HumanEva
        & 4 subjects, 5 actions, IMU data, 3D body scans, indoor environment. \citep{von2016human}\\
        \hline

        \tabincell{c}{MPI-INF-3DHP}
        & 14 & 15
        & \multicolumn{3}{c|}{$\approx$1.3M}
        & 3DPCK
        & 8 subjects, 8 actions, commercial markerless system, indoor and outdoor scenes. \citep{mehta2017monocular}\\
        \hline

        \tabincell{c}{TotalCapture}
        & 8 & 26
        & \multicolumn{3}{c|}{$\approx$1.9M}
        & MPJPE
        & 5 subjects, 5 actions, IMU and Vicon data, indoors environment. \citep{trumble2017total}\\
        \hline

        \multicolumn{8}{|l|}{\textbf{Multiple person}}  \\
        \hline
 
        \tabincell{c}{Panoptic}
        & 521 & 15
        & \multicolumn{3}{c|}{65 videos (5.5 hours)}
        & 3DPCK
        & up to 8 subjects in each video, social interactions, markerless studio, multi-annotation (3D joints, cloud points, optical flow), indoors environment. \citep{Joo_2017_TPAMI}\\
        \hline
        
        \tabincell{c}{3DPW}
        & 1 & 18
        & \multicolumn{3}{c|}{60 videos ($\approx$51k frames)}
        & MPJPE MPJAE
        & 7 subjects(up to 2), daily actions, estimated 3D poses from videos and attached IMUs, 3D body scans, SMPL model fitting, in the wild. \citep{von2018recovering}\\
        \hline

    \end{tabular}
\end{table*}

\textbf{HumanEva-I\&II Datasets} \citep{sigal2010humaneva}. The ground truth annotations of both datasets were captured with a commercial MoCap system from ViconPeak. The HumanEva-I dataset contains 7-view video sequences (4 grayscales and 3 colors) which are synchronized with 3D body poses. There are 4 subjects with markers on their bodies performing 6 common actions (e.g. walking, jogging, gesturing, throwing and catching a ball, boxing, combo) in an 3m x 2m capture area. HumanEva-II is an extension of HumanEva-I dataset for testing, which contains 2 subjects performing the action combo.

\textbf{Human3.6M Dataset} \citep{ionescu2014human3} was collected using accurate marker-based MoCap systems~\citep{Vicon} in an indoor laboratory setup with 11 professional actors (5 females and 6 males) dressing moderately realistic clothing. It contains 3.6 million 3D human poses and corresponding images from 4 different views. The performed $17$ daily activities include discussion, smoking, taking photos, talking on the phone, etc. Main capturing devices include 4 digital video cameras, 1 time-of-flight sensor, 10 motion cameras working synchronously. The capture area is about 4m x 3m. The provided annotations include 3D joint positions, joint angles, person bounding boxes, and 3D laser scans of each actor. For evaluation, there are three protocols with different training and testing data splits (protocol \#1, protocol \#2 and protocol \#3.)

\textbf{TNT15 Dataset} \citep{von2016human} consists of synchronized data streams from $8$ RGB-cameras and $10$ IMUs. It has been recorded in an office environment. The dataset records 4 actors performing five activities (e.g. walking, running on the spot, rotating arms, jumping and skiing exercises, dynamic punching.) and contains about $13$k frames including binary segmented images obtained by background subtraction, 3D laser scans and registered meshes of each actor.

\textbf{MPI-INF-3DHP} \citep{mehta2017monocular} was collected with a markerless multi-camera MoCap system~\citep{TheCaptury} in both indoor and outdoor scenes. It contains over $1.3$M frames from $14$ different views. Eight subjects (4 females and 4 males) are recorded performing $8$ activities (e.g. walking/standing, exercise, sitting, crouch/reach, on the floor, sports, miscellaneous.)

\textbf{TotalCapture Dataset} \citep{trumble2017total} was captured in indoors with space measuring roughly 8m x 4m with $8$ calibrated HD video cameras at a frame rate of 60Hz. There are $4$ male and $1$ female subjects each performing four diverse performances, repeated 3 times: Range Of Motion (ROM), Walking, Acting, and Freestyle. There is a total of $1,892,176$ frames of synchronized video, IMU and Vicon data. The variation and body motions contained in particular within the acting and freestyle sequences are very challenging with actions such as yoga, giving directions, bending over and crawling performed in both the train and test data.

\textbf{MARCOnI Dataset} \citep{elhayek2017marconi} is a test dataset containing sequences in a variety of uncontrolled indoor and outdoor scenarios. The sequences vary according to different data modalities captured (multiple videos, video + marker positions), in the numbers and identities of actors to track, the complexity of the motions, the number of cameras used, the existence and number of moving objects in the background, and the lighting conditions (i.e. some body parts lit and some in shadow). Cameras differ in the types (from cell phones to vision cameras), the frame resolutions, and the frame rates.

\textbf{Panoptic Dataset} \citep{Joo_2017_TPAMI} was captured with a markerless motion capturing using multiple view systems which contains $480$ VGA camera views, $31$ HD views, $10$ RGB-D sensors and hardware-based synchronized system. It contains $65$ sequences (5.5 hours) of social interaction with $1.5$ millions of 3D skeletons. The annotations include 3D keypoints, cloud points, optical flow, etc.

\textbf{3DPW Dataset} \citep{von2018recovering} was captured with a single hand-held camera in natural environments. 3D annotations are estimated from IMUs attached to subjects' limbs with proposed method Video Inertial Poser. All subjects are provided with 3D scans. The dataset consists of $60$ video sequences (more than $51,000$ frames) with daily actions including walking in the city, going up-stairs, having coffee or taking the bus.

In addition to the datasets collected with MoCap systems, there are other approaches to create a dataset for 3D human pose estimation.
\textbf{JTA} (Joint Track Auto) \citep{fabbri2018learning} is a fully synthetic dataset generated from highly photorealistic video game Grand Theft Auto V. It contains almost $10M$ annotated body poses and over $460,800$ densely annotated frames.
In \textbf{Human3D+} \citep{chen2016synthesizing}, the training images are obtained by integrating real background images and 3D textured models which generated from SCAPE model \citep{anguelov2005scape} with different texture deformation. The parameters for generating basic SCAPE models are captured from a MoCap system, or inferred from human-annotated 2D poses.
\textbf{SURREAL} (Synthetic hUmans foR REAL) \citep{varol2017learning} contains videos of single synthetic people with real unchanged background. It contains annotations of body parts segmentation, depth, optical flow, and surface normals. The dataset employs the SMPL body model for generating body poses and shapes.
\textbf{LSP-MPII-Ordinal} \citep{pavlakos2018ordinal} is an extension of two 2D human pose datasets (LSP \citep{Johnson10} and MPII \citep{andriluka20142d}) by adding the ordinal depth relation for each pair of joints.
\textbf{UP-3D} \citep{lassner2017unite} is a combination of color images from 2D human pose benchmarks like LSP \citep{Johnson10} and MPII \citep{andriluka20142d} and human body model SMPL \citet{bogo2016keep}. The 3D human shape candidates are fit to color images by human annotators.
\textbf{DensePose} \citep{guler2018densepose} is an extension on 50K COCO images with people. All RGB images are manually annotated with surface-based representations of the human body.
\textbf{AMASS Dataset} \citep{AMASS2019} unifies 15 different optical marker-based human motion capture datasets with SMPL \citet{loper2015smpl} body model as a standard fitting representation for human skeleton and surface mesh. Each body joint in this rich dataset has 3 rotational Degrees of Freedom (DoF) which are parametrized with exponential coordinates.

\subsection{Evaluation Metrics of 3D human pose estimation}\label{sec5.4}

There are several evaluation metrics for 3D human pose estimation with different limitation factors. Note that we only list widely used evaluation metrics as below.

\textbf{Mean Per Joint Position Error (MPJPE)} is the most widely used measures to evaluate the performance of 3D pose estimation. It calculates the Euclidean distance from the estimated 3D joints to the ground truth in millimeters, averaged over all joints in one image. In the case of a set of frames, the mean error is averaged over all frames. For different datasets and different protocols, there are different data post-processing of estimated joints before computing the MPJPE. For example, in the protocol \#1 of Human3.6M, the MPJPE is calculated after aligning the depths of the root joints (generally pelvis joint) \citep{tome2017lifting, yang20183d}, which is also called N-MPJPE \citep{rhodin2018unsupervised}. The MPJPE in HumanEva-I and the protocol \#2 \& \#3 of Human3.6M is calculated after the alignment of predictions and ground truth with a rigid transformation using Procrustes Analysis \citep{gower1975generalized}, which is also called reconstruction error \citep{kanazawa2018end, pavlakos2018learning}, P-MPJPE \citep{rhodin2018unsupervised} or PA-MPJPE \citep{sun2018integral}.

\textbf{Percentage of Correct Keypoints (PCK)} and \textbf{Area Under the Curve
(AUC)} are suggested by \citep{mehta2017monocular} for 3D pose evaluation similar to PCK and AUC in MPII for 2D pose evaluation. PCK counts the percentage of points that fall in a threshold also called \textbf{3DPCK}, and AUC is computed by a range of PCK thresholds. The general threshold in 3D space is 150mm, corresponding to roughly half of the head size.

In addition to the evaluation metrics for 3D joint coordinates, there is another evaluation measurement \textbf{Mean Per-vertex Error} to report the results of 3D body shape which report the error between predicted and ground truth meshes \citep{varol2018bodynet, pavlakos2018learning}.

\section{Conclusion and Future Research Directions}\label{sec6}

Human pose estimation is a hot research area in computer vision that evolved recently along with the blooming of deep learning. Due to limitations in hardware device capability and the quantity and quality of training data, early networks are relatively shallow, used in a very straightforward way and can only handle small images or patches \citep{toshev2014deeppose, tompson2015efficient, li20143d}. More recent networks are more powerful, deeper and efficient \citep{newell2016stacked, cao2016realtime, he2017mask, sun2019deep}. In this paper, we have reviewed the recent deep learning-based research addressing the 2D/3D human pose estimation problem from monocular images or video footage and organize approaches into four categories based on specific tasks: (1) 2D single person pose estimation, (2) 2D multi-person pose estimation, (3) 3D single person pose estimation, and (4) 3D multi-person pose estimation. Further, we have summarized the popular human pose datasets and evaluation protocols.

Despite the great development of monocular human pose estimation with deep learning, there still remain some unresolved challenges and gap between research and practical applications, such as the influence of body part occlusion and crowded people. Efficient networks and adequate training data are the most important requirements for deep learning-based approaches.

Future networks should explore both global and local contexts for more discriminative features of the human body while exploiting human body structures into the network for prior constraints. Current networks have validated some effective network design tricks such as multi-stage structure, intermediate supervision, multi-scale feature fusion, multi-task learning, body structure constrains. Network efficiency is also a very important factor to apply algorithms in real-life applications.

Diversity data can improve the robustness of networks to handle complex scenes with irregular poses, occluded body limbs and crowded people. Data collection for specific complex scenarios is an option and there are other ways to extend existing datasets. Synthetic technology can theoretically generate unlimited data while there is a domain gap between synthetic data and real data. Cross-dataset supplementation, especially to supplement 3D datasets with 2D datasets can mitigate the problem of insufficient diversity of training data.

\section*{Acknowledgments}

This manuscript is based upon the work supported by National Science Foundation (NSF) under award number IIS-1400802 and National Natural Science Foundation of China (61420106007, 61671387). Yucheng Chen's contribution was made when he was a visiting student at the City University of New York, sponsored by the Chinese Scholarship Council.


\bibliographystyle{model2-names}
\bibliography{refs}

\begin{thebibliography}{187}
\expandafter\ifx\csname natexlab\endcsname\relax\def\natexlab#1{#1}\fi
\providecommand{\url}[1]{\texttt{#1}}
\providecommand{\href}[2]{#2}
\providecommand{\path}[1]{#1}
\providecommand{\DOIprefix}{doi:}
\providecommand{\ArXivprefix}{arXiv:}
\providecommand{\URLprefix}{URL: }
\providecommand{\Pubmedprefix}{pmid:}
\providecommand{\doi}[1]{\href{http://dx.doi.org/#1}{\path{#1}}}
\providecommand{\Pubmed}[1]{\href{pmid:#1}{\path{#1}}}
\providecommand{\bibinfo}[2]{#2}
\ifx\xfnm\relax \def\xfnm[#1]{\unskip,\space#1}\fi
\bibitem[{Aggarwal and Cai(1999)}]{aggarwal1999human}
\bibinfo{author}{Aggarwal, J.K.}, \bibinfo{author}{Cai, Q.},
  \bibinfo{year}{1999}.
\newblock \bibinfo{title}{Human motion analysis: A review}.
\newblock \bibinfo{journal}{Computer Vision and Image Understanding}
  \bibinfo{volume}{73}, \bibinfo{pages}{428--440}.
\bibitem[{Ainsworth et~al.(2011)Ainsworth, Haskell, Herrmann, Meckes,
  Bassett~Jr, Tudor-Locke, Greer, Vezina, Whitt-Glover and
  Leon}]{ainsworth20112011}
\bibinfo{author}{Ainsworth, B.E.}, \bibinfo{author}{Haskell, W.L.},
  \bibinfo{author}{Herrmann, S.D.}, \bibinfo{author}{Meckes, N.},
  \bibinfo{author}{Bassett~Jr, D.R.}, \bibinfo{author}{Tudor-Locke, C.},
  \bibinfo{author}{Greer, J.L.}, \bibinfo{author}{Vezina, J.},
  \bibinfo{author}{Whitt-Glover, M.C.}, \bibinfo{author}{Leon, A.S.},
  \bibinfo{year}{2011}.
\newblock \bibinfo{title}{2011 compendium of physical activities: a second
  update of codes and met values}.
\newblock \bibinfo{journal}{Medicine \& science in sports \& exercise}
  \bibinfo{volume}{43}, \bibinfo{pages}{1575--1581}.
\bibitem[{Andriluka et~al.(2018)Andriluka, Iqbal, Milan, Insafutdinov,
  Pishchulin, Gall and Schiele}]{andriluka2018posetrack}
\bibinfo{author}{Andriluka, M.}, \bibinfo{author}{Iqbal, U.},
  \bibinfo{author}{Milan, A.}, \bibinfo{author}{Insafutdinov, E.},
  \bibinfo{author}{Pishchulin, L.}, \bibinfo{author}{Gall, J.},
  \bibinfo{author}{Schiele, B.}, \bibinfo{year}{2018}.
\newblock \bibinfo{title}{Posetrack: A benchmark for human pose estimation and
  tracking}, in: \bibinfo{booktitle}{Proc. IEEE Conference on Computer Vision
  and Pattern Recognition}, pp. \bibinfo{pages}{5167--5176}.
\bibitem[{Andriluka et~al.(2014)Andriluka, Pishchulin, Gehler and
  Schiele}]{andriluka20142d}
\bibinfo{author}{Andriluka, M.}, \bibinfo{author}{Pishchulin, L.},
  \bibinfo{author}{Gehler, P.}, \bibinfo{author}{Schiele, B.},
  \bibinfo{year}{2014}.
\newblock \bibinfo{title}{2d human pose estimation: New benchmark and state of
  the art analysis}, in: \bibinfo{booktitle}{Proc. IEEE Conference on Computer
  Vision and Pattern Recognition}, pp. \bibinfo{pages}{3686--3693}.
\bibitem[{Anguelov et~al.(2005)Anguelov, Srinivasan, Koller, Thrun, Rodgers and
  Davis}]{anguelov2005scape}
\bibinfo{author}{Anguelov, D.}, \bibinfo{author}{Srinivasan, P.},
  \bibinfo{author}{Koller, D.}, \bibinfo{author}{Thrun, S.},
  \bibinfo{author}{Rodgers, J.}, \bibinfo{author}{Davis, J.},
  \bibinfo{year}{2005}.
\newblock \bibinfo{title}{Scape: shape completion and animation of people}, in:
  \bibinfo{booktitle}{ACM transactions on graphics},
  \bibinfo{organization}{ACM}. pp. \bibinfo{pages}{408--416}.
\bibitem[{Arnab et~al.(2019)Arnab, Doersch and Zisserman}]{arnab2019exploiting}
\bibinfo{author}{Arnab, A.}, \bibinfo{author}{Doersch, C.},
  \bibinfo{author}{Zisserman, A.}, \bibinfo{year}{2019}.
\newblock \bibinfo{title}{Exploiting temporal context for 3d human pose
  estimation in the wild}, in: \bibinfo{booktitle}{Proc. IEEE Conference on
  Computer Vision and Pattern Recognition}, pp. \bibinfo{pages}{3395--3404}.
\bibitem[{Belagiannis and Zisserman(2017)}]{belagiannis2017recurrent}
\bibinfo{author}{Belagiannis, V.}, \bibinfo{author}{Zisserman, A.},
  \bibinfo{year}{2017}.
\newblock \bibinfo{title}{Recurrent human pose estimation}, in:
  \bibinfo{booktitle}{Proc. IEEE Conference on Automatic Face and Gesture
  Recognition}, \bibinfo{organization}{IEEE}. pp. \bibinfo{pages}{468--475}.
\bibitem[{Bogo et~al.(2016)Bogo, Kanazawa, Lassner, Gehler, Romero and
  Black}]{bogo2016keep}
\bibinfo{author}{Bogo, F.}, \bibinfo{author}{Kanazawa, A.},
  \bibinfo{author}{Lassner, C.}, \bibinfo{author}{Gehler, P.},
  \bibinfo{author}{Romero, J.}, \bibinfo{author}{Black, M.J.},
  \bibinfo{year}{2016}.
\newblock \bibinfo{title}{Keep it smpl: Automatic estimation of 3d human pose
  and shape from a single image}, in: \bibinfo{booktitle}{Proc. European
  Conference on Computer Vision}, \bibinfo{organization}{Springer}. pp.
  \bibinfo{pages}{561--578}.
\bibitem[{Bogo et~al.(2014)Bogo, Romero, Loper and Black}]{Bogo:CVPR:2014}
\bibinfo{author}{Bogo, F.}, \bibinfo{author}{Romero, J.},
  \bibinfo{author}{Loper, M.}, \bibinfo{author}{Black, M.J.},
  \bibinfo{year}{2014}.
\newblock \bibinfo{title}{{FAUST}: Dataset and evaluation for {3D} mesh
  registration}, in: \bibinfo{booktitle}{Proc. IEEE Conference on Computer
  Vision and Pattern Recognition}, pp. \bibinfo{pages}{3794--3801}.
\bibitem[{Bogo et~al.(2017)Bogo, Romero, Pons-Moll and
  Black}]{dfaust:CVPR:2017}
\bibinfo{author}{Bogo, F.}, \bibinfo{author}{Romero, J.},
  \bibinfo{author}{Pons-Moll, G.}, \bibinfo{author}{Black, M.J.},
  \bibinfo{year}{2017}.
\newblock \bibinfo{title}{Dynamic {FAUST}: {R}egistering human bodies in
  motion}, in: \bibinfo{booktitle}{Proc. IEEE Conference on Computer Vision and
  Pattern Recognition}, pp. \bibinfo{pages}{6233--6242}.
\bibitem[{Bourdev and Malik(2009)}]{bourdev2009poselets}
\bibinfo{author}{Bourdev, L.}, \bibinfo{author}{Malik, J.},
  \bibinfo{year}{2009}.
\newblock \bibinfo{title}{Poselets: Body part detectors trained using 3d human
  pose annotations}, in: \bibinfo{booktitle}{Proc. IEEE International
  Conference on Computer Vision}, \bibinfo{organization}{IEEE}. pp.
  \bibinfo{pages}{1365--1372}.
\bibitem[{Bulat and Tzimiropoulos(2016)}]{bulat2016human}
\bibinfo{author}{Bulat, A.}, \bibinfo{author}{Tzimiropoulos, G.},
  \bibinfo{year}{2016}.
\newblock \bibinfo{title}{Human pose estimation via convolutional part heatmap
  regression}, in: \bibinfo{booktitle}{Proc. European Conference on Computer
  Vision}, \bibinfo{organization}{Springer}. pp. \bibinfo{pages}{717--732}.
\bibitem[{Cao et~al.(2016)Cao, Simon, Wei and Sheikh}]{cao2016realtime}
\bibinfo{author}{Cao, Z.}, \bibinfo{author}{Simon, T.}, \bibinfo{author}{Wei,
  S.E.}, \bibinfo{author}{Sheikh, Y.}, \bibinfo{year}{2016}.
\newblock \bibinfo{title}{Realtime multi-person 2d pose estimation using part
  affinity fields}.
\newblock \bibinfo{journal}{arXiv preprint arXiv:1611.08050} .
\bibitem[{Carreira et~al.(2016)Carreira, Agrawal, Fragkiadaki and
  Malik}]{carreira2016human}
\bibinfo{author}{Carreira, J.}, \bibinfo{author}{Agrawal, P.},
  \bibinfo{author}{Fragkiadaki, K.}, \bibinfo{author}{Malik, J.},
  \bibinfo{year}{2016}.
\newblock \bibinfo{title}{Human pose estimation with iterative error feedback},
  in: \bibinfo{booktitle}{Proc. IEEE Conference on Computer Vision and Pattern
  Recognition}, pp. \bibinfo{pages}{4733--4742}.
\bibitem[{Charles et~al.(2014)Charles, Pfister, Everingham and
  Zisserman}]{charles2014automatic}
\bibinfo{author}{Charles, J.}, \bibinfo{author}{Pfister, T.},
  \bibinfo{author}{Everingham, M.}, \bibinfo{author}{Zisserman, A.},
  \bibinfo{year}{2014}.
\newblock \bibinfo{title}{Automatic and efficient human pose estimation for
  sign language videos}.
\newblock \bibinfo{journal}{International Journal of Computer Vision}
  \bibinfo{volume}{110}, \bibinfo{pages}{70--90}.
\bibitem[{Charles et~al.(2016)Charles, Pfister, Magee, Hogg and
  Zisserman}]{charles2016personalizing}
\bibinfo{author}{Charles, J.}, \bibinfo{author}{Pfister, T.},
  \bibinfo{author}{Magee, D.}, \bibinfo{author}{Hogg, D.},
  \bibinfo{author}{Zisserman, A.}, \bibinfo{year}{2016}.
\newblock \bibinfo{title}{Personalizing human video pose estimation}, in:
  \bibinfo{booktitle}{Proc. IEEE Conference on Computer Vision and Pattern
  Recognition}, pp. \bibinfo{pages}{3063--3072}.
\bibitem[{Chen and Ramanan(2017)}]{chen20173d}
\bibinfo{author}{Chen, C.H.}, \bibinfo{author}{Ramanan, D.},
  \bibinfo{year}{2017}.
\newblock \bibinfo{title}{3d human pose estimation= 2d pose estimation+
  matching}, in: \bibinfo{booktitle}{Proc. IEEE Conference on Computer Vision
  and Pattern Recognition}, pp. \bibinfo{pages}{7035--7043}.
\bibitem[{Chen et~al.(2013)Chen, Wei and Ferryman}]{chen2013survey}
\bibinfo{author}{Chen, L.}, \bibinfo{author}{Wei, H.},
  \bibinfo{author}{Ferryman, J.}, \bibinfo{year}{2013}.
\newblock \bibinfo{title}{A survey of human motion analysis using depth
  imagery}.
\newblock \bibinfo{journal}{Pattern Recognition Letters} \bibinfo{volume}{34},
  \bibinfo{pages}{1995--2006}.
\bibitem[{Chen et~al.(2016)Chen, Wang, Li, Su, Wang, Tu, Lischinski, Cohen-Or
  and Chen}]{chen2016synthesizing}
\bibinfo{author}{Chen, W.}, \bibinfo{author}{Wang, H.}, \bibinfo{author}{Li,
  Y.}, \bibinfo{author}{Su, H.}, \bibinfo{author}{Wang, Z.},
  \bibinfo{author}{Tu, C.}, \bibinfo{author}{Lischinski, D.},
  \bibinfo{author}{Cohen-Or, D.}, \bibinfo{author}{Chen, B.},
  \bibinfo{year}{2016}.
\newblock \bibinfo{title}{Synthesizing training images for boosting human 3d
  pose estimation}, in: \bibinfo{booktitle}{Proc. IEEE International Conference
  on 3D Vision}, \bibinfo{organization}{IEEE}. pp. \bibinfo{pages}{479--488}.
\bibitem[{Chen and Yuille(2014)}]{chen2014articulated}
\bibinfo{author}{Chen, X.}, \bibinfo{author}{Yuille, A.L.},
  \bibinfo{year}{2014}.
\newblock \bibinfo{title}{Articulated pose estimation by a graphical model with
  image dependent pairwise relations}, in: \bibinfo{booktitle}{Advances in
  neural information processing systems}, pp. \bibinfo{pages}{1736--1744}.
\bibitem[{Chen et~al.(2017)Chen, Shen, Wei, Liu and Yang}]{chen2017adversarial}
\bibinfo{author}{Chen, Y.}, \bibinfo{author}{Shen, C.}, \bibinfo{author}{Wei,
  X.S.}, \bibinfo{author}{Liu, L.}, \bibinfo{author}{Yang, J.},
  \bibinfo{year}{2017}.
\newblock \bibinfo{title}{Adversarial posenet: A structure-aware convolutional
  network for human pose estimation}.
\newblock \bibinfo{journal}{CoRR, abs/1705.00389} \bibinfo{volume}{2}.
\bibitem[{Chen et~al.(2018)Chen, Wang, Peng, Zhang, Yu and Sun}]{Chen2018CPN}
\bibinfo{author}{Chen, Y.}, \bibinfo{author}{Wang, Z.}, \bibinfo{author}{Peng,
  Y.}, \bibinfo{author}{Zhang, Z.}, \bibinfo{author}{Yu, G.},
  \bibinfo{author}{Sun, J.}, \bibinfo{year}{2018}.
\newblock \bibinfo{title}{Cascaded pyramid network for multi-person pose
  estimation}, in: \bibinfo{booktitle}{Proc. IEEE Conference on Computer Vision
  and Pattern Recognition}, pp. \bibinfo{pages}{7103--7112}.
\bibitem[{Chou et~al.(2017)Chou, Chien and Chen}]{chou2017self}
\bibinfo{author}{Chou, C.J.}, \bibinfo{author}{Chien, J.T.},
  \bibinfo{author}{Chen, H.T.}, \bibinfo{year}{2017}.
\newblock \bibinfo{title}{Self adversarial training for human pose estimation}.
\newblock \bibinfo{journal}{arXiv preprint arXiv:1707.02439} .
\bibitem[{Chu et~al.(2016)Chu, Ouyang, Li and Wang}]{chu2016structured}
\bibinfo{author}{Chu, X.}, \bibinfo{author}{Ouyang, W.}, \bibinfo{author}{Li,
  H.}, \bibinfo{author}{Wang, X.}, \bibinfo{year}{2016}.
\newblock \bibinfo{title}{Structured feature learning for pose estimation}, in:
  \bibinfo{booktitle}{Proc. IEEE Conference on Computer Vision and Pattern
  Recognition}, pp. \bibinfo{pages}{4715--4723}.
\bibitem[{Chu et~al.(2017)Chu, Yang, Ouyang, Ma, Yuille and
  Wang}]{chu2017multi}
\bibinfo{author}{Chu, X.}, \bibinfo{author}{Yang, W.}, \bibinfo{author}{Ouyang,
  W.}, \bibinfo{author}{Ma, C.}, \bibinfo{author}{Yuille, A.L.},
  \bibinfo{author}{Wang, X.}, \bibinfo{year}{2017}.
\newblock \bibinfo{title}{Multi-context attention for human pose estimation}.
\newblock \bibinfo{journal}{arXiv preprint arXiv:1702.07432}
  \bibinfo{volume}{1}.
\bibitem[{Cootes et~al.(1995)Cootes, Taylor, Cooper and
  Graham}]{cootes1995active}
\bibinfo{author}{Cootes, T.F.}, \bibinfo{author}{Taylor, C.J.},
  \bibinfo{author}{Cooper, D.H.}, \bibinfo{author}{Graham, J.},
  \bibinfo{year}{1995}.
\newblock \bibinfo{title}{Active shape models-their training and application}.
\newblock \bibinfo{journal}{Computer Vision and Image Understanding}
  \bibinfo{volume}{61}, \bibinfo{pages}{38--59}.
\bibitem[{Dantone et~al.(2013)Dantone, Gall, Leistner and
  Van~Gool}]{dantone2013human}
\bibinfo{author}{Dantone, M.}, \bibinfo{author}{Gall, J.},
  \bibinfo{author}{Leistner, C.}, \bibinfo{author}{Van~Gool, L.},
  \bibinfo{year}{2013}.
\newblock \bibinfo{title}{Human pose estimation using body parts dependent
  joint regressors}, in: \bibinfo{booktitle}{Proc. IEEE Conference on Computer
  Vision and Pattern Recognition}, pp. \bibinfo{pages}{3041--3048}.
\bibitem[{Debnath et~al.(2018)Debnath, O'Brien, Yamaguchi and
  Behera}]{debnath2018adapting}
\bibinfo{author}{Debnath, B.}, \bibinfo{author}{O'Brien, M.},
  \bibinfo{author}{Yamaguchi, M.}, \bibinfo{author}{Behera, A.},
  \bibinfo{year}{2018}.
\newblock \bibinfo{title}{Adapting mobilenets for mobile based upper body pose
  estimation}, in: \bibinfo{booktitle}{Proc. IEEE Conference on Advanced Video
  and Signal Based Surveillance}, pp. \bibinfo{pages}{1--6}.
\bibitem[{Eichner and Ferrari(2010)}]{eichner2010we}
\bibinfo{author}{Eichner, M.}, \bibinfo{author}{Ferrari, V.},
  \bibinfo{year}{2010}.
\newblock \bibinfo{title}{We are family: Joint pose estimation of multiple
  persons}, in: \bibinfo{booktitle}{Proc. European Conference on Computer
  Vision}, \bibinfo{organization}{Springer}. pp. \bibinfo{pages}{228--242}.
\bibitem[{Eichner and Ferrari(2012a)}]{eichner2010upperbodydetector}
\bibinfo{author}{Eichner, M.}, \bibinfo{author}{Ferrari, V.},
  \bibinfo{year}{2012}a.
\newblock \bibinfo{title}{Calvin upper-body detector v1.04}.
\newblock \URLprefix
  \url{http://groups.inf.ed.ac.uk/calvin/calvin_upperbody_detector/}.
\bibitem[{Eichner and Ferrari(2012b)}]{eichner2012human}
\bibinfo{author}{Eichner, M.}, \bibinfo{author}{Ferrari, V.},
  \bibinfo{year}{2012}b.
\newblock \bibinfo{title}{Human pose co-estimation and applications}.
\newblock \bibinfo{journal}{IEEE transactions on pattern analysis and machine
  intelligence} \bibinfo{volume}{34}, \bibinfo{pages}{2282--2288}.
\bibitem[{Eichner et~al.(2009)Eichner, Ferrari and Zurich}]{eichner2009better}
\bibinfo{author}{Eichner, M.}, \bibinfo{author}{Ferrari, V.},
  \bibinfo{author}{Zurich, S.}, \bibinfo{year}{2009}.
\newblock \bibinfo{title}{Better appearance models for pictorial structures},
  in: \bibinfo{booktitle}{Proc. British Machine Vision Conference},
  p.~\bibinfo{pages}{5}.
\bibitem[{Elhayek et~al.(2017)Elhayek, de~Aguiar, Jain, Thompson, Pishchulin,
  Andriluka, Bregler, Schiele and Theobalt}]{elhayek2017marconi}
\bibinfo{author}{Elhayek, A.}, \bibinfo{author}{de~Aguiar, E.},
  \bibinfo{author}{Jain, A.}, \bibinfo{author}{Thompson, J.},
  \bibinfo{author}{Pishchulin, L.}, \bibinfo{author}{Andriluka, M.},
  \bibinfo{author}{Bregler, C.}, \bibinfo{author}{Schiele, B.},
  \bibinfo{author}{Theobalt, C.}, \bibinfo{year}{2017}.
\newblock \bibinfo{title}{Marconi—convnet-based marker-less motion capture in
  outdoor and indoor scenes}.
\newblock \bibinfo{journal}{IEEE transactions on pattern analysis and machine
  intelligence} \bibinfo{volume}{39}, \bibinfo{pages}{501--514}.
\bibitem[{Everingham et~al.(2010)Everingham, Van~Gool, Williams, Winn and
  Zisserman}]{everingham2010pascal}
\bibinfo{author}{Everingham, M.}, \bibinfo{author}{Van~Gool, L.},
  \bibinfo{author}{Williams, C.K.}, \bibinfo{author}{Winn, J.},
  \bibinfo{author}{Zisserman, A.}, \bibinfo{year}{2010}.
\newblock \bibinfo{title}{The pascal visual object classes (voc) challenge}.
\newblock \bibinfo{journal}{International journal of computer vision}
  \bibinfo{volume}{88}, \bibinfo{pages}{303--338}.
\bibitem[{Fabbri et~al.(2018)Fabbri, Lanzi, Calderara, Palazzi, Vezzani and
  Cucchiara}]{fabbri2018learning}
\bibinfo{author}{Fabbri, M.}, \bibinfo{author}{Lanzi, F.},
  \bibinfo{author}{Calderara, S.}, \bibinfo{author}{Palazzi, A.},
  \bibinfo{author}{Vezzani, R.}, \bibinfo{author}{Cucchiara, R.},
  \bibinfo{year}{2018}.
\newblock \bibinfo{title}{Learning to detect and track visible and occluded
  body joints in a virtual world}, in: \bibinfo{booktitle}{Proc. European
  Conference on Computer Vision}, pp. \bibinfo{pages}{430--446}.
\bibitem[{Faessler et~al.(2014)Faessler, Mueggler, Schwabe and
  Scaramuzza}]{faessler2014monocular}
\bibinfo{author}{Faessler, M.}, \bibinfo{author}{Mueggler, E.},
  \bibinfo{author}{Schwabe, K.}, \bibinfo{author}{Scaramuzza, D.},
  \bibinfo{year}{2014}.
\newblock \bibinfo{title}{A monocular pose estimation system based on infrared
  leds}, in: \bibinfo{booktitle}{Proc. IEEE International Conference on
  Robotics and Automation}, \bibinfo{organization}{IEEE}. pp.
  \bibinfo{pages}{907--913}.
\bibitem[{Fan et~al.(2015)Fan, Zheng, Lin and Wang}]{fan2015combining}
\bibinfo{author}{Fan, X.}, \bibinfo{author}{Zheng, K.}, \bibinfo{author}{Lin,
  Y.}, \bibinfo{author}{Wang, S.}, \bibinfo{year}{2015}.
\newblock \bibinfo{title}{Combining local appearance and holistic view:
  Dual-source deep neural networks for human pose estimation}.
\newblock \bibinfo{journal}{arXiv preprint arXiv:1504.07159} .
\bibitem[{Fang et~al.(2017)Fang, Xie, Tai and Lu}]{fang2017rmpe}
\bibinfo{author}{Fang, H.}, \bibinfo{author}{Xie, S.}, \bibinfo{author}{Tai,
  Y.W.}, \bibinfo{author}{Lu, C.}, \bibinfo{year}{2017}.
\newblock \bibinfo{title}{Rmpe: Regional multi-person pose estimation}, in:
  \bibinfo{booktitle}{Proc. IEEE International Conference on Computer Vision},
  pp. \bibinfo{pages}{2334--2343}.
\bibitem[{Felzenszwalb and Huttenlocher(2005)}]{felzenszwalb2005pictorial}
\bibinfo{author}{Felzenszwalb, P.F.}, \bibinfo{author}{Huttenlocher, D.P.},
  \bibinfo{year}{2005}.
\newblock \bibinfo{title}{Pictorial structures for object recognition}.
\newblock \bibinfo{journal}{International journal of computer vision}
  \bibinfo{volume}{61}, \bibinfo{pages}{55--79}.
\bibitem[{Feng et~al.(2019)Feng, Xiatian and Mao}]{Zhang2019Fast}
\bibinfo{author}{Feng, Z.}, \bibinfo{author}{Xiatian, Z.},
  \bibinfo{author}{Mao, Y.}, \bibinfo{year}{2019}.
\newblock \bibinfo{title}{Fast human pose estimation}, in:
  \bibinfo{booktitle}{Proc. IEEE Conference on Computer Vision and Pattern
  Recognition}, pp. \bibinfo{pages}{1--8}.
\bibitem[{Ferrari et~al.(2008)Ferrari, Marin-Jimenez and
  Zisserman}]{ferrari2008progressive}
\bibinfo{author}{Ferrari, V.}, \bibinfo{author}{Marin-Jimenez, M.},
  \bibinfo{author}{Zisserman, A.}, \bibinfo{year}{2008}.
\newblock \bibinfo{title}{Progressive search space reduction for human pose
  estimation}, in: \bibinfo{booktitle}{Proc. IEEE Conference on Computer Vision
  and Pattern Recognition}, pp. \bibinfo{pages}{1--8}.
\bibitem[{Gavrila(1999)}]{gavrila1999visual}
\bibinfo{author}{Gavrila, D.M.}, \bibinfo{year}{1999}.
\newblock \bibinfo{title}{The visual analysis of human movement: A survey}.
\newblock \bibinfo{journal}{Computer Vision and Image Understanding}
  \bibinfo{volume}{73}, \bibinfo{pages}{82--98}.
\bibitem[{Gkioxari et~al.(2013)Gkioxari, Arbelaez, Bourdev and
  Malik}]{gkioxari2013articulated}
\bibinfo{author}{Gkioxari, G.}, \bibinfo{author}{Arbelaez, P.},
  \bibinfo{author}{Bourdev, L.}, \bibinfo{author}{Malik, J.},
  \bibinfo{year}{2013}.
\newblock \bibinfo{title}{Articulated pose estimation using discriminative
  armlet classifiers}, in: \bibinfo{booktitle}{Proc. IEEE Conference on
  Computer Vision and Pattern Recognition}, pp. \bibinfo{pages}{3342--3349}.
\bibitem[{Gkioxari et~al.(2014a)Gkioxari, Hariharan, Girshick and
  Malik}]{gkioxari2014r}
\bibinfo{author}{Gkioxari, G.}, \bibinfo{author}{Hariharan, B.},
  \bibinfo{author}{Girshick, R.}, \bibinfo{author}{Malik, J.},
  \bibinfo{year}{2014}a.
\newblock \bibinfo{title}{R-cnns for pose estimation and action detection}.
\newblock \bibinfo{journal}{arXiv preprint arXiv:1406.5212} .
\bibitem[{Gkioxari et~al.(2014b)Gkioxari, Hariharan, Girshick and
  Malik}]{gkioxari2014using}
\bibinfo{author}{Gkioxari, G.}, \bibinfo{author}{Hariharan, B.},
  \bibinfo{author}{Girshick, R.}, \bibinfo{author}{Malik, J.},
  \bibinfo{year}{2014}b.
\newblock \bibinfo{title}{Using k-poselets for detecting people and localizing
  their keypoints}, in: \bibinfo{booktitle}{Proc. IEEE Conference on Computer
  Vision and Pattern Recognition}, pp. \bibinfo{pages}{3582--3589}.
\bibitem[{Gkioxari et~al.(2016)Gkioxari, Toshev and
  Jaitly}]{gkioxari2016chained}
\bibinfo{author}{Gkioxari, G.}, \bibinfo{author}{Toshev, A.},
  \bibinfo{author}{Jaitly, N.}, \bibinfo{year}{2016}.
\newblock \bibinfo{title}{Chained predictions using convolutional neural
  networks}, in: \bibinfo{booktitle}{Proc. European Conference on Computer
  Vision}, \bibinfo{organization}{Springer}. pp. \bibinfo{pages}{728--743}.
\bibitem[{Gong et~al.(2016)Gong, Zhang, Gonz{\`a}lez, Sobral, Bouwmans, Tu and
  Zahzah}]{gong2016human}
\bibinfo{author}{Gong, W.}, \bibinfo{author}{Zhang, X.},
  \bibinfo{author}{Gonz{\`a}lez, J.}, \bibinfo{author}{Sobral, A.},
  \bibinfo{author}{Bouwmans, T.}, \bibinfo{author}{Tu, C.},
  \bibinfo{author}{Zahzah, E.h.}, \bibinfo{year}{2016}.
\newblock \bibinfo{title}{Human pose estimation from monocular images: A
  comprehensive survey}.
\newblock \bibinfo{journal}{Sensors} \bibinfo{volume}{16},
  \bibinfo{pages}{1966}.
\bibitem[{Gower(1975)}]{gower1975generalized}
\bibinfo{author}{Gower, J.C.}, \bibinfo{year}{1975}.
\newblock \bibinfo{title}{Generalized procrustes analysis}.
\newblock \bibinfo{journal}{Psychometrika} \bibinfo{volume}{40},
  \bibinfo{pages}{33--51}.
\bibitem[{G{\"u}ler et~al.(2018)G{\"u}ler, Neverova and
  Kokkinos}]{guler2018densepose}
\bibinfo{author}{G{\"u}ler, R.A.}, \bibinfo{author}{Neverova, N.},
  \bibinfo{author}{Kokkinos, I.}, \bibinfo{year}{2018}.
\newblock \bibinfo{title}{Densepose: Dense human pose estimation in the wild},
  in: \bibinfo{booktitle}{Proc. IEEE Conference on Computer Vision and Pattern
  Recognition}, pp. \bibinfo{pages}{7297--7306}.
\bibitem[{Hasler et~al.(2009)Hasler, Stoll, Sunkel, Rosenhahn and
  Seidel}]{hasler2009statistical}
\bibinfo{author}{Hasler, N.}, \bibinfo{author}{Stoll, C.},
  \bibinfo{author}{Sunkel, M.}, \bibinfo{author}{Rosenhahn, B.},
  \bibinfo{author}{Seidel, H.P.}, \bibinfo{year}{2009}.
\newblock \bibinfo{title}{A statistical model of human pose and body shape},
  in: \bibinfo{booktitle}{Computer Graphics Forum},
  \bibinfo{organization}{Wiley Online Library}. pp. \bibinfo{pages}{337--346}.
\bibitem[{He et~al.(2017)He, Gkioxari, Doll{\'a}r and Girshick}]{he2017mask}
\bibinfo{author}{He, K.}, \bibinfo{author}{Gkioxari, G.},
  \bibinfo{author}{Doll{\'a}r, P.}, \bibinfo{author}{Girshick, R.},
  \bibinfo{year}{2017}.
\newblock \bibinfo{title}{Mask r-cnn}, in: \bibinfo{booktitle}{Proc. IEEE
  International Conference on Computer Vision}, \bibinfo{organization}{IEEE}.
  pp. \bibinfo{pages}{2980--2988}.
\bibitem[{Holte et~al.(2012)Holte, Tran, Trivedi and Moeslund}]{holte2012human}
\bibinfo{author}{Holte, M.B.}, \bibinfo{author}{Tran, C.},
  \bibinfo{author}{Trivedi, M.M.}, \bibinfo{author}{Moeslund, T.B.},
  \bibinfo{year}{2012}.
\newblock \bibinfo{title}{Human pose estimation and activity recognition from
  multi-view videos: Comparative explorations of recent developments}.
\newblock \bibinfo{journal}{IEEE Journal of selected topics in signal
  processing} \bibinfo{volume}{6}, \bibinfo{pages}{538--552}.
\bibitem[{Howard et~al.(2017)Howard, Zhu, Chen, Kalenichenko, Wang, Weyand,
  Andreetto and Adam}]{howard2017mobilenets}
\bibinfo{author}{Howard, A.G.}, \bibinfo{author}{Zhu, M.},
  \bibinfo{author}{Chen, B.}, \bibinfo{author}{Kalenichenko, D.},
  \bibinfo{author}{Wang, W.}, \bibinfo{author}{Weyand, T.},
  \bibinfo{author}{Andreetto, M.}, \bibinfo{author}{Adam, H.},
  \bibinfo{year}{2017}.
\newblock \bibinfo{title}{Mobilenets: Efficient convolutional neural networks
  for mobile vision applications}.
\newblock \bibinfo{journal}{arXiv preprint arXiv:1704.04861} .
\bibitem[{Hu et~al.(2004)Hu, Tan, Wang and Maybank}]{hu2004survey}
\bibinfo{author}{Hu, W.}, \bibinfo{author}{Tan, T.}, \bibinfo{author}{Wang,
  L.}, \bibinfo{author}{Maybank, S.}, \bibinfo{year}{2004}.
\newblock \bibinfo{title}{A survey on visual surveillance of object motion and
  behaviors}.
\newblock \bibinfo{journal}{IEEE Transactions on Systems, Man, and Cybernetics,
  Part C (Applications and Reviews)} \bibinfo{volume}{34},
  \bibinfo{pages}{334--352}.
\bibitem[{Huang et~al.(2017)Huang, Gong and Tao}]{huang2017coarse}
\bibinfo{author}{Huang, S.}, \bibinfo{author}{Gong, M.}, \bibinfo{author}{Tao,
  D.}, \bibinfo{year}{2017}.
\newblock \bibinfo{title}{A coarse-fine network for keypoint localization}, in:
  \bibinfo{booktitle}{Proc. IEEE International Conference on Computer Vision},
  pp. \bibinfo{pages}{3028--3037}.
\bibitem[{INRIA4D(accessed on 2019)}]{INRIA4D}
\bibinfo{author}{INRIA4D}, \bibinfo{year}{accessed on 2019}.
\newblock \URLprefix \url{http://4drepository.inrialpes.fr}.
\bibitem[{Insafutdinov et~al.(2017)Insafutdinov, Andriluka, Pishchulin, Tang,
  Levinkov, Andres and Schiele}]{insafutdinov2017arttrack}
\bibinfo{author}{Insafutdinov, E.}, \bibinfo{author}{Andriluka, M.},
  \bibinfo{author}{Pishchulin, L.}, \bibinfo{author}{Tang, S.},
  \bibinfo{author}{Levinkov, E.}, \bibinfo{author}{Andres, B.},
  \bibinfo{author}{Schiele, B.}, \bibinfo{year}{2017}.
\newblock \bibinfo{title}{Arttrack: Articulated multi-person tracking in the
  wild}, in: \bibinfo{booktitle}{Proc. IEEE Conference on Computer Vision and
  Pattern Recognition}, pp. \bibinfo{pages}{6457--6465}.
\bibitem[{Insafutdinov et~al.(2016)Insafutdinov, Pishchulin, Andres, Andriluka
  and Schiele}]{insafutdinov2016deepercut}
\bibinfo{author}{Insafutdinov, E.}, \bibinfo{author}{Pishchulin, L.},
  \bibinfo{author}{Andres, B.}, \bibinfo{author}{Andriluka, M.},
  \bibinfo{author}{Schiele, B.}, \bibinfo{year}{2016}.
\newblock \bibinfo{title}{Deepercut: A deeper, stronger, and faster
  multi-person pose estimation model}, in: \bibinfo{booktitle}{Proc. European
  Conference on Computer Vision}, \bibinfo{organization}{Springer}. pp.
  \bibinfo{pages}{34--50}.
\bibitem[{Ionescu et~al.(2014)Ionescu, Papava, Olaru and
  Sminchisescu}]{ionescu2014human3}
\bibinfo{author}{Ionescu, C.}, \bibinfo{author}{Papava, D.},
  \bibinfo{author}{Olaru, V.}, \bibinfo{author}{Sminchisescu, C.},
  \bibinfo{year}{2014}.
\newblock \bibinfo{title}{Human3.6m: Large scale datasets and predictive
  methods for 3d human sensing in natural environments}.
\newblock \bibinfo{journal}{IEEE Transactions on Pattern Analysis and Machine
  Intelligence} \bibinfo{volume}{36}, \bibinfo{pages}{1325--1339}.
\bibitem[{Iqbal and Gall(2016)}]{iqbal2016multi}
\bibinfo{author}{Iqbal, U.}, \bibinfo{author}{Gall, J.}, \bibinfo{year}{2016}.
\newblock \bibinfo{title}{Multi-person pose estimation with local
  joint-to-person associations}, in: \bibinfo{booktitle}{Proc. European
  Conference on Computer Vision}, \bibinfo{organization}{Springer}. pp.
  \bibinfo{pages}{627--642}.
\bibitem[{Iqbal et~al.(2016)Iqbal, Milan and Gall}]{iqbal2016posetrack}
\bibinfo{author}{Iqbal, U.}, \bibinfo{author}{Milan, A.},
  \bibinfo{author}{Gall, J.}, \bibinfo{year}{2016}.
\newblock \bibinfo{title}{Posetrack: Joint multi-person pose estimation and
  tracking}.
\newblock \bibinfo{journal}{arXiv:1611.07727} .
\bibitem[{Jaderberg et~al.(2015)Jaderberg, Simonyan, Zisserman
  et~al.}]{jaderberg2015spatial}
\bibinfo{author}{Jaderberg, M.}, \bibinfo{author}{Simonyan, K.},
  \bibinfo{author}{Zisserman, A.}, et~al., \bibinfo{year}{2015}.
\newblock \bibinfo{title}{Spatial transformer networks}, in:
  \bibinfo{booktitle}{Advances in neural information processing systems}, pp.
  \bibinfo{pages}{2017--2025}.
\bibitem[{Jain et~al.(2013)Jain, Tompson, Andriluka, Taylor and
  Bregler}]{jain2013learning}
\bibinfo{author}{Jain, A.}, \bibinfo{author}{Tompson, J.},
  \bibinfo{author}{Andriluka, M.}, \bibinfo{author}{Taylor, G.W.},
  \bibinfo{author}{Bregler, C.}, \bibinfo{year}{2013}.
\newblock \bibinfo{title}{Learning human pose estimation features with
  convolutional networks}.
\newblock \bibinfo{journal}{arXiv preprint arXiv:1312.7302} .
\bibitem[{Jain et~al.(2014)Jain, Tompson, LeCun and Bregler}]{jain2014modeep}
\bibinfo{author}{Jain, A.}, \bibinfo{author}{Tompson, J.},
  \bibinfo{author}{LeCun, Y.}, \bibinfo{author}{Bregler, C.},
  \bibinfo{year}{2014}.
\newblock \bibinfo{title}{Modeep: A deep learning framework using motion
  features for human pose estimation}, in: \bibinfo{booktitle}{Proc. Asian
  conference on computer vision}, \bibinfo{organization}{Springer}. pp.
  \bibinfo{pages}{302--315}.
\bibitem[{Jhuang et~al.(2013)Jhuang, Gall, Zuffi, Schmid and
  Black}]{Jhuang:ICCV:2013}
\bibinfo{author}{Jhuang, H.}, \bibinfo{author}{Gall, J.},
  \bibinfo{author}{Zuffi, S.}, \bibinfo{author}{Schmid, C.},
  \bibinfo{author}{Black, M.J.}, \bibinfo{year}{2013}.
\newblock \bibinfo{title}{Towards understanding action recognition}, in:
  \bibinfo{booktitle}{Proc. IEEE International Conference on Computer Vision},
  pp. \bibinfo{pages}{3192--3199}.
\bibitem[{Jhuang et~al.(2011)Jhuang, Garrote, Poggio, Serre and
  Hmdb}]{jhuang2011large}
\bibinfo{author}{Jhuang, H.}, \bibinfo{author}{Garrote, H.},
  \bibinfo{author}{Poggio, E.}, \bibinfo{author}{Serre, T.},
  \bibinfo{author}{Hmdb, T.}, \bibinfo{year}{2011}.
\newblock \bibinfo{title}{A large video database for human motion recognition},
  in: \bibinfo{booktitle}{Proc. IEEE International Conference on Computer
  Vision}, p.~\bibinfo{pages}{6}.
\bibitem[{Ji and Liu(2010)}]{ji2010advances}
\bibinfo{author}{Ji, X.}, \bibinfo{author}{Liu, H.}, \bibinfo{year}{2010}.
\newblock \bibinfo{title}{Advances in view-invariant human motion analysis: A
  review}.
\newblock \bibinfo{journal}{IEEE Transactions on Systems, Man, and Cybernetics,
  Part C (Applications and Reviews)} \bibinfo{volume}{40},
  \bibinfo{pages}{13--24}.
\bibitem[{Johnson and Everingham(2010)}]{Johnson10}
\bibinfo{author}{Johnson, S.}, \bibinfo{author}{Everingham, M.},
  \bibinfo{year}{2010}.
\newblock \bibinfo{title}{Clustered pose and nonlinear appearance models for
  human pose estimation}, in: \bibinfo{booktitle}{Proc. British Machine Vision
  Conference}, p.~\bibinfo{pages}{5}.
\bibitem[{Johnson and Everingham(2011)}]{Johnson11}
\bibinfo{author}{Johnson, S.}, \bibinfo{author}{Everingham, M.},
  \bibinfo{year}{2011}.
\newblock \bibinfo{title}{Learning effective human pose estimation from
  inaccurate annotation}, in: \bibinfo{booktitle}{Proc. IEEE Conference on
  Computer Vision and Pattern Recognition}, pp. \bibinfo{pages}{1465--1472}.
\bibitem[{Joo et~al.(2017)Joo, Simon, Li, Liu, Tan, Gui, Banerjee, Godisart,
  Nabbe, Matthews, Kanade, Nobuhara and Sheikh}]{Joo_2017_TPAMI}
\bibinfo{author}{Joo, H.}, \bibinfo{author}{Simon, T.}, \bibinfo{author}{Li,
  X.}, \bibinfo{author}{Liu, H.}, \bibinfo{author}{Tan, L.},
  \bibinfo{author}{Gui, L.}, \bibinfo{author}{Banerjee, S.},
  \bibinfo{author}{Godisart, T.S.}, \bibinfo{author}{Nabbe, B.},
  \bibinfo{author}{Matthews, I.}, \bibinfo{author}{Kanade, T.},
  \bibinfo{author}{Nobuhara, S.}, \bibinfo{author}{Sheikh, Y.},
  \bibinfo{year}{2017}.
\newblock \bibinfo{title}{Panoptic studio: A massively multiview system for
  social interaction capture}.
\newblock \bibinfo{journal}{IEEE Transactions on Pattern Analysis and Machine
  Intelligence} \bibinfo{volume}{41}, \bibinfo{pages}{190--204}.
\bibitem[{Joo et~al.(2018)Joo, Simon and Sheikh}]{joo2018total}
\bibinfo{author}{Joo, H.}, \bibinfo{author}{Simon, T.},
  \bibinfo{author}{Sheikh, Y.}, \bibinfo{year}{2018}.
\newblock \bibinfo{title}{Total capture: A 3d deformation model for tracking
  faces, hands, and bodies}, in: \bibinfo{booktitle}{Proc. IEEE Conference on
  Computer Vision and Pattern Recognition}, pp. \bibinfo{pages}{8320--8329}.
\bibitem[{Ju et~al.(1996)Ju, Black and Yacoob}]{ju1996cardboard}
\bibinfo{author}{Ju, S.X.}, \bibinfo{author}{Black, M.J.},
  \bibinfo{author}{Yacoob, Y.}, \bibinfo{year}{1996}.
\newblock \bibinfo{title}{Cardboard people: A parameterized model of
  articulated image motion}, in: \bibinfo{booktitle}{Proc. IEEE Conference on
  Automatic Face and Gesture Recognition}, pp. \bibinfo{pages}{38--44}.
\bibitem[{Kanazawa et~al.(2018)Kanazawa, Black, Jacobs and
  Malik}]{kanazawa2018end}
\bibinfo{author}{Kanazawa, A.}, \bibinfo{author}{Black, M.J.},
  \bibinfo{author}{Jacobs, D.W.}, \bibinfo{author}{Malik, J.},
  \bibinfo{year}{2018}.
\newblock \bibinfo{title}{End-to-end recovery of human shape and pose}, in:
  \bibinfo{booktitle}{Proc. IEEE Conference on Computer Vision and Pattern
  Recognition}, pp. \bibinfo{pages}{7122--7131}.
\bibitem[{Ke et~al.(2018)Ke, Chang, Qi and Lyu}]{ke2018multi}
\bibinfo{author}{Ke, L.}, \bibinfo{author}{Chang, M.C.}, \bibinfo{author}{Qi,
  H.}, \bibinfo{author}{Lyu, S.}, \bibinfo{year}{2018}.
\newblock \bibinfo{title}{Multi-scale structure-aware network for human pose
  estimation}.
\newblock \bibinfo{journal}{arXiv preprint arXiv:1803.09894} .
\bibitem[{Kinect(accessed on 2019)}]{Kinect}
\bibinfo{author}{Kinect}, \bibinfo{year}{accessed on 2019}.
\newblock \URLprefix
  \url{https://developer.microsoft.com/en-us/windows/kinect}.
\bibitem[{Kocabas et~al.(2018)Kocabas, Karagoz and
  Akbas}]{kocabas2018multiposenet}
\bibinfo{author}{Kocabas, M.}, \bibinfo{author}{Karagoz, S.},
  \bibinfo{author}{Akbas, E.}, \bibinfo{year}{2018}.
\newblock \bibinfo{title}{Multiposenet: Fast multi-person pose estimation using
  pose residual network}, in: \bibinfo{booktitle}{Proc. European Conference on
  Computer Vision}, \bibinfo{organization}{Springer}. pp.
  \bibinfo{pages}{437--453}.
\bibitem[{Kreiss et~al.(2019)Kreiss, Bertoni and Alahi}]{kreiss2019pifpaf}
\bibinfo{author}{Kreiss, S.}, \bibinfo{author}{Bertoni, L.},
  \bibinfo{author}{Alahi, A.}, \bibinfo{year}{2019}.
\newblock \bibinfo{title}{Pifpaf: Composite fields for human pose estimation},
  in: \bibinfo{booktitle}{Proc. IEEE Conference on Computer Vision and Pattern
  Recognition}, pp. \bibinfo{pages}{11977--11986}.
\bibitem[{Krizhevsky et~al.(2012)Krizhevsky, Sutskever and
  Hinton}]{krizhevsky2012imagenet}
\bibinfo{author}{Krizhevsky, A.}, \bibinfo{author}{Sutskever, I.},
  \bibinfo{author}{Hinton, G.E.}, \bibinfo{year}{2012}.
\newblock \bibinfo{title}{Imagenet classification with deep convolutional
  neural networks}, in: \bibinfo{booktitle}{Advances in neural information
  processing systems}, pp. \bibinfo{pages}{1097--1105}.
\bibitem[{Lassner et~al.(2017)Lassner, Romero, Kiefel, Bogo, Black and
  Gehler}]{lassner2017unite}
\bibinfo{author}{Lassner, C.}, \bibinfo{author}{Romero, J.},
  \bibinfo{author}{Kiefel, M.}, \bibinfo{author}{Bogo, F.},
  \bibinfo{author}{Black, M.J.}, \bibinfo{author}{Gehler, P.V.},
  \bibinfo{year}{2017}.
\newblock \bibinfo{title}{Unite the people: Closing the loop between 3d and 2d
  human representations}, in: \bibinfo{booktitle}{Proc. IEEE Conference on
  Computer Vision and Pattern Recognition}, pp. \bibinfo{pages}{4704--4713}.
\bibitem[{Li et~al.(2017a)Li, Chen, Chen, Dai and He}]{li2017Skeletonbox}
\bibinfo{author}{Li, B.}, \bibinfo{author}{Chen, H.}, \bibinfo{author}{Chen,
  Y.}, \bibinfo{author}{Dai, Y.}, \bibinfo{author}{He, M.},
  \bibinfo{year}{2017}a.
\newblock \bibinfo{title}{Skeleton boxes: Solving skeleton based action
  detection with a single deep convolutional neural network}, in:
  \bibinfo{booktitle}{Proc. IEEE International Conference on Multimedia and
  Expo Workshops}, pp. \bibinfo{pages}{613--616}.
\bibitem[{Li et~al.(2017b)Li, Dai, Cheng, Chen, Lin and He}]{li2017Skeleton}
\bibinfo{author}{Li, B.}, \bibinfo{author}{Dai, Y.}, \bibinfo{author}{Cheng,
  X.}, \bibinfo{author}{Chen, H.}, \bibinfo{author}{Lin, Y.},
  \bibinfo{author}{He, M.}, \bibinfo{year}{2017}b.
\newblock \bibinfo{title}{Skeleton based action recognition using
  translation-scale invariant image mapping and multi-scale deep cnn}, in:
  \bibinfo{booktitle}{Proc. IEEE International Conference on Multimedia and
  Expo Workshops}, pp. \bibinfo{pages}{601--604}.
\bibitem[{Li et~al.(2018a)Li, Dai and He}]{li2018monocular}
\bibinfo{author}{Li, B.}, \bibinfo{author}{Dai, Y.}, \bibinfo{author}{He, M.},
  \bibinfo{year}{2018}a.
\newblock \bibinfo{title}{Monocular depth estimation with hierarchical fusion
  of dilated cnns and soft-weighted-sum inference}.
\newblock \bibinfo{journal}{Pattern Recognition} \bibinfo{volume}{83},
  \bibinfo{pages}{328--339}.
\bibitem[{Li et~al.(2018b)Li, He, Dai, Cheng and Chen}]{li20183d}
\bibinfo{author}{Li, B.}, \bibinfo{author}{He, M.}, \bibinfo{author}{Dai, Y.},
  \bibinfo{author}{Cheng, X.}, \bibinfo{author}{Chen, Y.},
  \bibinfo{year}{2018}b.
\newblock \bibinfo{title}{3d skeleton based action recognition by video-domain
  translation-scale invariant mapping and multi-scale dilated cnn}.
\newblock \bibinfo{journal}{Multimedia Tools and Applications}
  \bibinfo{volume}{77}, \bibinfo{pages}{22901--22921}.
\bibitem[{Li et~al.(2015a)Li, Shen, Dai, Hengel and He}]{li2015depth}
\bibinfo{author}{Li, B.}, \bibinfo{author}{Shen, C.}, \bibinfo{author}{Dai,
  Y.}, \bibinfo{author}{Hengel, A.}, \bibinfo{author}{He, M.},
  \bibinfo{year}{2015}a.
\newblock \bibinfo{title}{Depth and surface normal estimation from monocular
  images using regression on deep features and hierarchical crfs}, in:
  \bibinfo{booktitle}{Proc. IEEE Conference on Computer Vision and Pattern
  Recognition}, pp. \bibinfo{pages}{1119--1127}.
\bibitem[{Li and Lee(2019)}]{li2019generating}
\bibinfo{author}{Li, C.}, \bibinfo{author}{Lee, G.H.}, \bibinfo{year}{2019}.
\newblock \bibinfo{title}{Generating multiple hypotheses for 3d human pose
  estimation with mixture density network}, in: \bibinfo{booktitle}{Proc. IEEE
  Conference on Computer Vision and Pattern Recognition}, pp.
  \bibinfo{pages}{9887--9895}.
\bibitem[{Li and Fei-fei(2007)}]{li2007and}
\bibinfo{author}{Li, L.}, \bibinfo{author}{Fei-fei, L.}, \bibinfo{year}{2007}.
\newblock \bibinfo{title}{What, where and who? classifying events by scene and
  object recognition}, in: \bibinfo{booktitle}{Proc. IEEE International
  Conference on Computer Vision}, p.~\bibinfo{pages}{6}.
\bibitem[{Li and Chan(2014)}]{li20143d}
\bibinfo{author}{Li, S.}, \bibinfo{author}{Chan, A.B.}, \bibinfo{year}{2014}.
\newblock \bibinfo{title}{3d human pose estimation from monocular images with
  deep convolutional neural network}, in: \bibinfo{booktitle}{Proc. Asian
  Conference on Computer Vision}, \bibinfo{organization}{Springer}. pp.
  \bibinfo{pages}{332--347}.
\bibitem[{Li et~al.(2014)Li, Liu and Chan}]{li2014heterogeneous}
\bibinfo{author}{Li, S.}, \bibinfo{author}{Liu, Z.Q.}, \bibinfo{author}{Chan,
  A.B.}, \bibinfo{year}{2014}.
\newblock \bibinfo{title}{Heterogeneous multi-task learning for human pose
  estimation with deep convolutional neural network}, in:
  \bibinfo{booktitle}{Proc. IEEE Conference on Computer Vision and Pattern
  Recognition Workshops}, pp. \bibinfo{pages}{482--489}.
\bibitem[{Li et~al.(2015b)Li, Zhang and Chan}]{li2015maximum}
\bibinfo{author}{Li, S.}, \bibinfo{author}{Zhang, W.}, \bibinfo{author}{Chan,
  A.B.}, \bibinfo{year}{2015}b.
\newblock \bibinfo{title}{Maximum-margin structured learning with deep networks
  for 3d human pose estimation}, in: \bibinfo{booktitle}{Proc. IEEE
  International Conference on Computer Vision}, pp.
  \bibinfo{pages}{2848--2856}.
\bibitem[{Li et~al.(2019)Li, Dekel, Cole, Tucker, Snavely, Liu and
  Freeman}]{li2019learning}
\bibinfo{author}{Li, Z.}, \bibinfo{author}{Dekel, T.}, \bibinfo{author}{Cole,
  F.}, \bibinfo{author}{Tucker, R.}, \bibinfo{author}{Snavely, N.},
  \bibinfo{author}{Liu, C.}, \bibinfo{author}{Freeman, W.},
  \bibinfo{year}{2019}.
\newblock \bibinfo{title}{Learning the depths of moving people by watching
  frozen} , \bibinfo{pages}{4521--4530}.
\bibitem[{Lifshitz et~al.(2016)Lifshitz, Fetaya and Ullman}]{lifshitz2016human}
\bibinfo{author}{Lifshitz, I.}, \bibinfo{author}{Fetaya, E.},
  \bibinfo{author}{Ullman, S.}, \bibinfo{year}{2016}.
\newblock \bibinfo{title}{Human pose estimation using deep consensus voting},
  in: \bibinfo{booktitle}{Proc. European Conference on Computer Vision},
  \bibinfo{organization}{Springer}. pp. \bibinfo{pages}{246--260}.
\bibitem[{Lin et~al.(2017)Lin, Doll{\'a}r, Girshick, He, Hariharan and
  Belongie}]{lin2017feature}
\bibinfo{author}{Lin, T.Y.}, \bibinfo{author}{Doll{\'a}r, P.},
  \bibinfo{author}{Girshick, R.}, \bibinfo{author}{He, K.},
  \bibinfo{author}{Hariharan, B.}, \bibinfo{author}{Belongie, S.},
  \bibinfo{year}{2017}.
\newblock \bibinfo{title}{Feature pyramid networks for object detection}, in:
  \bibinfo{booktitle}{Proc. IEEE Conference on Computer Vision and Pattern
  Recognition}, pp. \bibinfo{pages}{2117--2125}.
\bibitem[{Lin et~al.(2014)Lin, Maire, Belongie, Hays, Perona, Ramanan,
  Doll{\'a}r and Zitnick}]{lin2014microsoft}
\bibinfo{author}{Lin, T.Y.}, \bibinfo{author}{Maire, M.},
  \bibinfo{author}{Belongie, S.}, \bibinfo{author}{Hays, J.},
  \bibinfo{author}{Perona, P.}, \bibinfo{author}{Ramanan, D.},
  \bibinfo{author}{Doll{\'a}r, P.}, \bibinfo{author}{Zitnick, C.L.},
  \bibinfo{year}{2014}.
\newblock \bibinfo{title}{Microsoft coco: Common objects in context}, in:
  \bibinfo{booktitle}{Proc. European Conference on Computer Vision},
  \bibinfo{organization}{Springer}. pp. \bibinfo{pages}{740--755}.
\bibitem[{Liu et~al.(2015)Liu, Zhu, Bu and Chen}]{liu2015survey}
\bibinfo{author}{Liu, Z.}, \bibinfo{author}{Zhu, J.}, \bibinfo{author}{Bu, J.},
  \bibinfo{author}{Chen, C.}, \bibinfo{year}{2015}.
\newblock \bibinfo{title}{A survey of human pose estimation: the body parts
  parsing based methods}.
\newblock \bibinfo{journal}{Journal of Visual Communication and Image
  Representation} \bibinfo{volume}{32}, \bibinfo{pages}{10--19}.
\bibitem[{Long et~al.(2015)Long, Shelhamer and Darrell}]{long2015fully}
\bibinfo{author}{Long, J.}, \bibinfo{author}{Shelhamer, E.},
  \bibinfo{author}{Darrell, T.}, \bibinfo{year}{2015}.
\newblock \bibinfo{title}{Fully convolutional networks for semantic
  segmentation}, in: \bibinfo{booktitle}{Proc. IEEE Conference on Computer
  Vision and Pattern Recognition}, pp. \bibinfo{pages}{3431--3440}.
\bibitem[{Loper et~al.(2015)Loper, Mahmood, Romero, Pons-Moll and
  Black}]{loper2015smpl}
\bibinfo{author}{Loper, M.}, \bibinfo{author}{Mahmood, N.},
  \bibinfo{author}{Romero, J.}, \bibinfo{author}{Pons-Moll, G.},
  \bibinfo{author}{Black, M.J.}, \bibinfo{year}{2015}.
\newblock \bibinfo{title}{Smpl: A skinned multi-person linear model}.
\newblock \bibinfo{journal}{ACM Transactions on Graphics} \bibinfo{volume}{34},
  \bibinfo{pages}{248}.
\bibitem[{Luo et~al.(2018)Luo, Ren, Wang, Sun, Pan, Liu, Pang and
  Lin}]{luo2018lstm}
\bibinfo{author}{Luo, Y.}, \bibinfo{author}{Ren, J.}, \bibinfo{author}{Wang,
  Z.}, \bibinfo{author}{Sun, W.}, \bibinfo{author}{Pan, J.},
  \bibinfo{author}{Liu, J.}, \bibinfo{author}{Pang, J.}, \bibinfo{author}{Lin,
  L.}, \bibinfo{year}{2018}.
\newblock \bibinfo{title}{Lstm pose machines}, in: \bibinfo{booktitle}{Proc.
  IEEE Conference on Computer Vision and Pattern Recognition}, pp.
  \bibinfo{pages}{5207--5215}.
\bibitem[{Luvizon et~al.(2018)Luvizon, Picard and Tabia}]{luvizon20182d}
\bibinfo{author}{Luvizon, D.C.}, \bibinfo{author}{Picard, D.},
  \bibinfo{author}{Tabia, H.}, \bibinfo{year}{2018}.
\newblock \bibinfo{title}{2d/3d pose estimation and action recognition using
  multitask deep learning}, in: \bibinfo{booktitle}{Proc. IEEE Conference on
  Computer Vision and Pattern Recognition}, pp. \bibinfo{pages}{5137--5146}.
\bibitem[{Luvizon et~al.(2017)Luvizon, Tabia and Picard}]{luvizon2017human}
\bibinfo{author}{Luvizon, D.C.}, \bibinfo{author}{Tabia, H.},
  \bibinfo{author}{Picard, D.}, \bibinfo{year}{2017}.
\newblock \bibinfo{title}{Human pose regression by combining indirect part
  detection and contextual information}.
\newblock \bibinfo{journal}{arXiv preprint arXiv:1710.02322} .
\bibitem[{Mahmood et~al.(2019)Mahmood, Ghorbani, Troje, Pons-Moll and
  Black}]{AMASS2019}
\bibinfo{author}{Mahmood, N.}, \bibinfo{author}{Ghorbani, N.},
  \bibinfo{author}{Troje, N.F.}, \bibinfo{author}{Pons-Moll, G.},
  \bibinfo{author}{Black, M.J.}, \bibinfo{year}{2019}.
\newblock \bibinfo{title}{Amass: Archive of motion capture as surface shapes}.
\newblock \bibinfo{journal}{arXiv preprint arXiv:1904.03278} .
\bibitem[{von Marcard et~al.(2018)von Marcard, Henschel, Black, Rosenhahn and
  Pons-Moll}]{von2018recovering}
\bibinfo{author}{von Marcard, T.}, \bibinfo{author}{Henschel, R.},
  \bibinfo{author}{Black, M.J.}, \bibinfo{author}{Rosenhahn, B.},
  \bibinfo{author}{Pons-Moll, G.}, \bibinfo{year}{2018}.
\newblock \bibinfo{title}{Recovering accurate 3d human pose in the wild using
  imus and a moving camera}, in: \bibinfo{booktitle}{Proc. European Conference
  on Computer Vision}, pp. \bibinfo{pages}{601--617}.
\bibitem[{von Marcard et~al.(2016)von Marcard, Pons-Moll and
  Rosenhahn}]{von2016human}
\bibinfo{author}{von Marcard, T.}, \bibinfo{author}{Pons-Moll, G.},
  \bibinfo{author}{Rosenhahn, B.}, \bibinfo{year}{2016}.
\newblock \bibinfo{title}{Human pose estimation from video and imus}.
\newblock \bibinfo{journal}{IEEE transactions on pattern analysis and machine
  intelligence} \bibinfo{volume}{38}, \bibinfo{pages}{1533--1547}.
\bibitem[{Martinez et~al.(2017)Martinez, Hossain, Romero and
  Little}]{martinez2017simple}
\bibinfo{author}{Martinez, J.}, \bibinfo{author}{Hossain, R.},
  \bibinfo{author}{Romero, J.}, \bibinfo{author}{Little, J.J.},
  \bibinfo{year}{2017}.
\newblock \bibinfo{title}{A simple yet effective baseline for 3d human pose
  estimation}, in: \bibinfo{booktitle}{Proc. IEEE International Conference on
  Computer Vision}, pp. \bibinfo{pages}{2640--2649}.
\bibitem[{Mehta et~al.(2017a)Mehta, Rhodin, Casas, Fua, Sotnychenko, Xu and
  Theobalt}]{mehta2017monocular}
\bibinfo{author}{Mehta, D.}, \bibinfo{author}{Rhodin, H.},
  \bibinfo{author}{Casas, D.}, \bibinfo{author}{Fua, P.},
  \bibinfo{author}{Sotnychenko, O.}, \bibinfo{author}{Xu, W.},
  \bibinfo{author}{Theobalt, C.}, \bibinfo{year}{2017}a.
\newblock \bibinfo{title}{Monocular 3d human pose estimation in the wild using
  improved cnn supervision}, in: \bibinfo{booktitle}{Proc. IEEE International
  Conference on 3D Vision}, \bibinfo{organization}{IEEE}. pp.
  \bibinfo{pages}{506--516}.
\bibitem[{Mehta et~al.(2019)Mehta, Sotnychenko, Mueller, Xu, Elgharib, Fua,
  Seidel, Rhodin, Pons-Moll and Theobalt}]{mehta2019xnect}
\bibinfo{author}{Mehta, D.}, \bibinfo{author}{Sotnychenko, O.},
  \bibinfo{author}{Mueller, F.}, \bibinfo{author}{Xu, W.},
  \bibinfo{author}{Elgharib, M.}, \bibinfo{author}{Fua, P.},
  \bibinfo{author}{Seidel, H.P.}, \bibinfo{author}{Rhodin, H.},
  \bibinfo{author}{Pons-Moll, G.}, \bibinfo{author}{Theobalt, C.},
  \bibinfo{year}{2019}.
\newblock \bibinfo{title}{Xnect: Real-time multi-person 3d human pose
  estimation with a single rgb camera}.
\newblock \bibinfo{journal}{arXiv:1907.00837} .
\bibitem[{Mehta et~al.(2017b)Mehta, Sotnychenko, Mueller, Xu, Sridhar,
  Pons-Moll and Theobalt}]{mehta2017single}
\bibinfo{author}{Mehta, D.}, \bibinfo{author}{Sotnychenko, O.},
  \bibinfo{author}{Mueller, F.}, \bibinfo{author}{Xu, W.},
  \bibinfo{author}{Sridhar, S.}, \bibinfo{author}{Pons-Moll, G.},
  \bibinfo{author}{Theobalt, C.}, \bibinfo{year}{2017}b.
\newblock \bibinfo{title}{Single-shot multi-person 3d body pose estimation from
  monocular rgb input}.
\newblock \bibinfo{journal}{arXiv preprint arXiv:1712.03453} .
\bibitem[{Mehta et~al.(2017c)Mehta, Sridhar, Sotnychenko, Rhodin, Shafiei,
  Seidel, Xu, Casas and Theobalt}]{mehta2017vnect}
\bibinfo{author}{Mehta, D.}, \bibinfo{author}{Sridhar, S.},
  \bibinfo{author}{Sotnychenko, O.}, \bibinfo{author}{Rhodin, H.},
  \bibinfo{author}{Shafiei, M.}, \bibinfo{author}{Seidel, H.P.},
  \bibinfo{author}{Xu, W.}, \bibinfo{author}{Casas, D.},
  \bibinfo{author}{Theobalt, C.}, \bibinfo{year}{2017}c.
\newblock \bibinfo{title}{Vnect: Real-time 3d human pose estimation with a
  single rgb camera}.
\newblock \bibinfo{journal}{ACM Transactions on Graphics} \bibinfo{volume}{36},
  \bibinfo{pages}{44}.
\bibitem[{Meredith et~al.(2001)Meredith, Maddock et~al.}]{meredith2001motion}
\bibinfo{author}{Meredith, M.}, \bibinfo{author}{Maddock, S.}, et~al.,
  \bibinfo{year}{2001}.
\newblock \bibinfo{title}{Motion capture file formats explained}.
\newblock \bibinfo{journal}{Department of Computer Science, University of
  Sheffield} \bibinfo{volume}{211}, \bibinfo{pages}{241--244}.
\bibitem[{Moeslund and Granum(2001)}]{moeslund2001survey}
\bibinfo{author}{Moeslund, T.B.}, \bibinfo{author}{Granum, E.},
  \bibinfo{year}{2001}.
\newblock \bibinfo{title}{A survey of computer vision-based human motion
  capture}.
\newblock \bibinfo{journal}{Computer Vision and Image Understanding}
  \bibinfo{volume}{81}, \bibinfo{pages}{231--268}.
\bibitem[{Moeslund et~al.(2006)Moeslund, Hilton and
  Kr{\"u}ger}]{moeslund2006survey}
\bibinfo{author}{Moeslund, T.B.}, \bibinfo{author}{Hilton, A.},
  \bibinfo{author}{Kr{\"u}ger, V.}, \bibinfo{year}{2006}.
\newblock \bibinfo{title}{A survey of advances in vision-based human motion
  capture and analysis}.
\newblock \bibinfo{journal}{Computer Vision and Image Understanding}
  \bibinfo{volume}{104}, \bibinfo{pages}{90--126}.
\bibitem[{Moeslund et~al.(2011)Moeslund, Hilton, Kr{\"u}ger and
  Sigal}]{moeslund2011visual}
\bibinfo{author}{Moeslund, T.B.}, \bibinfo{author}{Hilton, A.},
  \bibinfo{author}{Kr{\"u}ger, V.}, \bibinfo{author}{Sigal, L.},
  \bibinfo{year}{2011}.
\newblock \bibinfo{title}{Visual analysis of humans}.
\newblock \bibinfo{publisher}{Springer}.
\bibitem[{Moon et~al.(2019)Moon, Chang and Lee}]{moon2019posefix}
\bibinfo{author}{Moon, G.}, \bibinfo{author}{Chang, J.Y.},
  \bibinfo{author}{Lee, K.M.}, \bibinfo{year}{2019}.
\newblock \bibinfo{title}{Posefix: Model-agnostic general human pose refinement
  network}, in: \bibinfo{booktitle}{Proc. IEEE Conference on Computer Vision
  and Pattern Recognition}, pp. \bibinfo{pages}{7773--7781}.
\bibitem[{Moreno-Noguer(2017)}]{moreno20173d}
\bibinfo{author}{Moreno-Noguer, F.}, \bibinfo{year}{2017}.
\newblock \bibinfo{title}{3d human pose estimation from a single image via
  distance matrix regression}, in: \bibinfo{booktitle}{Proc. IEEE Conference on
  Computer Vision and Pattern Recognition}, pp. \bibinfo{pages}{1561--1570}.
\bibitem[{Newell et~al.(2017)Newell, Huang and Deng}]{newell2017associative}
\bibinfo{author}{Newell, A.}, \bibinfo{author}{Huang, Z.},
  \bibinfo{author}{Deng, J.}, \bibinfo{year}{2017}.
\newblock \bibinfo{title}{Associative embedding: End-to-end learning for joint
  detection and grouping}, in: \bibinfo{booktitle}{Advances in Neural
  Information Processing Systems}, pp. \bibinfo{pages}{2277--2287}.
\bibitem[{Newell et~al.(2016)Newell, Yang and Deng}]{newell2016stacked}
\bibinfo{author}{Newell, A.}, \bibinfo{author}{Yang, K.},
  \bibinfo{author}{Deng, J.}, \bibinfo{year}{2016}.
\newblock \bibinfo{title}{Stacked hourglass networks for human pose
  estimation}, in: \bibinfo{booktitle}{Proc. European Conference on Computer
  Vision}, \bibinfo{organization}{Springer}. pp. \bibinfo{pages}{483--499}.
\bibitem[{Nibali et~al.(2018)Nibali, He, Morgan and
  Prendergast}]{nibali2018numerical}
\bibinfo{author}{Nibali, A.}, \bibinfo{author}{He, Z.},
  \bibinfo{author}{Morgan, S.}, \bibinfo{author}{Prendergast, L.},
  \bibinfo{year}{2018}.
\newblock \bibinfo{title}{Numerical coordinate regression with convolutional
  neural networks}.
\newblock \bibinfo{journal}{arXiv preprint arXiv:1801.07372} .
\bibitem[{Nie et~al.(2017)Nie, Wei and Zhu}]{xiaohan2017monocular}
\bibinfo{author}{Nie, B.X.}, \bibinfo{author}{Wei, P.}, \bibinfo{author}{Zhu,
  S.C.}, \bibinfo{year}{2017}.
\newblock \bibinfo{title}{Monocular 3d human pose estimation by predicting
  depth on joints}, in: \bibinfo{booktitle}{Proc. IEEE International Conference
  on Computer Vision}, pp. \bibinfo{pages}{3447--3455}.
\bibitem[{Nie et~al.(2018)Nie, Feng, Xing and Yan}]{nie2018pose}
\bibinfo{author}{Nie, X.}, \bibinfo{author}{Feng, J.}, \bibinfo{author}{Xing,
  J.}, \bibinfo{author}{Yan, S.}, \bibinfo{year}{2018}.
\newblock \bibinfo{title}{Pose partition networks for multi-person pose
  estimation}, in: \bibinfo{booktitle}{Proc. European Conference on Computer
  Vision}, pp. \bibinfo{pages}{684--699}.
\bibitem[{Ning et~al.(2018)Ning, Zhang and He}]{ning2018knowledge}
\bibinfo{author}{Ning, G.}, \bibinfo{author}{Zhang, Z.}, \bibinfo{author}{He,
  Z.}, \bibinfo{year}{2018}.
\newblock \bibinfo{title}{Knowledge-guided deep fractal neural networks for
  human pose estimation}.
\newblock \bibinfo{journal}{IEEE Transactions on Multimedia}
  \bibinfo{volume}{20}, \bibinfo{pages}{1246--1259}.
\bibitem[{Omran et~al.(2018)Omran, Lassner, Pons-Moll, Gehler and
  Schiele}]{omran2018neural}
\bibinfo{author}{Omran, M.}, \bibinfo{author}{Lassner, C.},
  \bibinfo{author}{Pons-Moll, G.}, \bibinfo{author}{Gehler, P.},
  \bibinfo{author}{Schiele, B.}, \bibinfo{year}{2018}.
\newblock \bibinfo{title}{Neural body fitting: Unifying deep learning and model
  based human pose and shape estimation}, in: \bibinfo{booktitle}{Proc. IEEE
  International Conference on 3D Vision}, \bibinfo{organization}{IEEE}. pp.
  \bibinfo{pages}{484--494}.
\bibitem[{Ouyang et~al.(2014)Ouyang, Chu and Wang}]{ouyang2014multi}
\bibinfo{author}{Ouyang, W.}, \bibinfo{author}{Chu, X.}, \bibinfo{author}{Wang,
  X.}, \bibinfo{year}{2014}.
\newblock \bibinfo{title}{Multi-source deep learning for human pose
  estimation}, in: \bibinfo{booktitle}{Proc. IEEE Conference on Computer Vision
  and Pattern Recognition}, pp. \bibinfo{pages}{2329--2336}.
\bibitem[{Papandreou et~al.(2018)Papandreou, Zhu, Chen, Gidaris, Tompson and
  Murphy}]{papandreou2018personlab}
\bibinfo{author}{Papandreou, G.}, \bibinfo{author}{Zhu, T.},
  \bibinfo{author}{Chen, L.C.}, \bibinfo{author}{Gidaris, S.},
  \bibinfo{author}{Tompson, J.}, \bibinfo{author}{Murphy, K.},
  \bibinfo{year}{2018}.
\newblock \bibinfo{title}{Personlab: Person pose estimation and instance
  segmentation with a bottom-up, part-based, geometric embedding model}.
\newblock \bibinfo{journal}{arXiv preprint arXiv:1803.08225} .
\bibitem[{Papandreou et~al.(2017)Papandreou, Zhu, Kanazawa, Toshev, Tompson,
  Bregler and Murphy}]{papandreou2017towards}
\bibinfo{author}{Papandreou, G.}, \bibinfo{author}{Zhu, T.},
  \bibinfo{author}{Kanazawa, N.}, \bibinfo{author}{Toshev, A.},
  \bibinfo{author}{Tompson, J.}, \bibinfo{author}{Bregler, C.},
  \bibinfo{author}{Murphy, K.}, \bibinfo{year}{2017}.
\newblock \bibinfo{title}{Towards accurate multi-person pose estimation in the
  wild}, in: \bibinfo{booktitle}{Proc. IEEE Conference on Computer Vision and
  Pattern Recognition}, pp. \bibinfo{pages}{4903--4911}.
\bibitem[{Pavlakos et~al.(2018a)Pavlakos, Zhou and
  Daniilidis}]{pavlakos2018ordinal}
\bibinfo{author}{Pavlakos, G.}, \bibinfo{author}{Zhou, X.},
  \bibinfo{author}{Daniilidis, K.}, \bibinfo{year}{2018}a.
\newblock \bibinfo{title}{Ordinal depth supervision for 3d human pose
  estimation}.
\newblock \bibinfo{journal}{arXiv preprint arXiv:1805.04095} .
\bibitem[{Pavlakos et~al.(2017)Pavlakos, Zhou, Derpanis and
  Daniilidis}]{pavlakos2017coarse}
\bibinfo{author}{Pavlakos, G.}, \bibinfo{author}{Zhou, X.},
  \bibinfo{author}{Derpanis, K.G.}, \bibinfo{author}{Daniilidis, K.},
  \bibinfo{year}{2017}.
\newblock \bibinfo{title}{Coarse-to-fine volumetric prediction for single-image
  3d human pose}, in: \bibinfo{booktitle}{Proc. IEEE Conference on Computer
  Vision and Pattern Recognition}, pp. \bibinfo{pages}{1263--1272}.
\bibitem[{Pavlakos et~al.(2018b)Pavlakos, Zhu, Zhou and
  Daniilidis}]{pavlakos2018learning}
\bibinfo{author}{Pavlakos, G.}, \bibinfo{author}{Zhu, L.},
  \bibinfo{author}{Zhou, X.}, \bibinfo{author}{Daniilidis, K.},
  \bibinfo{year}{2018}b.
\newblock \bibinfo{title}{Learning to estimate 3d human pose and shape from a
  single color image}.
\newblock \bibinfo{journal}{arXiv preprint arXiv:1805.04092} .
\bibitem[{Peng et~al.(2018)Peng, Tang, Yang, Feris and
  Metaxas}]{peng2018jointly}
\bibinfo{author}{Peng, X.}, \bibinfo{author}{Tang, Z.}, \bibinfo{author}{Yang,
  F.}, \bibinfo{author}{Feris, R.S.}, \bibinfo{author}{Metaxas, D.},
  \bibinfo{year}{2018}.
\newblock \bibinfo{title}{Jointly optimize data augmentation and network
  training: Adversarial data augmentation in human pose estimation}, in:
  \bibinfo{booktitle}{Proc. IEEE Conference on Computer Vision and Pattern
  Recognition}, pp. \bibinfo{pages}{2226--2234}.
\bibitem[{Perez-Sala et~al.(2014)Perez-Sala, Escalera, Angulo and
  Gonzalez}]{perez2014survey}
\bibinfo{author}{Perez-Sala, X.}, \bibinfo{author}{Escalera, S.},
  \bibinfo{author}{Angulo, C.}, \bibinfo{author}{Gonzalez, J.},
  \bibinfo{year}{2014}.
\newblock \bibinfo{title}{A survey on model based approaches for 2d and 3d
  visual human pose recovery}.
\newblock \bibinfo{journal}{Sensors} \bibinfo{volume}{14},
  \bibinfo{pages}{4189--4210}.
\bibitem[{Pfister et~al.(2015)Pfister, Charles and
  Zisserman}]{pfister2015flowing}
\bibinfo{author}{Pfister, T.}, \bibinfo{author}{Charles, J.},
  \bibinfo{author}{Zisserman, A.}, \bibinfo{year}{2015}.
\newblock \bibinfo{title}{Flowing convnets for human pose estimation in
  videos}, in: \bibinfo{booktitle}{Proc. IEEE International Conference on
  Computer Vision}, pp. \bibinfo{pages}{1913--1921}.
\bibitem[{Pfister et~al.(2014)Pfister, Simonyan, Charles and
  Zisserman}]{pfister2014deep}
\bibinfo{author}{Pfister, T.}, \bibinfo{author}{Simonyan, K.},
  \bibinfo{author}{Charles, J.}, \bibinfo{author}{Zisserman, A.},
  \bibinfo{year}{2014}.
\newblock \bibinfo{title}{Deep convolutional neural networks for efficient pose
  estimation in gesture videos}, in: \bibinfo{booktitle}{Proc. Asian Conference
  on Computer Vision}, \bibinfo{organization}{Springer}. pp.
  \bibinfo{pages}{538--552}.
\bibitem[{Pishchulin et~al.(2016)Pishchulin, Insafutdinov, Tang, Andres,
  Andriluka, Gehler and Schiele}]{pishchulin2016deepcut}
\bibinfo{author}{Pishchulin, L.}, \bibinfo{author}{Insafutdinov, E.},
  \bibinfo{author}{Tang, S.}, \bibinfo{author}{Andres, B.},
  \bibinfo{author}{Andriluka, M.}, \bibinfo{author}{Gehler, P.V.},
  \bibinfo{author}{Schiele, B.}, \bibinfo{year}{2016}.
\newblock \bibinfo{title}{Deepcut: Joint subset partition and labeling for
  multi person pose estimation}, in: \bibinfo{booktitle}{Proc. IEEE Conference
  on Computer Vision and Pattern Recognition}, pp. \bibinfo{pages}{4929--4937}.
\bibitem[{Pons-Moll et~al.(2015)Pons-Moll, Romero, Mahmood and
  Black}]{pons2015dyna}
\bibinfo{author}{Pons-Moll, G.}, \bibinfo{author}{Romero, J.},
  \bibinfo{author}{Mahmood, N.}, \bibinfo{author}{Black, M.J.},
  \bibinfo{year}{2015}.
\newblock \bibinfo{title}{Dyna: A model of dynamic human shape in motion}.
\newblock \bibinfo{journal}{ACM Transactions on Graphics} \bibinfo{volume}{34},
  \bibinfo{pages}{120}.
\bibitem[{Popa et~al.(2017)Popa, Zanfir and Sminchisescu}]{popa2017deep}
\bibinfo{author}{Popa, A.I.}, \bibinfo{author}{Zanfir, M.},
  \bibinfo{author}{Sminchisescu, C.}, \bibinfo{year}{2017}.
\newblock \bibinfo{title}{Deep multitask architecture for integrated 2d and 3d
  human sensing}, in: \bibinfo{booktitle}{Proc. IEEE Conference on Computer
  Vision and Pattern Recognition}, pp. \bibinfo{pages}{4714--4723}.
\bibitem[{Poppe(2007)}]{poppe2007vision}
\bibinfo{author}{Poppe, R.}, \bibinfo{year}{2007}.
\newblock \bibinfo{title}{Vision-based human motion analysis: An overview}.
\newblock \bibinfo{journal}{Computer Vision and Image Understanding}
  \bibinfo{volume}{108}, \bibinfo{pages}{4--18}.
\bibitem[{Qammaz and Argyros(2019)}]{qammaz2019mocapnet}
\bibinfo{author}{Qammaz, A.}, \bibinfo{author}{Argyros, A.},
  \bibinfo{year}{2019}.
\newblock \bibinfo{title}{Mocapnet: Ensemble of snn encoders for 3d human pose
  estimation in rgb images}, in: \bibinfo{booktitle}{Proc. British Machine
  VIsion Conference}.
\bibitem[{Rafi et~al.(2016)Rafi, Leibe, Gall and Kostrikov}]{rafi2016efficient}
\bibinfo{author}{Rafi, U.}, \bibinfo{author}{Leibe, B.}, \bibinfo{author}{Gall,
  J.}, \bibinfo{author}{Kostrikov, I.}, \bibinfo{year}{2016}.
\newblock \bibinfo{title}{An efficient convolutional network for human pose
  estimation}, in: \bibinfo{booktitle}{Proc. British Machine Vision
  Conference}, p.~\bibinfo{pages}{2}.
\bibitem[{Ramakrishna et~al.(2014)Ramakrishna, Munoz, Hebert, Bagnell and
  Sheikh}]{ramakrishna2014pose}
\bibinfo{author}{Ramakrishna, V.}, \bibinfo{author}{Munoz, D.},
  \bibinfo{author}{Hebert, M.}, \bibinfo{author}{Bagnell, J.A.},
  \bibinfo{author}{Sheikh, Y.}, \bibinfo{year}{2014}.
\newblock \bibinfo{title}{Pose machines: Articulated pose estimation via
  inference machines}, in: \bibinfo{booktitle}{Proc. European Conference on
  Computer Vision}, \bibinfo{organization}{Springer}. pp.
  \bibinfo{pages}{33--47}.
\bibitem[{Ren et~al.(2015)Ren, He, Girshick and Sun}]{ren2015faster}
\bibinfo{author}{Ren, S.}, \bibinfo{author}{He, K.}, \bibinfo{author}{Girshick,
  R.}, \bibinfo{author}{Sun, J.}, \bibinfo{year}{2015}.
\newblock \bibinfo{title}{Faster r-cnn: Towards real-time object detection with
  region proposal networks}, in: \bibinfo{booktitle}{Advances in neural
  information processing systems}, pp. \bibinfo{pages}{91--99}.
\bibitem[{Rhodin et~al.(2018a)Rhodin, Salzmann and
  Fua}]{rhodin2018unsupervised}
\bibinfo{author}{Rhodin, H.}, \bibinfo{author}{Salzmann, M.},
  \bibinfo{author}{Fua, P.}, \bibinfo{year}{2018}a.
\newblock \bibinfo{title}{Unsupervised geometry-aware representation for 3d
  human pose estimation}.
\newblock \bibinfo{journal}{arXiv:1804.01110} .
\bibitem[{Rhodin et~al.(2018b)Rhodin, Sp{\"o}rri, Katircioglu, Constantin,
  Meyer, M{\"u}ller, Salzmann and Fua}]{rhodin2018learning}
\bibinfo{author}{Rhodin, H.}, \bibinfo{author}{Sp{\"o}rri, J.},
  \bibinfo{author}{Katircioglu, I.}, \bibinfo{author}{Constantin, V.},
  \bibinfo{author}{Meyer, F.}, \bibinfo{author}{M{\"u}ller, E.},
  \bibinfo{author}{Salzmann, M.}, \bibinfo{author}{Fua, P.},
  \bibinfo{year}{2018}b.
\newblock \bibinfo{title}{Learning monocular 3d human pose estimation from
  multi-view images}, in: \bibinfo{booktitle}{Proc. IEEE Conference on Computer
  Vision and Pattern Recognition}, pp. \bibinfo{pages}{8437--8446}.
\bibitem[{Rogez et~al.(2017)Rogez, Weinzaepfel and Schmid}]{rogez2017lcr}
\bibinfo{author}{Rogez, G.}, \bibinfo{author}{Weinzaepfel, P.},
  \bibinfo{author}{Schmid, C.}, \bibinfo{year}{2017}.
\newblock \bibinfo{title}{Lcr-net: Localization-classification-regression for
  human pose}, in: \bibinfo{booktitle}{Proc. IEEE Conference on Computer Vision
  and Pattern Recognition}, pp. \bibinfo{pages}{3433--3441}.
\bibitem[{Rohrbach et~al.(2012)Rohrbach, Amin, Andriluka and
  Schiele}]{rohrbach2012database}
\bibinfo{author}{Rohrbach, M.}, \bibinfo{author}{Amin, S.},
  \bibinfo{author}{Andriluka, M.}, \bibinfo{author}{Schiele, B.},
  \bibinfo{year}{2012}.
\newblock \bibinfo{title}{A database for fine grained activity detection of
  cooking activities}, in: \bibinfo{booktitle}{Proc. IEEE Conference on
  Computer Vision and Pattern Recognition}, pp. \bibinfo{pages}{1194--1201}.
\bibitem[{Sapp and Taskar(2013)}]{modec13}
\bibinfo{author}{Sapp, B.}, \bibinfo{author}{Taskar, B.}, \bibinfo{year}{2013}.
\newblock \bibinfo{title}{Modec: Multimodal decomposable models for human pose
  estimation}, in: \bibinfo{booktitle}{Proc. IEEE Conference on Computer Vision
  and Pattern Recognition}, pp. \bibinfo{pages}{3674--3681}.
\bibitem[{Sapp et~al.(2011)Sapp, Weiss and Taskar}]{sapp2011parsing}
\bibinfo{author}{Sapp, B.}, \bibinfo{author}{Weiss, D.},
  \bibinfo{author}{Taskar, B.}, \bibinfo{year}{2011}.
\newblock \bibinfo{title}{Parsing human motion with stretchable models}, in:
  \bibinfo{booktitle}{Proc. IEEE Conference on Computer Vision and Pattern
  Recognition}, pp. \bibinfo{pages}{1281--1288}.
\bibitem[{Sarafianos et~al.(2016)Sarafianos, Boteanu, Ionescu and
  Kakadiaris}]{sarafianos20163d}
\bibinfo{author}{Sarafianos, N.}, \bibinfo{author}{Boteanu, B.},
  \bibinfo{author}{Ionescu, B.}, \bibinfo{author}{Kakadiaris, I.A.},
  \bibinfo{year}{2016}.
\newblock \bibinfo{title}{3d human pose estimation: A review of the literature
  and analysis of covariates}.
\newblock \bibinfo{journal}{Computer Vision and Image Understanding}
  \bibinfo{volume}{152}, \bibinfo{pages}{1--20}.
\bibitem[{Shahroudy et~al.(2016)Shahroudy, Liu, Ng and Wang}]{shahroudy2016ntu}
\bibinfo{author}{Shahroudy, A.}, \bibinfo{author}{Liu, J.},
  \bibinfo{author}{Ng, T.T.}, \bibinfo{author}{Wang, G.}, \bibinfo{year}{2016}.
\newblock \bibinfo{title}{Ntu rgb+ d: A large scale dataset for 3d human
  activity analysis}, in: \bibinfo{booktitle}{Proc. IEEE Conference on Computer
  Vision and Pattern Recognition}, pp. \bibinfo{pages}{1010--1019}.
\bibitem[{Shotton et~al.(2012)Shotton, Girshick, Fitzgibbon, Sharp, Cook,
  Finocchio, Moore, Kohli, Criminisi, Kipman et~al.}]{shotton2012efficient}
\bibinfo{author}{Shotton, J.}, \bibinfo{author}{Girshick, R.},
  \bibinfo{author}{Fitzgibbon, A.}, \bibinfo{author}{Sharp, T.},
  \bibinfo{author}{Cook, M.}, \bibinfo{author}{Finocchio, M.},
  \bibinfo{author}{Moore, R.}, \bibinfo{author}{Kohli, P.},
  \bibinfo{author}{Criminisi, A.}, \bibinfo{author}{Kipman, A.}, et~al.,
  \bibinfo{year}{2012}.
\newblock \bibinfo{title}{Efficient human pose estimation from single depth
  images}.
\newblock \bibinfo{journal}{IEEE transactions on pattern analysis and machine
  intelligence} \bibinfo{volume}{35}, \bibinfo{pages}{2821--2840}.
\bibitem[{Sidenbladh et~al.(2000)Sidenbladh, De~la Torre and
  Black}]{sidenbladh2000framework}
\bibinfo{author}{Sidenbladh, H.}, \bibinfo{author}{De~la Torre, F.},
  \bibinfo{author}{Black, M.J.}, \bibinfo{year}{2000}.
\newblock \bibinfo{title}{A framework for modeling the appearance of 3d
  articulated figures}, in: \bibinfo{booktitle}{Proc. IEEE Conference on
  Automatic Face and Gesture Recognition}, \bibinfo{organization}{IEEE}. pp.
  \bibinfo{pages}{368--375}.
\bibitem[{Sigal et~al.(2010)Sigal, Balan and Black}]{sigal2010humaneva}
\bibinfo{author}{Sigal, L.}, \bibinfo{author}{Balan, A.O.},
  \bibinfo{author}{Black, M.J.}, \bibinfo{year}{2010}.
\newblock \bibinfo{title}{Humaneva: Synchronized video and motion capture
  dataset and baseline algorithm for evaluation of articulated human motion}.
\newblock \bibinfo{journal}{International journal of computer vision}
  \bibinfo{volume}{87}, \bibinfo{pages}{4}.
\bibitem[{Sminchisescu(2008)}]{sminchisescu20083d}
\bibinfo{author}{Sminchisescu, C.}, \bibinfo{year}{2008}.
\newblock \bibinfo{title}{3d human motion analysis in monocular video:
  techniques and challenges}, in: \bibinfo{booktitle}{Human Motion}.
  \bibinfo{publisher}{Springer}, pp. \bibinfo{pages}{185--211}.
\bibitem[{Sun et~al.(2019)Sun, Xiao, Liu and Wang}]{sun2019deep}
\bibinfo{author}{Sun, K.}, \bibinfo{author}{Xiao, B.}, \bibinfo{author}{Liu,
  D.}, \bibinfo{author}{Wang, J.}, \bibinfo{year}{2019}.
\newblock \bibinfo{title}{Deep high-resolution representation learning for
  human pose estimation}, in: \bibinfo{booktitle}{Proc. IEEE Conference on
  Computer Vision and Pattern Recognition}.
\bibitem[{Sun et~al.(2017)Sun, Shang, Liang and Wei}]{sun2017compositional}
\bibinfo{author}{Sun, X.}, \bibinfo{author}{Shang, J.}, \bibinfo{author}{Liang,
  S.}, \bibinfo{author}{Wei, Y.}, \bibinfo{year}{2017}.
\newblock \bibinfo{title}{Compositional human pose regression}, in:
  \bibinfo{booktitle}{Proc. IEEE International Conference on Computer Vision},
  p.~\bibinfo{pages}{7}.
\bibitem[{Sun et~al.(2018)Sun, Xiao, Wei, Liang and Wei}]{sun2018integral}
\bibinfo{author}{Sun, X.}, \bibinfo{author}{Xiao, B.}, \bibinfo{author}{Wei,
  F.}, \bibinfo{author}{Liang, S.}, \bibinfo{author}{Wei, Y.},
  \bibinfo{year}{2018}.
\newblock \bibinfo{title}{Integral human pose regression}, in:
  \bibinfo{booktitle}{Proc. European Conference on Computer Vision}, pp.
  \bibinfo{pages}{529--545}.
\bibitem[{Szegedy et~al.(2016)Szegedy, Vanhoucke, Ioffe, Shlens and
  Wojna}]{szegedy2016rethinking}
\bibinfo{author}{Szegedy, C.}, \bibinfo{author}{Vanhoucke, V.},
  \bibinfo{author}{Ioffe, S.}, \bibinfo{author}{Shlens, J.},
  \bibinfo{author}{Wojna, Z.}, \bibinfo{year}{2016}.
\newblock \bibinfo{title}{Rethinking the inception architecture for computer
  vision}, in: \bibinfo{booktitle}{Proc. IEEE Conference on Computer Vision and
  Pattern Recognition}, pp. \bibinfo{pages}{2818--2826}.
\bibitem[{Tan et~al.(2017)Tan, Budvytis and Cipolla}]{tan2017indirect}
\bibinfo{author}{Tan, J.}, \bibinfo{author}{Budvytis, I.},
  \bibinfo{author}{Cipolla, R.}, \bibinfo{year}{2017}.
\newblock \bibinfo{title}{Indirect deep structured learning for 3d human body
  shape and pose prediction}, in: \bibinfo{booktitle}{Proc. British Machine
  Vision Conference}.
\bibitem[{Tang and Wu(2019)}]{tang2019does}
\bibinfo{author}{Tang, W.}, \bibinfo{author}{Wu, Y.}, \bibinfo{year}{2019}.
\newblock \bibinfo{title}{Does learning specific features for related parts
  help human pose estimation?}, in: \bibinfo{booktitle}{Proc. IEEE Conference
  on Computer Vision and Pattern Recognition}, pp. \bibinfo{pages}{1107--1116}.
\bibitem[{Tang et~al.(2018a)Tang, Yu and Wu}]{tang2018deeply}
\bibinfo{author}{Tang, W.}, \bibinfo{author}{Yu, P.}, \bibinfo{author}{Wu, Y.},
  \bibinfo{year}{2018}a.
\newblock \bibinfo{title}{Deeply learned compositional models for human pose
  estimation}, in: \bibinfo{booktitle}{Proc. European Conference on Computer
  Vision}, pp. \bibinfo{pages}{190--206}.
\bibitem[{Tang et~al.(2018b)Tang, Peng, Geng, Wu, Zhang and
  Metaxas}]{tang2018quantized}
\bibinfo{author}{Tang, Z.}, \bibinfo{author}{Peng, X.}, \bibinfo{author}{Geng,
  S.}, \bibinfo{author}{Wu, L.}, \bibinfo{author}{Zhang, S.},
  \bibinfo{author}{Metaxas, D.}, \bibinfo{year}{2018}b.
\newblock \bibinfo{title}{Quantized densely connected u-nets for efficient
  landmark localization}, in: \bibinfo{booktitle}{Proc. European Conference on
  Computer Vision}, pp. \bibinfo{pages}{339--354}.
\bibitem[{Tekin et~al.(2016)Tekin, Katircioglu, Salzmann, Lepetit and
  Fua}]{tekin2016structured}
\bibinfo{author}{Tekin, B.}, \bibinfo{author}{Katircioglu, I.},
  \bibinfo{author}{Salzmann, M.}, \bibinfo{author}{Lepetit, V.},
  \bibinfo{author}{Fua, P.}, \bibinfo{year}{2016}.
\newblock \bibinfo{title}{Structured prediction of 3d human pose with deep
  neural networks}.
\newblock \bibinfo{journal}{arXiv preprint arXiv:1605.05180} .
\bibitem[{Tekin et~al.(2017)Tekin, Marquez~Neila, Salzmann and
  Fua}]{tekin2017learning}
\bibinfo{author}{Tekin, B.}, \bibinfo{author}{Marquez~Neila, P.},
  \bibinfo{author}{Salzmann, M.}, \bibinfo{author}{Fua, P.},
  \bibinfo{year}{2017}.
\newblock \bibinfo{title}{Learning to fuse 2d and 3d image cues for monocular
  body pose estimation}, in: \bibinfo{booktitle}{Proc. IEEE International
  Conference on Computer Vision}, pp. \bibinfo{pages}{3941--3950}.
\bibitem[{TheCaptury(accessed on 2019)}]{TheCaptury}
\bibinfo{author}{TheCaptury}, \bibinfo{year}{accessed on 2019}.
\newblock \URLprefix \url{https://thecaptury.com/}.
\bibitem[{Tome et~al.(2017)Tome, Russell and Agapito}]{tome2017lifting}
\bibinfo{author}{Tome, D.}, \bibinfo{author}{Russell, C.},
  \bibinfo{author}{Agapito, L.}, \bibinfo{year}{2017}.
\newblock \bibinfo{title}{Lifting from the deep: Convolutional 3d pose
  estimation from a single image}, pp. \bibinfo{pages}{2500--2509}.
\bibitem[{Tompson et~al.(2015)Tompson, Goroshin, Jain, LeCun and
  Bregler}]{tompson2015efficient}
\bibinfo{author}{Tompson, J.}, \bibinfo{author}{Goroshin, R.},
  \bibinfo{author}{Jain, A.}, \bibinfo{author}{LeCun, Y.},
  \bibinfo{author}{Bregler, C.}, \bibinfo{year}{2015}.
\newblock \bibinfo{title}{Efficient object localization using convolutional
  networks}, in: \bibinfo{booktitle}{Proc. IEEE Conference on Computer Vision
  and Pattern Recognition}, pp. \bibinfo{pages}{648--656}.
\bibitem[{Tompson et~al.(2014)Tompson, Jain, LeCun and
  Bregler}]{tompson2014joint}
\bibinfo{author}{Tompson, J.J.}, \bibinfo{author}{Jain, A.},
  \bibinfo{author}{LeCun, Y.}, \bibinfo{author}{Bregler, C.},
  \bibinfo{year}{2014}.
\newblock \bibinfo{title}{Joint training of a convolutional network and a
  graphical model for human pose estimation}, in: \bibinfo{booktitle}{Advances
  in neural information processing systems}, pp. \bibinfo{pages}{1799--1807}.
\bibitem[{Toshev and Szegedy(2014)}]{toshev2014deeppose}
\bibinfo{author}{Toshev, A.}, \bibinfo{author}{Szegedy, C.},
  \bibinfo{year}{2014}.
\newblock \bibinfo{title}{Deeppose: Human pose estimation via deep neural
  networks}, in: \bibinfo{booktitle}{Proc. IEEE Conference on Computer Vision
  and Pattern Recognition}, pp. \bibinfo{pages}{1653--1660}.
\bibitem[{Trumble et~al.(2017)Trumble, Gilbert, Malleson, Hilton and
  Collomosse}]{trumble2017total}
\bibinfo{author}{Trumble, M.}, \bibinfo{author}{Gilbert, A.},
  \bibinfo{author}{Malleson, C.}, \bibinfo{author}{Hilton, A.},
  \bibinfo{author}{Collomosse, J.}, \bibinfo{year}{2017}.
\newblock \bibinfo{title}{Total capture: 3d human pose estimation fusing video
  and inertial sensors}, in: \bibinfo{booktitle}{Proc. British Machine Vision
  Conference}, pp. \bibinfo{pages}{1--13}.
\bibitem[{Varol et~al.(2018)Varol, Ceylan, Russell, Yang, Yumer, Laptev and
  Schmid}]{varol2018bodynet}
\bibinfo{author}{Varol, G.}, \bibinfo{author}{Ceylan, D.},
  \bibinfo{author}{Russell, B.}, \bibinfo{author}{Yang, J.},
  \bibinfo{author}{Yumer, E.}, \bibinfo{author}{Laptev, I.},
  \bibinfo{author}{Schmid, C.}, \bibinfo{year}{2018}.
\newblock \bibinfo{title}{Bodynet: Volumetric inference of 3d human body
  shapes}.
\newblock \bibinfo{journal}{arXiv preprint arXiv:1804.04875} .
\bibitem[{Varol et~al.(2017)Varol, Romero, Martin, Mahmood, Black, Laptev and
  Schmid}]{varol2017learning}
\bibinfo{author}{Varol, G.}, \bibinfo{author}{Romero, J.},
  \bibinfo{author}{Martin, X.}, \bibinfo{author}{Mahmood, N.},
  \bibinfo{author}{Black, M.J.}, \bibinfo{author}{Laptev, I.},
  \bibinfo{author}{Schmid, C.}, \bibinfo{year}{2017}.
\newblock \bibinfo{title}{Learning from synthetic humans}, in:
  \bibinfo{booktitle}{Proc. IEEE Conference on Computer Vision and Pattern
  Recognition}, pp. \bibinfo{pages}{4627--4635}.
\bibitem[{Vicon(accessed on 2019)}]{Vicon}
\bibinfo{author}{Vicon}, \bibinfo{year}{accessed on 2019}.
\newblock \URLprefix \url{https://www.vicon.com/}.
\bibitem[{Vondrick et~al.(2013)Vondrick, Patterson and
  Ramanan}]{vondrick2013efficiently}
\bibinfo{author}{Vondrick, C.}, \bibinfo{author}{Patterson, D.},
  \bibinfo{author}{Ramanan, D.}, \bibinfo{year}{2013}.
\newblock \bibinfo{title}{Efficiently scaling up crowdsourced video
  annotation}.
\newblock \bibinfo{journal}{International Journal of Computer Vision}
  \bibinfo{volume}{101}, \bibinfo{pages}{184--204}.
\bibitem[{Wang et~al.(2018a)Wang, Chen, Liu, Qian, Lin and
  Ma}]{wang2018drpose3d}
\bibinfo{author}{Wang, M.}, \bibinfo{author}{Chen, X.}, \bibinfo{author}{Liu,
  W.}, \bibinfo{author}{Qian, C.}, \bibinfo{author}{Lin, L.},
  \bibinfo{author}{Ma, L.}, \bibinfo{year}{2018}a.
\newblock \bibinfo{title}{Drpose3d: Depth ranking in 3d human pose estimation}.
\newblock \bibinfo{journal}{arXiv preprint arXiv:1805.08973} .
\bibitem[{Wang et~al.(2018b)Wang, Li, Ogunbona, Wan and Escalera}]{wang2018rgb}
\bibinfo{author}{Wang, P.}, \bibinfo{author}{Li, W.},
  \bibinfo{author}{Ogunbona, P.}, \bibinfo{author}{Wan, J.},
  \bibinfo{author}{Escalera, S.}, \bibinfo{year}{2018}b.
\newblock \bibinfo{title}{Rgb-d-based human motion recognition with deep
  learning: A survey}.
\newblock \bibinfo{journal}{Computer Vision and Image Understanding}
  \bibinfo{volume}{171}, \bibinfo{pages}{118--139}.
\bibitem[{Wang et~al.(2011)Wang, Tran and Liao}]{wang2011learning}
\bibinfo{author}{Wang, Y.}, \bibinfo{author}{Tran, D.}, \bibinfo{author}{Liao,
  Z.}, \bibinfo{year}{2011}.
\newblock \bibinfo{title}{Learning hierarchical poselets for human parsing},
  in: \bibinfo{booktitle}{Proc. IEEE Conference on Computer Vision and Pattern
  Recognition}, pp. \bibinfo{pages}{1705--1712}.
\bibitem[{Wei et~al.(2016)Wei, Ramakrishna, Kanade and
  Sheikh}]{wei2016convolutional}
\bibinfo{author}{Wei, S.E.}, \bibinfo{author}{Ramakrishna, V.},
  \bibinfo{author}{Kanade, T.}, \bibinfo{author}{Sheikh, Y.},
  \bibinfo{year}{2016}.
\newblock \bibinfo{title}{Convolutional pose machines}, in:
  \bibinfo{booktitle}{Proc. IEEE Conference on Computer Vision and Pattern
  Recognition}, pp. \bibinfo{pages}{4724--4732}.
\bibitem[{Wu et~al.(2017)Wu, Zheng, Zhao, Li, Yan, Liang, Wang, Zhou, Lin, Fu
  et~al.}]{wu2017ai}
\bibinfo{author}{Wu, J.}, \bibinfo{author}{Zheng, H.}, \bibinfo{author}{Zhao,
  B.}, \bibinfo{author}{Li, Y.}, \bibinfo{author}{Yan, B.},
  \bibinfo{author}{Liang, R.}, \bibinfo{author}{Wang, W.},
  \bibinfo{author}{Zhou, S.}, \bibinfo{author}{Lin, G.}, \bibinfo{author}{Fu,
  Y.}, et~al., \bibinfo{year}{2017}.
\newblock \bibinfo{title}{Ai challenger: A large-scale dataset for going deeper
  in image understanding}.
\newblock \bibinfo{journal}{arXiv preprint arXiv:1711.06475} .
\bibitem[{Xiao et~al.(2018)Xiao, Wu and Wei}]{xiao2018simple}
\bibinfo{author}{Xiao, B.}, \bibinfo{author}{Wu, H.}, \bibinfo{author}{Wei,
  Y.}, \bibinfo{year}{2018}.
\newblock \bibinfo{title}{Simple baselines for human pose estimation and
  tracking}, in: \bibinfo{booktitle}{Proc. European Conference on Computer
  Vision}, pp. \bibinfo{pages}{466--481}.
\bibitem[{Yang et~al.(2017)Yang, Li, Ouyang, Li and Wang}]{yang2017learning}
\bibinfo{author}{Yang, W.}, \bibinfo{author}{Li, S.}, \bibinfo{author}{Ouyang,
  W.}, \bibinfo{author}{Li, H.}, \bibinfo{author}{Wang, X.},
  \bibinfo{year}{2017}.
\newblock \bibinfo{title}{Learning feature pyramids for human pose estimation},
  in: \bibinfo{booktitle}{Proc. IEEE International Conference on Computer
  Vision}, pp. \bibinfo{pages}{1281--1290}.
\bibitem[{Yang et~al.(2016)Yang, Ouyang, Li and Wang}]{yang2016end}
\bibinfo{author}{Yang, W.}, \bibinfo{author}{Ouyang, W.}, \bibinfo{author}{Li,
  H.}, \bibinfo{author}{Wang, X.}, \bibinfo{year}{2016}.
\newblock \bibinfo{title}{End-to-end learning of deformable mixture of parts
  and deep convolutional neural networks for human pose estimation}, in:
  \bibinfo{booktitle}{Proc. IEEE Conference on Computer Vision and Pattern
  Recognition}, pp. \bibinfo{pages}{3073--3082}.
\bibitem[{Yang et~al.(2018)Yang, Ouyang, Wang, Ren, Li and Wang}]{yang20183d}
\bibinfo{author}{Yang, W.}, \bibinfo{author}{Ouyang, W.},
  \bibinfo{author}{Wang, X.}, \bibinfo{author}{Ren, J.}, \bibinfo{author}{Li,
  H.}, \bibinfo{author}{Wang, X.}, \bibinfo{year}{2018}.
\newblock \bibinfo{title}{3d human pose estimation in the wild by adversarial
  learning}, in: \bibinfo{booktitle}{Proc. IEEE Conference on Computer Vision
  and Pattern Recognition}, pp. \bibinfo{pages}{5255--5264}.
\bibitem[{Yang and Ramanan(2013)}]{yang2013articulated}
\bibinfo{author}{Yang, Y.}, \bibinfo{author}{Ramanan, D.},
  \bibinfo{year}{2013}.
\newblock \bibinfo{title}{Articulated human detection with flexible mixtures of
  parts}.
\newblock \bibinfo{journal}{IEEE transactions on pattern analysis and machine
  intelligence} \bibinfo{volume}{35}, \bibinfo{pages}{2878--2890}.
\bibitem[{Zanfir et~al.(2018)Zanfir, Marinoiu and
  Sminchisescu}]{zanfir2018monocular}
\bibinfo{author}{Zanfir, A.}, \bibinfo{author}{Marinoiu, E.},
  \bibinfo{author}{Sminchisescu, C.}, \bibinfo{year}{2018}.
\newblock \bibinfo{title}{Monocular 3d pose and shape estimation of multiple
  people in natural scenes-the importance of multiple scene constraints}, in:
  \bibinfo{booktitle}{Proc. IEEE Conference on Computer Vision and Pattern
  Recognition}, pp. \bibinfo{pages}{2148--2157}.
\bibitem[{Zhang et~al.(2013)Zhang, Zhu and Derpanis}]{zhang2013actemes}
\bibinfo{author}{Zhang, W.}, \bibinfo{author}{Zhu, M.},
  \bibinfo{author}{Derpanis, K.G.}, \bibinfo{year}{2013}.
\newblock \bibinfo{title}{From actemes to action: A strongly-supervised
  representation for detailed action understanding}, in:
  \bibinfo{booktitle}{Proc. IEEE International Conference on Computer Vision},
  pp. \bibinfo{pages}{2248--2255}.
\bibitem[{Zhao et~al.(2018)Zhao, Li, Abu~Alsheikh, Tian, Zhao, Torralba and
  Katabi}]{zhao2018through}
\bibinfo{author}{Zhao, M.}, \bibinfo{author}{Li, T.},
  \bibinfo{author}{Abu~Alsheikh, M.}, \bibinfo{author}{Tian, Y.},
  \bibinfo{author}{Zhao, H.}, \bibinfo{author}{Torralba, A.},
  \bibinfo{author}{Katabi, D.}, \bibinfo{year}{2018}.
\newblock \bibinfo{title}{Through-wall human pose estimation using radio
  signals}, in: \bibinfo{booktitle}{Proc. IEEE Conference on Computer Vision
  and Pattern Recognition}, pp. \bibinfo{pages}{7356--7365}.
\bibitem[{Zhou et~al.(2017)Zhou, Huang, Sun, Xue and Wei}]{zhou2017towards}
\bibinfo{author}{Zhou, X.}, \bibinfo{author}{Huang, Q.}, \bibinfo{author}{Sun,
  X.}, \bibinfo{author}{Xue, X.}, \bibinfo{author}{Wei, Y.},
  \bibinfo{year}{2017}.
\newblock \bibinfo{title}{Towards 3d human pose estimation in the wild: a
  weakly-supervised approach}, in: \bibinfo{booktitle}{Proc. IEEE International
  Conference on Computer Vision}, pp. \bibinfo{pages}{398--407}.
\bibitem[{Zhou et~al.(2016)Zhou, Sun, Zhang, Liang and Wei}]{zhou2016deep}
\bibinfo{author}{Zhou, X.}, \bibinfo{author}{Sun, X.}, \bibinfo{author}{Zhang,
  W.}, \bibinfo{author}{Liang, S.}, \bibinfo{author}{Wei, Y.},
  \bibinfo{year}{2016}.
\newblock \bibinfo{title}{Deep kinematic pose regression}, in:
  \bibinfo{booktitle}{Proc. European Conference on Computer Vision},
  \bibinfo{organization}{Springer}. pp. \bibinfo{pages}{186--201}.
\bibitem[{Zuffi and Black(2015)}]{zuffi2015stitched}
\bibinfo{author}{Zuffi, S.}, \bibinfo{author}{Black, M.J.},
  \bibinfo{year}{2015}.
\newblock \bibinfo{title}{The stitched puppet: A graphical model of 3d human
  shape and pose}, in: \bibinfo{booktitle}{Proc. IEEE Conference on Computer
  Vision and Pattern Recognition}, pp. \bibinfo{pages}{3537--3546}.
\bibitem[{Zuffi et~al.(2012)Zuffi, Freifeld and Black}]{zuffi2012pictorial}
\bibinfo{author}{Zuffi, S.}, \bibinfo{author}{Freifeld, O.},
  \bibinfo{author}{Black, M.J.}, \bibinfo{year}{2012}.
\newblock \bibinfo{title}{From pictorial structures to deformable structures},
  in: \bibinfo{booktitle}{Proc. IEEE Conference on Computer Vision and Pattern
  Recognition}, pp. \bibinfo{pages}{3546--3553}.

\end{thebibliography}

\end{document}